%% file: main_FBS_CSR_ver1.tex
\documentclass[lettersize,journal]{IEEEtran}
\usepackage{amsmath,amsfonts}
\usepackage{algorithm}
\usepackage[noend]{algpseudocode}

\usepackage{array}
\usepackage[caption=false,font=normalsize,labelfont=sf,textfont=sf]{subfig}
\usepackage{textcomp}
\usepackage{stfloats}
\usepackage{url}
\usepackage{verbatim}
\usepackage{graphicx}
\usepackage{cite}
\usepackage{booktabs}
\usepackage{multirow}

\usepackage{color}

\usepackage{amsthm}

\theoremstyle{definition}

\newtheorem*{Proof}{Proof}

\theoremstyle{remark}

\global\long\def\Ours{CSRFM}
\global\long\def\Ourss{CSRFM }

\DeclareMathOperator{\prox}{\mathrm{prox}}

\newcommand{\argmin}{\mathop{\mathrm{argmin}}\limits}

\global\long\def\ZeroElem{\mathbf{0}}
\global\long\def\ElemVideoClean{u}
\global\long\def\VecVideoClean{\mathbf{\ElemVideoClean}}
\global\long\def\ElemCompBack{b}
\global\long\def\CompBack{\mathbf{\ElemCompBack}}
\global\long\def\ElemCompFore{f}
\global\long\def\CompFore{\mathbf{\ElemCompFore}}
\global\long\def\CompSpar{\mathbf{s}}
\global\long\def\CompStri{\mathbf{l}}
\global\long\def\CoefSpar{\mathbf{a}}
\global\long\def\Dict{\mathbf{d}}
\global\long\def\VecZero{\mathbf{0}}
\global\long\def\MatCompBack{\mathbf{B}}

\global\long\def\SetRealNum{\mathbb{R}}

\global\long\def\ALFuncMultiConvSymb{f}
\global\long\def\ALFuncMultiConv#1{\ALFuncMultiConvSymb(#1)}
\global\long\def\ALFuncMultiConvFixed#1#2#3{\ALFuncMultiConvSymb_{#1}^{#2}(#3)}
\global\long\def\ALFuncConvSymb{g}
\global\long\def\ALFuncConv#1#2{\ALFuncConvSymb_{#1}(#2)}

\global\long\def\ALVari#1{\mathbf{x}_{#1}}
\global\long\def\ALNumVari{n}
\global\long\def\ALIdxVari{i}
\global\long\def\ALVaris{\ALVari{1}, \ldots, \ALVari{\ALNumVari}}
\global\long\def\ALDimSet#1{N_{#1}}

\global\long\def\NumPixVer{n_{1}}
\global\long\def\NumPixHor{n_{2}}
\global\long\def\NumFrames{n_{3}}
\global\long\def\ParamStri{ }

\global\long\def\MatImagObsv{\mathbf{V}}
\global\long\def\ImagObsv{\mathbf{v}}

\global\long\def\VarSymDual{\mathbf{z}}
\global\long\def\VarDual#1{\mathbf{z}_{#1}}

\global\long\def\ConsSpar{\eta_{\CompSpar}}
\global\long\def\ConsFore{\eta_{\CompFore}}
\global\long\def\ConsFidel{\varepsilon}

\global\long\def\BallDict{B_{2, 1}^{\ZeroElem}}
\global\long\def\BallSpar{B_{1, \ConsSpar}^{\ZeroElem}}
\global\long\def\BallFore{B_{1, \ConsFore}^{\ZeroElem}}
\global\long\def\BallFidel{B_{2, \ConsFidel}^{\ImagObsv}}

\global\long\def\NormOp#1{\| #1 \|_{\mathrm{op}}}

\global\long\def\NumDict{D}
\global\long\def\ValDict{d}
\global\long\def\NumFuncBack{K}
\global\long\def\ValFuncBack{i}
\global\long\def\FuncBack#1{R_{#1}}
\global\long\def\LinOpBack#1{\mathbf{L}_{#1}}

\global\long\def\ConvA#1{\mathfrak{A}_{#1}}
\global\long\def\ConvD#1{\mathfrak{C}_{#1}}

\global\long\def\ParamOptTV{\lambda_{2}}
\global\long\def\ParamOptSparCoef{\lambda_{1}}

\global\long\def\DiffVer{\mathbf{D}_{\mathrm{v}}}

\global\long\def\DiffHor{\mathbf{D}_{\mathrm{h}}}

\global\long\def\DiffTemp{\mathbf{D}_{\mathrm{t}}}
\global\long\def\DiffTempT{\mathbf{D}_{\mathrm{t}}^{\top}}
\global\long\def\DiffVerTemp{\mathbf{D}_{\mathrm{l}}}
\global\long\def\DiffSpa{\mathbf{D}}

\global\long\def\FuncVec{\mathrm{vec}}

\global\long\def\IterOuter{t}
\global\long\def\IterInner{l}

\global\long\def\StepsizeSubFISTA{\tau}

\global\long\def\StanDevGauss{\sigma}
\global\long\def\RateSparse{p_{\CompSpar}}

\global\long\def\Valbest#1{\textbf{#1}}
\global\long\def\ValSecnd#1{\underline{#1}}

\global\long\def\PSNR{\mathrm{MPSNR}}

\global\long\def\ParamOursLRB{\lambda_{\mathrm{LR}}}
\global\long\def\ParamOursTVFB{\ParamOptTV}

\global\long\def\TextSizeGraph{\small}

\makeatletter
\def\bstctlcite{\@ifnextchar[{\@bstctlcite}{\@bstctlcite[@auxout]}}
\def\@bstctlcite[#1]#2{\@bsphack
	\@for\@citeb:=#2\do{%
		\edef\@citeb{\expandafter\@firstofone\@citeb}%
		\if@filesw\immediate\write\csname #1\endcsname{\string\citation{\@citeb}}\fi}%
	\@esphack}
\makeatother

\begin{document}
\bstctlcite{IEEEexample:BSTcontrol}
\title{Robust Foreground-Background Separation for Severely-Degraded Videos Using Convolutional Sparse Representation Modeling}

\author{Kazuki~Naganuma,~\IEEEmembership{Member,~IEEE,}
	Shunsuke~Ono,~\IEEEmembership{Senior~Member,~IEEE,}
	\thanks{Manuscript received XXX, XXX; revised XXX XXX, XXX.}%
	\thanks{K. Naganuma is with the Institute of Engineering of Tokyo University of Agriculture and Technology, Tokyo, 184-8588, Japan (e-mail: k-naganuma@go.tuat.ac.jp).}
	\thanks{S. Ono is with the Department of Computer Science, Institute of Science Tokyo, Yokohama, 226-8503, Japan (e-mail: ono@c.titech.ac.jp).}
	\thanks{This work was supported in part by JST ACT-X Grant Number JPMJAX23CJ, JST PRESTO under Grant JPMJPR21C4, and JST AdCORP under Grant JPMJKB2307, in part by JSPS KAKENHI under Grant 22H03610, 22H00512, 23H01415, 23K17461, 24K03119, and 24K22291, and in part by Grant-in-Aid for JSPS Fellows under Grant 25KJ0117.}}

\markboth{Journal of \LaTeX\ Class Files,~Vol.~14, No.~8, August~2021}%
{Shell \MakeLowercase{\textit{et al.}}: A Sample Article Using IEEEtran.cls for IEEE Journals}


\maketitle

\begin{abstract}
This paper proposes a foreground-background separation (FBS) method with a novel foreground model based on convolutional sparse representation (CSR). In order to analyze the dynamic and static components of videos acquired under undesirable conditions, such as hardware, environmental, and power limitations, it is essential to establish an FBS method that can handle videos with low frame rates and various types of noise. Existing FBS methods have two limitations that prevent us from accurately separating foreground and background components from such degraded videos. First, they only capture either data-specific or general features of the components. Second, they do not include explicit models for various types of noise to remove them in the FBS process. To this end, we propose a robust FBS method with a CSR-based foreground model. This model can adaptively capture specific spatial structures scattered in imaging data. Then, we formulate FBS as a constrained multiconvex optimization problem that incorporates CSR, functions that capture general features, and explicit noise characterization functions for multiple types of noise. Thanks to these functions, our method captures both data-specific and general features to accurately separate the components from various types of noise even under low frame rates. To obtain a solution of the optimization problem, we develop an algorithm that alternately solves its two convex subproblems by newly established algorithms. Experiments demonstrate the superiority of our method over existing methods using two types of degraded videos: infrared and microscope videos.
\end{abstract}


\section{Introduction}
\label{sec:intro}
\IEEEPARstart{F}{oreground-background} separation (FBS) is a preprocessing technique for decomposing video data into dynamic foreground and static background components~\cite{BOUWMANS20171,Rev_video_process_Bouwmans_2018}.
Since these components are crucial for subsequent processing, such as motion detection~\cite{motion_detection_Huang_2011}, moving object detection~\cite{moving_object_detection_Zhou_2013}, small target detection~\cite{ISTD_DAI_2016,TRD_MolReg_Deng_2022}, background subtraction~\cite{background_subtraction_SOBRAL20144,OTenRPCA_Li_2023}, and cell segmentation~\cite{CellSeg_rev_Vicar_2019}, FBS is widely applied to the various types of videos, such as natural videos, infrared (IR) videos,  and microscope (MS) videos.
However, some video data often suffer from several types of high-level noise~\cite{RevMicImg_Meiniel_2018_TIP,RevMicImg_RICCI2022}, such as Gaussian-like random noise, outliers, missing values, and stripe noise. 
In addition, they may suffer from low frame rates due to hardware limitations and power consumption.
These severe degradations have a deleterious impact on the FBS performance.
Therefore, the development of an FBS method that handles such \textit{severely-degraded} videos can help resolve issues in related fields such as remote sensing, astronomical imaging, and biomedical/biological imaging~\cite{weng2009thermal,Lichtman2005Fluorescence}.

Existing FBS methods are broadly classified into a neural network-based approach~\cite{NI2021,GAAM_Sultana_2021_ICCV,LEE2023,FBSGenerative_Dombrowski_2023_ICCV,UVCNet_FBS_Zhao_2023,FDVP_Miao_2024} and an optimization-based approach~\cite{RPCA,CORPCAFO_jimaging4070090_2018,TRPCA,GNNSSN_Yang_2020,SPMD_wang_2020,GNNLSM_YANG_2020,TVTRPCA_Cao_2016,MSCL_Javed_2017_TIP,TVRPCA,PRPCA,SSSRPCA,SRTC_FBS_Shen_2022,RTD_FBS_Shen_2023}.
The first approach constructs a neural network model that decomposes target videos into foreground and background components, capturing intricate features that are difficult to model manually.
Methods taking the approach accurately separate featureful components from high-quality videos, but do not from severely-degraded videos because they do not incorporate appropriate characterizations for the various types of noise.
In addition, because of the lack of clarity regarding criteria that FBS networks use to estimate these components, they might produce unexplainable results, especially when handling severely-degraded videos. Such lack of explainability can be fatal in the aforementioned fields, which often require clarity in the entire process of extracting meaningful data from observed videos, interpreting the information, and making decisions.



On the other hand, an optimization-based approach formulates FBS as an optimization problem that incorporates regularization functions that model the nature of foreground and background components and functions that characterize the noise. 
By leveraging these functions, this approach effectively separates noise from the components, enabling the development of robust FBS methods.
Unlike a neural network-based approach, an optimization-based approach does not require training data, which makes it particularly suitable for severely-degraded videos. 
In terms of foreground modeling functions, there are two main categories: a sparsity model~\cite{RPCA,CORPCAFO_jimaging4070090_2018,TRPCA,GNNSSN_Yang_2020,SPMD_wang_2020,GNNLSM_YANG_2020} and a smoothness and continuity model~\cite{TVTRPCA_Cao_2016,MSCL_Javed_2017_TIP,TVRPCA,PRPCA,SSSRPCA,SRTC_FBS_Shen_2022,RTD_FBS_Shen_2023}.

The well-known FBS method adopting the first model is robust principal component analysis~\cite{RPCA}, which inspired subsequent methods~\cite{CORPCAFO_jimaging4070090_2018,TRPCA,GNNSSN_Yang_2020,SPMD_wang_2020,GNNLSM_YANG_2020}. 
Although these methods are capable of capturing the sparsity of foreground components, they are unable to accurately estimate foreground components from videos contaminated with high-level noise or sparse noise because these types of noise have sparsely distributed high-intensity values.
The second model characterizes foreground objects as spatial piecewise smoothness and slow motion based on their spatial correlation and temporal continuity, respectively~\cite{TVTRPCA_Cao_2016,MSCL_Javed_2017_TIP,TVRPCA,PRPCA,SSSRPCA,SRTC_FBS_Shen_2022,RTD_FBS_Shen_2023}.
However, this model may fail to separate the foreground component from low frame-rate videos that contain fast-moving objects.
In addition, the two models are limited in their ability to recover the intricate detail structures of foreground objects specific to each video because they only capture features that appear in many videos.

To recover the intricate structures of a foregrounzd component while removing various types of noise, FBS methods need to take an optimization approach equipped with mechanisms that capture data-specific features.
One promising approach to build such an FBS method is by incorporating convolutional sparse representation (CSR)~\cite{SRrev1_Rubinstein_2010,CSC1_Bristow_2013_CVPR,CSC2,CSC3_Wholberg_2016,CSC4_Garcia_2018}. 
CSR models a component as the sum of convolutions between a basis (called a dictionary) and sparse coefficients, and thus efficiently captures specific spatial structures scattered in imaging data.\footnote{In the context of FBS, to capture the static nature of a background component for noiseless videos, some methods introduced SR~\cite{david2009foreground,zhao2011background,SCFBS_Stag_2015_TIP,BREW_DLHPM_Dong_2016_TIP}, which represents a dictionary as a matrix~\cite{SRrev2_Wright_2010,SRrev3_Zhang_2015}.}
Due to its simple representation, CSR can learn a dictionary for data-specific features even from a single severely-degraded video, while guaranteeing interpretability in the FBS process. 
Here, a natural question arises: 
\textit{By adopting CSC and regularization functions to capture data-specific and general features simultaneously, can we establish an FBS method that appropriately separates foreground and background components from severely-degraded videos?}
To answer the question, we need to overcome the following technical difficulties.
\begin{itemize}
	\item To capture both general features and data-specific features, CSR techniques need to be innovated in ranging from the design of an optimization problem to the development of an optimization algorithm.
	\item Optimization problems are typically designed by incorporating terms (e.g., CSR, noise characterizations, and video regularization functions). However, this increases the number of parameters that need to be adjusted manually and carefully. 
	\item Although CSR has simpler structure than neural networks, optimization problems involving CSR are still nonconvex. Consequently, this presents a considerable challenge in developing an optimization algorithm that obtains a solution with meaningful background and foreground components and a basis of CSR.
\end{itemize}




Based on the above discussion, we propose a robust FBS with CSR-based foreground modeling (\Ours).
The main contributions are as follows.
\begin{itemize}
	\item We introduce CSR into FBS as foreground modeling.
	Since foreground objects are scattered in several frames, CSR can adaptively capture their shapes and edges. Therefore, \Ourss thus accurately separates them even under highly noisy and low frame rate cases.
	\item We formulate FBS as a constrained multiconvex optimization problem incorporating the CSR-based modeling function. Our optimization problem is designed to separate foreground and background components while removing Gaussian, sparse, and stripe noise.
	In addition, instead of adding terms characterizing these types of noise to the objective function, they are imposed as hard constraints. This transforms complicated interdependent hyperparameters into independent parameters that can be set based on statistical information for noise.
	\item We develop an algorithm for solving optimization problems based on an alternating minimization (ALM). ALM requires solving convex subproblems with certain variables fixed. In our case, solvers for two convex subproblems are established based on the preconditioned primal-dual splitting algorithm (P-PDS)~\cite{DP-PDS,P-PDS_naganuma_2023} and the fast iterative shrinkage-thresholding algorithm (FISTA)~\cite{FISTA}. P-PDS can automatically adjust the appropriate stepsizes based on the problem structure that changes with each iteration. In addition, FISTA also obtains a solution of the subproblem at a fast convergence rate without stepsize adjustment. These two algorithms allow us to solve each subproblem efficiently. 
\end{itemize}
Experiments using IR videos and MS videos demonstrate the superiority of \Ourss to existing methods.

This paper is organized as follows.
Section~\ref{sec:preliminaries} gives preliminaries on CSR and ALM. 
In Section~\ref{sec:proposed}, we present \Ours. 
Experimental results and discussion are given in Section~\ref{sec:experiments}. 
Finally, we conclude the paper in Section~\ref{sec:conclusion}.

The preliminary version of this work, without the generalization of our method, comprehensive experimental comparison, or deeper discussion, has appeared in conference proceedings~\cite{CSRFBS_naganuma_2023_APSIPA}.

\section{Preliminaries}
\label{sec:preliminaries}
\subsection{Notation}
In this paper, 
we denote the sets of real numbers as $\SetRealNum$. Vectors and matrices are denoted by lowercase bold letters (e.g., $\mathbf{x}$) and uppercase bold letters (e.g., $\mathbf{X}$), respectively. The $i$-th element of a vector $\mathbf{x}$ is denoted by $x_{i}$ or $[\mathbf{x}]_{i}$. Similarly, the element at the $i$-th row and $j$-th column of a matrix $\mathbf{X}$ is denoted by $X_{i,j}$ or $[\mathbf{X}]_{i,j}$.
The $\ell_{1}$, $\ell_{2}$, and operator norms are defined as $\| \mathbf{x} \|_{1} := \sum_{i} | \mathbf{x}_{i} |$, $\| \mathbf{x} \|_{2} := \sqrt{\sum_{i} \mathbf{x}_{i}^{2} }$, and $\| \mathfrak{L} \|_{\mbox{op}} := \sup_{\mathbf{x} \neq \mathbf{0}} \| \mathfrak{L}(\mathbf{x}) \|_{2} / \| \mathbf{x} \|_{2}$, respectively.
We denote the $\ell_{p}$ ball centered by $\mathbf{c}$ with a radius $\alpha$ as $B_{p, \alpha}^{\mathbf{c}} := \{ \mathbf{y} \, | \, \| \mathbf{y} - \mathbf{c} \|_{p} \leq \alpha \}$.
The adjoint operator of a linear operator $\mathfrak{L}$ is denoted as $\mathfrak{L}^{*}$. 

\subsection{Convolutional Sparse Representation (CSR)}
CSR models a signal $\mathbf{x}$ as the sum of convolutions between a dictionary basis $\Dict := \{ \Dict_{1}, \ldots, \Dict_{\NumDict} \}$ and a sparse coefficient $\CoefSpar := \{\CoefSpar_{1}, \ldots, \CoefSpar_{\NumDict} \}$~\cite{SRrev1_Rubinstein_2010,CSC1_Bristow_2013_CVPR,CSC2,CSC3_Wholberg_2016,CSC4_Garcia_2018}:
\begin{equation}
	\min_{\Dict, \CoefSpar }  \:
	\frac{1}{2} \left\| \mathbf{x} - \sum_{\ValDict = 1}^{\NumDict} \Dict_{\ValDict} * \CoefSpar_{\ValDict} \right\|_{2}^{2} 
	+ \lambda \sum_{\ValDict = 1}^{\NumDict} \|\CoefSpar_{\ValDict}\|_{1} \:
	\mathrm{s.t.} \: 
	\begin{cases}
		\Dict_{1} \in \BallDict, \\
		\vdots, \\
		\Dict_{\NumDict} \in \BallDict, \\
	\end{cases}
	\label{prob:convolutional_sparse_representation}
\end{equation}
where $*$ denotes the convolution operator. 
The first term guarantees the fidelity between the signal $\mathbf{x}$ and the sum of convolutions between $\Dict$ and $\CoefSpar$.
The second term promotes the sparsity of the coefficient $\CoefSpar$. 
By combining these terms using the appropriate balancing parameter $\lambda > 0$, CSR can capture specific spatial structure scattered throughout the signal $\mathbf{x}$. The unit $\ell_{2}$ ball constraints prevent the basis $\Dict$ from absorbing all of the energy of the target signal $\mathbf{x}$. 


\subsection{Alternating Minimization for Multiconvex Optimization}
This section introduces alternating minimization, which solves multiconvex optimization problems.
We consider a multiconvex optimization problem of the following form:
\begin{equation}
\label{prob:gene_AL}
\min_{\ALVaris} \ALFuncMultiConv{\ALVaris} + \sum_{\ALIdxVari = 1}^{\ALNumVari} \ALFuncConv{\ALIdxVari}{\ALVari{\ALIdxVari}},
\end{equation}
where $\ALFuncMultiConvSymb : \SetRealNum^{\ALDimSet{1}} \times \cdots \times \SetRealNum^{\ALDimSet{\ALNumVari}} \rightarrow \SetRealNum$ is a differentiable and \textit{block multiconvex} function\footnote{
	A function $\ALFuncMultiConv{\ALVaris}$ is called \textit{block multiconvex} if, for each $\ALIdxVari$, $\ALFuncMultiConvSymb$ is a convex function of $\ALVari{\ALIdxVari}$ while all the other blocks are fixed.
} and $\ALFuncConvSymb_{\ALIdxVari}: \SetRealNum^{\ALDimSet{\ALIdxVari}} \rightarrow (-\infty, \infty]$ ($\forall \ALIdxVari = 1, \ldots, \ALNumVari$) are proper lower-semicontinuous convex functions.

\input{fig_tex/overall.tex}

Under some conditions, the following procedures generate a sequence that converges to an optimal solution in a certain region close to an initial point~\cite{xu2013block}: 
\begin{equation}
	\label{eq:AL}
	\left\lfloor
	\begin{array}{ll}
		\ALVari{1}^{(\IterOuter + 1)} 
		& \leftarrow \argmin_{\ALVari{1}} \ALFuncMultiConvFixed{1}{(\IterOuter)}{\ALVari{1}} + \ALFuncConv{1}{\ALVari{1}^{(\IterOuter)}}, \\
		& \vdots \\
		\ALVari{\ALNumVari}^{(\IterOuter + 1)} 
		& \leftarrow \argmin_{\ALVari{\ALNumVari}} \ALFuncMultiConvFixed{\ALNumVari}{(\IterOuter)}{\ALVari{\ALNumVari}} + \ALFuncConv{\ALNumVari}{\ALVari{\ALNumVari}^{(\IterOuter)}}, \\
	\end{array}
	\right.
\end{equation}
where for any $\ALIdxVari = 1, \ldots, \ALNumVari$ 
\begin{equation}
	\ALFuncMultiConvFixed{\ALIdxVari}{(\IterOuter)}{\ALVari{\ALIdxVari}} = \ALFuncMultiConv{\ALVari{1}^{(\IterOuter + 1)}, \ldots, \ALVari{\ALIdxVari - 1}^{(\IterOuter + 1)}, \ALVari{\ALIdxVari}, \ALVari{\ALIdxVari + 1}^{(\IterOuter)}, \ldots, \ALVari{\ALNumVari}^{(\IterOuter)}}.
\end{equation}

\section{Proposed Method}
\label{sec:proposed}
Figure~\ref{fig:overall} shows a general diagram of the proposed method, \Ours. 
In the following, we first introduce an observation model with three types of noise. Based on the model, we then formulate the FBS problem as a constrained multiconvex optimization problem involving CSR-based foreground modeling. 
Finally, we describe an ALM-based algorithm to solve the optimization problem.

\subsection{Problem Formulation}
Let $\MatImagObsv_{1}, \ldots, \MatImagObsv_{\NumFrames} \in \mathbb{R}^{\NumPixVer \times \NumPixHor}$ be $\NumFrames$ observed frames. We handle the observed frames in a vectorized form as $\ImagObsv = [\FuncVec(\MatImagObsv_{1})^{\top} \: \cdots \: \FuncVec(\MatImagObsv_{\NumFrames})^{\top}]^{\top} \in \mathbb{R}^{\NumPixVer \NumPixHor \NumFrames}$, where $\FuncVec : \mathbb{R}^{\NumPixVer \times \NumPixHor} \rightarrow \mathbb{R}^{\NumPixVer \NumPixHor}$ represents the vectorization operator. 
Then, consider that the observed video is modeled by 
\begin{equation}
	\label{eq:observation}
	\ImagObsv = \bar{\CompFore} + \bar{\CompBack} + \bar{\CompSpar} + \bar{\CompStri} + \mathbf{n},
\end{equation}
where $\bar{\CompFore}$, $\bar{\CompBack}$, $\bar{\CompSpar}$, $\bar{\CompStri}$, and $\mathbf{n}$ are the true foreground component, the true background component, sparsely distributed noise (e.g., missing values and outliers), stripe noise, and random noise, respectively.

Based on the model in~\eqref{eq:observation}, we newly formulate FBS as a multiconvex optimization problem of the following form:
\begin{align}
	\min_{\substack{\CompFore,\CompBack, \CompSpar, \CompStri \\ \Dict, \CoefSpar} } \: &
	\frac{1}{2} \left\|\CompFore - \sum_{\ValDict = 1}^{\NumDict} \Dict_{\ValDict}*\CoefSpar_{\ValDict} \right\|_{2}^2
	+ \ParamOptSparCoef \sum_{\ValDict = 1}^{\NumDict} \|\CoefSpar_{\ValDict}\|_{1} 
	+ \ParamOptTV \| \DiffSpa(\CompFore + \CompBack)\|_{1} 
	\nonumber \\
	& + \FuncBack{0}(\CompBack)
	+ \sum_{\ValFuncBack = 1}^{\NumFuncBack} \FuncBack{i}(\LinOpBack{i}\CompBack)
	+ \ParamStri \| \CompStri \|_{1} \nonumber \\
	\mathrm{s.t.} \: & 
	\begin{cases}
		\CompFore \in \BallFore, \\
		\CompSpar \in \BallSpar, \\
		\CompFore + \CompBack + \CompSpar + \CompStri \in \BallFidel, \\
		\DiffVerTemp \CompStri = \ZeroElem, \\
		\Dict_{1} \in \BallDict, \ldots, \Dict_{\NumDict} \in \BallDict.
	\end{cases}
	\label{prob:CSC_FBS}
\end{align}
These terms and constraints have the following roles.

\input{./algo/algo_CSR_FBS}

\input{./algo/algo_sub_FBSA_PDS_no_conj}

\subsubsection{CSR for $\CompFore$}
The first and second terms (with a balancing parameter $\ParamOptSparCoef > 0$) and the constraints in fifth row build CSR, which captures similar structures over frames by simultaneously estimating a basis $\Dict_{1}, \ldots, \Dict_{\NumDict}$ (representing the edges or shapes of foreground objects) and their corresponding sparse coefficients $\CoefSpar_{1}, \ldots, \CoefSpar_{\NumDict}$ (representing the locations).
By referring to the sum of the convolutions of $\Dict_{1}, \ldots, \Dict_{\NumDict}$ and $\CoefSpar_{1}, \ldots, \CoefSpar_{\NumDict}$ estimated simultaneously with $\CompFore$, \Ourss can recover the structures of $\CompFore$ more accurately.

\subsubsection{A Sparsity Constraint for $\CompFore$}
The first constraint guarantees the sparsity of the foreground component $\CompFore$ with the $\ell_{1}$ ball with a radius $\ConsFore$. By varying $\ConsFore$, we can control the sparsity of  $\CompFore$. 
Using this constraint instead of a sparse term avoids adjusting the hyperparameter in response to changes of background modeling functions or noise conditions.

\input{./algo/algo_sub_D_FISTA}

\subsubsection{Background Modeling}
The sum of the fourth and fifth terms is a general form of background modeling including linear operators $\LinOpBack{1}, \ldots, \LinOpBack{\NumFuncBack}$ and proper lower-semicontinuous convex functions $\FuncBack{0}, \ldots, \FuncBack{\NumFuncBack}$.
Some examples of applying specific background modeling to the general form will be shown in Section~\ref{ssec:ex_back}.

\subsubsection{Total Variation Regularization for $\CompFore$ and $\CompBack$}
The third term promotes the spatial piecewise smoothness of $\CompFore$ and $\CompBack$ by the total variation regularization with a balancing parameter $\ParamOptTV > 0$.
The total variation regularization incoprates a spatial difference operator $\DiffSpa := [\DiffVer^{\top} ~ \DiffHor^{\top}]^{\top} \in \mathbb{R}^{2 \NumPixVer \NumPixHor \NumFrames \times \NumPixVer \NumPixHor \NumFrames}$, where $\DiffVer$ and $\DiffHor$ are the vertical and horizontal difference operators with the Neumann boundary condition: let $[\FuncVec(\mathbf{Y}_{1})^{\top} \: \cdots \: \FuncVec(\mathbf{Y}_{\NumFrames})^{\top}]^{\top} = \DiffVer\mathbf{x}$ and $[\FuncVec(\mathbf{W}_{1})^{\top} \: \cdots \: \FuncVec(\mathbf{W}_{\NumFrames})^{\top}]^{\top} = \DiffHor\mathbf{x}$ for a time series component $\mathbf{x} = [\FuncVec(\mathbf{X}_{1})^{\top} \: \cdots \: \FuncVec(\mathbf{X}_{\NumFrames})^{\top}]^{\top}$, then for any $k = 1, \ldots, \NumFrames$
\begin{align}
	[\mathbf{Y}_{k}]_{i, j} = 
	\begin{cases}
		[\mathbf{X}_{k}]_{i + 1, j} - [\mathbf{X}_{k}]_{i, j}, & \text{if }i < \NumPixVer; \\
		0, & \text{otherwise,}
	\end{cases}
	\label{eq:diff_Vec} \\
	[\mathbf{W}_{k}]_{i, j} = 
	\begin{cases}
		[\mathbf{X}_{k}]_{i, j + 1} - [\mathbf{X}_{k}]_{i, j}, & \text{if }j < \NumPixHor; \\
		0, & \text{otherwise.}
	\end{cases}
\end{align}

\subsubsection{Handling Three Types of Noise}
The second constraints guarantee the sparsity of $\CompSpar$.
The third constraint guarantees the data-fidelity to an observed video $\ImagObsv$. 
Using these constraints instead of data-fidelity and sparse terms allows us to easily adjust the hyperparameters, since the parameters $\ConsFidel$ and $\ConsSpar$ can be determined based only on the statistical information of these types of noise, rather than relative weights to the terms of the objective function.
Indeed, this kind of constrained formulation has played an important role in facilitating the parameter setting of signal recovery problems~\cite{CSALSA,EPIpre,ono_2015,ono2017primal,ono_2019}.


The sixth term controls the intensity of a stripe noise component $\CompStri$.
The fourth constraint is the flatness constraint~\cite{FC_naganuma_2022}, which captures the vertical flatness and temporal invariance by imposing zero to all the values of the vertical and temporal gradients of $\CompStri$ using a difference operator 
$\DiffVerTemp = [\DiffVer^{\top} ~ \DiffTemp^{\top}]^{\top} \in \mathbb{R}^{2 \NumPixVer \NumPixHor \NumFrames \times \NumPixVer \NumPixHor \NumFrames}$.\footnote{For data with horizontally featured stripe noise, we rotate the data $90$ degrees in the spatial direction before optimization.}
Here, $\DiffVer$ is the vertical difference operator defined in Eq.~\eqref{eq:diff_Vec} and $\DiffTemp$ is the temporal difference operator with the Neumann boundary condition: let $[\FuncVec(\mathbf{Y}_{1})^{\top} \: \cdots \: \FuncVec(\mathbf{Y}_{\NumFrames})^{\top}]^{\top} = \DiffTemp\mathbf{x}$, then for any $k = 1, \ldots, \NumFrames$
\begin{equation}
	[\mathbf{Y}_{k}]_{i, j} = 
	\begin{cases}
		[\mathbf{X}_{k + 1}]_{i, j} - [\mathbf{X}_{k}]_{i, j}, & \text{if }k < \NumFrames; \\
		0, & \text{otherwise.}
	\end{cases}
\end{equation}
Actually, this flatness constraint plays an important role to accurately characterize the stripe noise in the context of remote sensing data analysis~\cite{sato2024robusthyperspectralanomalydetection,Naganuma_unmixing_2024,Takemoto_2024}.
If stripe noise changes over time, the difference operator $\DiffVerTemp$ is designed as $\DiffVerTemp = \DiffVer \in \mathbb{R}^{\NumPixVer \NumPixHor \NumFrames \times \NumPixVer \NumPixHor \NumFrames}$.

\input{./fig_tex/GT_images.tex}

\subsection{Optimization}

Using indicator functions\footnote{For a given nonempty closed set $C$, the indicator function of $C$ is defined as $\iota_{C}(\mathcal{X}):=0$, if $\mathcal{X}\in C$; $\infty$, otherwise.} $\iota_{\BallFidel}$, $\iota_{\BallFore}$, and $\iota_{\BallSpar}$ of norm balls $\BallFidel$, $\BallSpar$, and $\BallFore$, 
we reformulate Prob.~\eqref{prob:CSC_FBS} into the following equivalent optimization problem:
\begin{align}
\min_{\substack{\CompFore, \CompBack, \CompSpar, \\ \CompStri, \CoefSpar, \Dict}} \:
& 
\frac{1}{2} \left\|\CompFore - \sum_{\ValDict = 1}^{\NumDict} \Dict_{\ValDict}*\CoefSpar_{\ValDict} \right\|_{2}^{2}
+ \ParamOptSparCoef \sum_{\ValDict = 1}^{\NumDict} \left\| \CoefSpar_{\ValDict} \right\|_{1}
+ \ParamOptTV \left\| \mathbf{D}(\CompFore + \CompBack) \right\|_{1}
\nonumber \\
& 
+ \FuncBack{0}(\CompBack)
+ \sum_{\ValFuncBack = 1}^{\NumFuncBack} \FuncBack{\ValFuncBack}(\LinOpBack{\ValFuncBack}\CompBack)
+ \ParamStri \| \CompStri \|_{1} 
+ \iota_{\BallFore}(\CompFore) 
+ \iota_{\BallSpar}(\CompSpar) 
\nonumber \\
& 
+ \iota_{\BallFidel} (\CompFore + \CompBack + \CompSpar + \CompStri)
+ \iota_{\{\ZeroElem\}}(\DiffVerTemp \CompStri)
+ \sum_{\ValDict = 1}^{\NumDict} \iota_{\BallDict}(\Dict_{\ValDict}).
\label{prob:FBS_main_reform}
\end{align}
Since the first term is a multiconvex function and the other terms are convex functions, Prob.~\eqref{prob:CSC_FBS} can be seen as a specific form of Prob.~\eqref{prob:gene_AL}.
To solve this type of optimization problems, one often updates the variables one by one. 
However, since each the update process strongly depends on the current states of the other variables, this approach can produce a sequence that converges to a local minimum that includes meaningless components. 
Therefore, we take an approach to update some variables simultaneously.
Specifically, we set
\begin{align}
	\ALFuncMultiConv{\CompFore, \CompBack, \CompSpar, \CompStri, \CoefSpar, \Dict} 
	= 
	& \frac{1}{2}\left\|\CompFore - \sum_{\ValDict = 1}^{\NumDict} \Dict_{\ValDict}*\CoefSpar_{\ValDict} \right\|_{2}^{2},
	\nonumber \\
	\ALFuncConv{1}{\CompFore, \CompBack, \CompSpar, \CompStri, \CoefSpar}
	= 
	& \ParamOptSparCoef \sum_{\ValDict = 1}^{\NumDict} \| \CoefSpar_{\ValDict}\|_{1}
	+ \ParamOptTV \| \mathbf{D}(\CompFore + \CompBack)\|_{1}
	+ \FuncBack{0}(\CompBack)
	\nonumber \\
	& + \sum_{\ValFuncBack = 0}^{\NumFuncBack} \FuncBack{\ValFuncBack}(\LinOpBack{\ValFuncBack}\CompBack)  
	+ \ParamStri \| \CompStri \|_{1} 
	+ \iota_{\BallFore}(\CompFore) 
	\nonumber \\
	& + \iota_{\BallSpar}(\CompSpar) 
	+ \iota_{\BallFidel} (\CompFore + \CompBack + \CompSpar + \CompStri)
	\nonumber \\
	& + \iota_{\{\ZeroElem\}}(\DiffVerTemp \CompStri),
	\nonumber \\
	\ALFuncConv{2}{\Dict} 
	= 
	& \sum_{\ValDict = 1}^{\NumDict} \iota_{\BallDict}(\Dict_{\ValDict}),
	\label{eq:def_obj_func}
\end{align}
and alternately solving two subproblems for $\ALFuncMultiConvSymb_{1}^{\IterOuter} + \ALFuncConvSymb_{1}$ and $\ALFuncMultiConvSymb_{2}^{\IterOuter} + \ALFuncConvSymb_{2}$ to update the variables $\CompFore$, $\CompBack$, $\CompSpar$, $\CompStri$, $\CoefSpar$ and the variable $\Dict$.

Algorithm~\ref{algo:CSR_FBS} shows an ALM-based algorithm to solve Prob.~\eqref{algo:CSR_FBS}.\footnote{It is difficult to confirm whether Algorithm~\ref{algo:CSR_FBS} satisfies the theoretical convergence condition described in~\cite{xu2013block}. However, we observed in expereiments (see Section~\ref{sec:experiments}) that our algorithm generate stable results.}
Here, we describe the details of algorithms solving each subproblem.


\subsubsection{Updating $\CompFore$, $\CompBack$, $\CompSpar$, $\CompStri$, and $\CoefSpar$}

With $\Dict^{(\IterOuter)}$ fixed, the first subproblem of Prob.~\eqref{prob:FBS_main_reform} with respect to $\CompFore$, $\CompBack$, $\CompSpar$, $\CompStri$, and $\CoefSpar$ is written as the following optimization problem:
\begin{align}
	\min_{\substack{\CompFore, \CompBack, \CompSpar, \\ \CompStri, \CoefSpar}} \:
	& \iota_{\BallFore}(\CompFore) 
	+ \FuncBack{0}(\CompBack)
	+ \iota_{\BallSpar}(\CompSpar) 
	+ \ParamStri \| \CompStri \|_{1} 
	+ \ParamOptSparCoef \sum_{\ValDict = 1}^{\NumDict} \| \CoefSpar_{\ValDict}\|_{1} \nonumber \\
	& + \sum_{\ValFuncBack = 1}^{\NumFuncBack} \FuncBack{\ValFuncBack}(\LinOpBack{\ValFuncBack}\CompBack)  
	+ \frac{1}{2} \left\|\CompFore - \sum_{\ValDict = 1}^{\NumDict} \Dict_{\ValDict}^{(\IterOuter)}*\CoefSpar_{\ValDict} \right\|_{2}^{2}
	\nonumber \\
	& 
	+ \ParamOptTV \| \mathbf{D}(\CompFore + \CompBack)\|_{1}
	+ \iota_{\BallFidel} (\CompFore + \CompBack + \CompSpar + \CompStri)
	+ \iota_{\{\ZeroElem\}}(\DiffVerTemp \CompStri). 
	\label{prob:sub_FBSA_PDS}
\end{align}
Prob.~\eqref{prob:sub_FBSA_PDS} can be solved by P-PDS~\cite{DP-PDS,P-PDS_naganuma_2023}.



We show the detailed algorithm in Algorithm~\ref{algo:sub_FBSA_PDS}. Following a method~\cite{P-PDS_naganuma_2023} that selects reasonable stepsize, the parameters $\gamma_{\CompFore},\gamma_{\CompBack},\gamma_{\CompSpar},\gamma_{\CoefSpar}$, and $\gamma_{\mathbf{z}}$ are determined as follows:
\begin{align}
	\label{eq:stepsize_setting_P_PDS}
		& \gamma_{\CompFore} = \tfrac{1}{10}, 
		\gamma_{\CompBack} = \tfrac{1}{ 9 + \sum_{\ValFuncBack = 1}^{\NumFuncBack} \NormOp{\LinOpBack{\ValFuncBack}}^{2} }, 
		\gamma_{\CompSpar} = 1, 
		\gamma_{\CompStri} = \tfrac{1}{9}, \nonumber \\
		& \gamma_{\CoefSpar} = \tfrac{1}{ \sum_{\ValDict = 1}^{\NumDict} \NormOp{\ConvD{\Dict_{\ValDict}}}^{2} }, 
		\gamma_{\VarSymDual} = \tfrac{1}{ 4 + \NumDict },
\end{align}
where $\ConvD{\Dict_{\ValDict}^{(\IterOuter)}}(\CoefSpar) = \Dict_{\ValDict}^{(\IterOuter)}*\CoefSpar$.
The proximity operators\footnote{The proximity operator of a proper lower semicontinuous convex function $f$ with a parameter $\alpha > 0$ is defined as $\prox_{\alpha f}(\mathbf{x}) := \mathrm{argmin}_{\mathbf{y}} f(\mathbf{y}) + \frac{1}{2\alpha}\|\mathbf{x} - \mathbf{y}\|_{2}^{2}$.} of $\|\cdot\|_{1}$, $\frac{1}{2} \|\cdot\|_{2}^{2}$, and $\iota_{\BallFidel}$ (in lines 11, 16, and 18 of Algorithm~\ref{algo:sub_FBSA_PDS}) are respectively calculated as
\begin{align}
	[\mathrm{prox}_{\alpha \|\cdot\|_{1}}(\mathbf{x})]_{i}
	& = \mathrm{sgn}(x_{i})\max\{|x_{i}|-\alpha, 0\},
	\label{eq:prox_of_l1} \\
	\mathrm{prox}_{\frac{\alpha}{2} \|\cdot\|_{2}^{2}}(\mathbf{x}) 
	& = \frac{\mathbf{x}}{1 + \alpha},
	\label{eq:prox_of_l2} \\
	\mathrm{prox}_{\alpha \iota_{\BallFidel}}(\mathbf{x}) 
	& = \begin{cases} 
		\mathbf{x}, & \mathrm{if} \: \mathbf{x} \in \BallFidel; \\ 
		\ImagObsv + \frac{\varepsilon(\mathbf{x}-\ImagObsv)}{\|\mathbf{x}-\ImagObsv\|_{2}}, & \mathrm{otherwise}. 
	\end{cases} 
	\label{eq:prox_of_l2ball}
\end{align}
The proximity operator of $\iota_{B_{1, \eta}^{\ZeroElem}}$ can be efficiently computed by the $\ell_{1}$ ball projection algorithm~\cite{fast_l1_ball_projection}.

\input{./table_tex/params.tex}

\input{./table_tex/params_CSR.tex}

\input{./table_tex/result_PSNR_SSIM.tex}

\input{./table_tex/result_PSNR_SSIM_gs.tex}

\input{./table_tex/result_PSNR_SSIM_gsst.tex}

\subsubsection{Updating $\Dict$}
With $\CompFore$, $\CompBack$, $\CompSpar$, $\CompStri$, and $\CoefSpar$ fixed, the subproblem of Prob.~\eqref{prob:CSC_FBS} with respect to $\Dict$ is written as the following optimization problem:
\begin{equation}
	\label{prob:sub_D_FISTA}
	\min_{\Dict}
	\frac{1}{2} \left\|\CompFore^{(\IterOuter + 1)} - \sum_{\ValDict = 1}^{\NumDict}\Dict_{\ValDict}*\CoefSpar_{\ValDict}^{(\IterOuter + 1)} \right\|_{2}^{2} 
	+ \sum_{\ValDict = 1}^{\NumDict} \iota_{\BallDict}(\Dict_{\ValDict}).
\end{equation}
Prob.~\eqref{prob:sub_D_FISTA} is can be approximately solved by FISTA~\cite{FISTA}.
Algorithm~\ref{algo:sub_D_FISTA} shows the detailed processes.
The stepsize parameter $\gamma$ is set as $\gamma = \frac{1}{\sum_{\ValDict = 1}^{\NumDict} \NormOp{\ConvA{\CoefSpar_{\ValDict}}}^{2}}$, where $\ConvA{\CoefSpar_{\ValDict}}(\Dict) = \Dict*\CoefSpar_{\ValDict}$. 

\subsection{Examples of Applying Specific Background Modeling}
\label{ssec:ex_back}
We apply some specific functions into the background modeling functions in Prob.~\eqref{prob:CSC_FBS}.

First, let us consider low-rank modeling, which often appears in numerous FBS methods.
This modeling promotes the low rankness of a background component matrix $\MatCompBack = [\CompBack_{1}, \ldots, \CompBack_{\NumFrames}] \in \mathbb{R}^{\NumFrames \times \NumPixVer \NumPixHor}$ using the nuclear norm, defined as 
\begin{equation}
	\| \MatCompBack \|_{*} := \sum_{i}^{q} s_{i}(\MatCompBack),
\end{equation}
where $s_{i}(\MatCompBack)$ is the $i$-th singular value of $\MatCompBack$ and $q = \min\{\NumPixVer\NumPixHor, \NumFrames\}$.
Letting $\NumFuncBack = 0$ and $\FuncBack{0} = \ParamOursLRB\| \cdot \|_{*}$, we can apply the low-rank modeling to Prob.~\eqref{prob:CSC_FBS}.
By using a vectorizing operator $\mathrm{vec}(\MatCompBack) = [\CompBack_{1}^{\top}, \ldots, \CompBack_{\CompBack}^{\top}]^{\top}$ and a matrixizing operator $\mathrm{mat}(\CompBack) = [\CompBack_{1}, \ldots, \CompBack_{\NumFrames}]$, the steps 7 and 8 in Algorithm~\ref{algo:sub_FBSA_PDS} become 
\begin{align}
	& \CompBack^{\prime} \leftarrow \CompBack^{(\IterInner)} - \gamma_{\CompBack} (\mathbf{D}^{\top}\VarDual{3}^{(\IterInner)} + \VarDual{4}^{(\IterInner)}), \\
	& \CompBack^{(\IterInner + 1)} \leftarrow \mathrm{vec}(\prox_{\gamma_{\CompBack}\ParamOursLRB \|\cdot\|_{*}}(\mathrm{mat}(\CompBack^{\prime}))),
\end{align}
and the steps 20-22 are eliminated.
Here, the proximity operator of $\| \cdot \|_{*}$ with a parameter $\gamma > 0$ can be calculated as follows: let $\MatCompBack = \mathbf{U}_{\MatCompBack} \mathbf{S}_{\MatCompBack} \mathbf{V}_{\MatCompBack}^{\top}$ be the singular value decomposition of $\MatCompBack$, then
\begin{equation}
	\mathrm{prox}_{\gamma \| \cdot \|_{*}}(\MatCompBack) = \mathbf{U}_{\MatCompBack} \widetilde{\mathbf{S}}_{\gamma} \mathbf{V}_{\MatCompBack}^{\top},
\end{equation}
where $\widetilde{\mathbf{S}}_{\gamma}$ is a diagonal matrix whose $i$-th element is $\max\{s_{i}(\MatCompBack) - \gamma, 0\}$.
The stepsize parameter $\gamma_{\CompBack}$ is determined as $\gamma_{\CompBack} = 1/9$.

As a second example, let us consider static scene constraint modeling~\cite{SSC_naganuma_2023}, which characterizes a background component as exactly static.
This modeling makes an algorithm relatively efficient for low-rank modeling by imposing zero to all the values of the temporal gradient of $\CompBack$ as follows:
\begin{equation}
	\DiffTemp\CompBack = \VecZero.
\end{equation}
Using this constraint is identical to adding $\iota_{\{\VecZero\}}(\DiffTemp\CompBack)$ into the objective function.
Therefore, by removing $\FuncBack{0}(\CompBack)$ and letting $\NumFuncBack = 1$, $\LinOpBack{1} = \DiffTemp$, and $\FuncBack{1} = \iota_{\{\VecZero\}}$, we can apply the static scene constraint modeling to Prob.~\eqref{prob:CSC_FBS}.
In Algorithm~\ref{algo:sub_FBSA_PDS}, the steps 7, 8, 21, and 22 respectively become
\begin{align}
	& \CompBack^{\prime} \leftarrow \CompBack^{(\IterInner)} - \gamma_{\CompBack} (\DiffTempT\VarDual{1, 1}^{(\IterInner)} + \DiffSpa^{\top}\VarDual{3}^{(\IterInner)} + \VarDual{4}^{(\IterInner)}), \\
	& \CompBack^{(\IterInner + 1)} \leftarrow \CompBack^{\prime}, \\
	& \VarDual{1, 1}^{\prime} \leftarrow \VarDual{1, 1}^{(\IterInner)} + \gamma_{\VarSymDual} \DiffTemp \widetilde{\CompBack}, \\
	& \VarDual{1, 1}^{(\IterInner + 1)} 
	\leftarrow 
	\VarDual{1, 1}^{\prime}.
\end{align}
The stepsize parameter $\gamma_{\CompBack}$ is determined as $\gamma_{\CompBack} = 1/13$.

\section{Experiments}
\label{sec:experiments}
To demonstrate the effectiveness of \Ours, we perform FBS experiments on low frame-rate videos with several types of noise. 
In the experiments, we compare \Ourss to seven existing FBS methods: RPCA~\cite{RPCA}, GNNLSM~\cite{GNNLSM_YANG_2020}, TVRPCA~\cite{TVRPCA}, PRPCA~\cite{PRPCA}, SRTC~\cite{SRTC_FBS_Shen_2022}, SS-RTD~\cite{RTD_FBS_Shen_2023}, and FactorDVP-T~\cite{FDVP_Miao_2024}.
As the background modeling of \Ours, we adopt the low-rank modeling (referred to as \Ourss (LR)) and the static scene constraint modeling (referred to as \Ourss (SC)), described in Section~\ref{ssec:ex_back}.

\subsection{Datasets}
As the ground-truth IR and MS videos, we used six IR videos with static background from CAMEL Dataset\footnote{\url{https://camel.ece.gatech.edu/}} (cropped into a size of $128 \times 128 \times 30$)~\cite{CAMEL_DATASET,CAMEL_DATASET2}, one IR video from Bird Dataset\footnote{\url{ https://universe.roboflow.com/antiuav-9-aniket/bird-6le8u }} (cropped into a size of $200 \times 200$ and sampled $30$ frames)~\cite{bird-6le8u_dataset}, one MS video from T cell signaling proteins 3D confocal microscope movies\footnote{\url{https://kilthub.cmu.edu/articles/dataset/T_cell_signaling_proteins_3D_confocal_microscope_movies/9963566}} (cropped into a size of $128 \times 128$ and sampled $30$ frames), and one MS video from Cell Tracking Challenge Datasets\footnote{\url{http://celltrackingchallenge.net/2d-datasets/}} (cropped into a size of $200 \times 200$ and sampled $30$ frames).
We sampled frames every few frames to compare performance in low frame rate situations, and normalized pixel values into $[0, 1]$.
By applying RPCA with the recommended parameter value to these ground-truth videos, the pseudo ground-truth background and foreground components were generated.
In addition, through threshold processing on the pseudo ground-truth foreground components, we obtain the pseudo ground-truth foreground maps for calculating the area under the reciever operating characteristic curve (AUC) (described in Section~\ref{sssec:evaluation_metric}).
Fig.~\ref{fig:GT_images} shows the frames of the ground-truth videos and the pseudo ground-truth foreground maps.


\subsection{Experimental Setup}
\subsubsection{Noise Setup}
We consider the following cases of noise.

\textit{Case 1 (Gaussian noise):} The observed HS image is contaminated by white Gaussian noise with the standard deviation $\StanDevGauss = 0.1$.

\textit{Case 2 (Gaussian noise + salt-and-pepper noise):} The observed HS image is contaminated by white Gaussian noise with the standard deviation $\StanDevGauss = 0.1$ and salt-and-pepper noise with the rate $\RateSparse = 0.05$.

\textit{Case 3 (Gaussian noise + salt-and-pepper noise + stripe noise):} The observed HS image is contaminated by white Gaussian noise with the standard deviation $\StanDevGauss = 0.1$ and salt-and-pepper noise with the rate $\RateSparse = 0.05$. In addition, the observed HS image is corrupted by vertical stripe noise whose intensity is random in the range $[-0.2, 0.2]$. We only added time-invariant stipe noise because the stripe location of video data often does not change over time~\cite{IRI_TDFPN1}.

\subsubsection{Parameter Setup}

The parameters of existing methods and \Ourss were adjusted in the ranges shown in Tab.~\ref{tab:parameter_settings} to minimize the recovery error of objects of a foreground component $\bar{\CompFore}$, a background component $\bar{\CompBack}$, and a clean video $\bar{\VecVideoClean} = \bar{\CompFore} + \bar{\CompBack}$. Specifically, we determined them to take small values in measure calculated by the following equation:
\begin{equation}
	\label{eq:measure_param}
	\sum_{i = 1}^{\NumPixVer\NumPixHor\NumFrames} |\bar{\ElemVideoClean}_{i} - \mu_{\bar{\VecVideoClean}}| (\ElemVideoClean_{i} - \bar{\ElemVideoClean}_{i})^{2}
	+ |\bar{\ElemCompFore}_{i} - \mu_{\bar{\CompFore}}| (\ElemCompFore_{i} - \bar{\ElemCompFore}_{i})^{2}
	+ |\bar{\ElemCompBack}_{i} - \mu_{\bar{\CompBack}}| (\ElemCompBack_{i} - \bar{\ElemCompBack}_{i})^{2},
\end{equation}
where $\mu_{\bar{\VecVideoClean}}$, $\mu_{\bar{\CompFore}}$, and $\mu_{\bar{\CompBack}}$ are the mean values of $\bar{\VecVideoClean}$, $\bar{\CompFore}$, and $\bar{\CompBack}$, respectively.
The reason for weighting by the absolute values of the difference between pixel values and means, rather than the absolute values of pixel values, is to enable appropriate evaluation even for videos that have high-intensity pixel values in regions where any objects do not exist. Specifically, since MS videos (cell 1 and cell 2) have such regions, the measure with weights by the absolute values do not allow us to evaluate the object recovery performance. To avoid this, we used the measure in Eq.~\eqref{eq:measure_param}.
The parameter $\ParamOptSparCoef$ was determined by $0.05$ to achieve good performance, with reference to Ref.~\cite{CSRFBS_naganuma_2023_APSIPA}.
For the parameters $\ConsSpar$ and $\ConsFidel$ in~\eqref{prob:CSC_FBS}, we set them as $\ConsSpar = 0.5\RateSparse n_{1}n_{2}n_{3}$ and $\ConsFidel = 0.9\StanDevGauss\sqrt{(1 - \RateSparse)n_{1}n_{2}n_{3}}$, respectively.
The parameter $\ConsFore$, the number of filters $\NumDict$, and the size of filters were roughly determined from the intensity, the number, and the size of the foreground objects, as shown in Tab.~\ref{tab:parameter_settings_CSR}.
For Algorithm~\ref{algo:CSR_FBS}, we set the maximum iterations to $300$ and the stopping criterion to $\|\CompBack^{(n + 1)} - \CompBack^{(n)}\|_{2}/\|\CompBack^{(n)}\|_{2} \leq 10^{-5}$ and $\|\CompFore^{(n + 1)} - \CompFore^{(n)}\|_{2}/\|\CompFore^{(n)}\|_{2} \leq 10^{-5}$.
Moreover, we set the maximum iterations of Algorithms~\ref{algo:sub_FBSA_PDS} and~\ref{algo:sub_D_FISTA} to $500$ and $10$, respectively.
After obtaining the dictionaries, we finally estimated $\CompFore$ and $\CompBack$ using Algorithm 2 with $20000$ iterations.

\input{./fig_tex/result_image_cell2_g.tex}

\input{./fig_tex/result_image_cell1_gs.tex}

\input{./fig_tex/result_image_CAMEL_17_gsst.tex}

\subsubsection{Evaluation Metric}
\label{sssec:evaluation_metric}
For the quality measures, we used the mean of the peak-signal-to-noise ratio (MPSNR) [dB]: 
\begin{equation}
	\PSNR(\mathbf{q}, \bar{\mathbf{q}}) := \frac{1}{n_{3}} \sum_{k = 1}^{n_{3}} 10 \log_{10}\frac{n_{1}n_{2}}{\|\mathbf{q}_{n_{3}} - \bar{\mathbf{q}}_{n_{3}}\|_{2}^{2}},
\end{equation}
and the mean of structural similarity (MSSIM)~\cite{MSSIM}:
\begin{equation}
	\mathrm{MSSIM}(\mathbf{q}, \bar{\mathbf{q}}) := \frac{1}{n_{3}} \sum_{k = 1}^{n_{3}} \mathrm{SSIM}(\mathbf{q}_{k}, \bar{\mathbf{q}}_{k}), 
\end{equation}
for $\mathbf{q} = \CompFore$ and $\mathbf{q} = \CompBack$,
where $\mathbf{q}_{k}$ is the $k$-th frame of a video to be evaluated and $\bar{\mathbf{q}}_{k}$ is a reference frame to $\mathbf{q}_{k}$.
The higher the MPSNR and MSSIM values are, the better the estimation performance.
In addition, we adopted AUC~\cite{ACU_Chang_2021}, which is the area under reciever operating characteristic curve with the true positive rate on the vertical axis and the false positive rate plotted on the horizontal axis. The closer the AUC value is to $1$, the better the foreground estimation performance.

\subsection{Experimental Results and Discussion}
Tables~\ref{tab:PSNR_and_SSIM},~\ref{tab:PSNR_and_SSIM_gs}, and~\ref{tab:PSNR_and_SSIM_gsst} show the MPSNR, MSSIM, and AUC results in Case 1, Case 2, and Case 3, respectively. The best and second-best values are highlighted in bold and underlined, respectively. RPCA, GNNLSM, TVRPCA, and PRPCA produced worse results than the other methods in almost all cases. GNNLSM resulted in the best/second-best MPSNR $\CompFore$ and MSSIM $\CompFore$ values for some videos, but its ACU values are close to $0.5$. This indicates that it does not adequately estimate foreground components, although its performance appears to be good in  MPSNR $\CompFore$ and MSSIM $\CompFore$.
SRTC and SS-RTD performed better than the other existing methods. However, they do not adequately model multiple types of noise (particularly sparse and stripe noise), reducing the capability of FBS in Cases 2 and 3.
FactorDVP-T did not perform well overall. In particular, its performance is degraded for \textit{Bird}, \textit{Cell1}, and \textit{Cell2}, which contain small foreground objects.
In contrast, \Ourss achieved best/second-best results in almost all cases.

\input{./fig_tex/result_filter.tex}

\input{./table_tex/result_abr.tex}

Figures~\ref{fig:result_image_cell2_g},~\ref{fig:result_image_cell1_gs}, and~\ref{fig:result_image_CAMEL_17_gsst} depict the FBS results for \textit{Cell2} in Case 1, \textit{Cell1} in Case 2, and for \textit{CAMEL seq17} in Case 3, respectively.
Figure~\ref{fig:result_filter} visualize some bases $\Dict$ obtained in the FBS process of \Ours.
RPCA and GNNLSM only captured the sparsity of the foreground components and thus did not separate the foreground components from noise (Figs.~\ref{fig:result_image_cell2_g},~\ref{fig:result_image_cell1_gs}, and~\ref{fig:result_image_CAMEL_17_gsst} (c) and (d)). 
TVRPCA also failed to separate them (Figs.~\ref{fig:result_image_cell2_g},~\ref{fig:result_image_cell1_gs}, and~\ref{fig:result_image_CAMEL_17_gsst} (e)) because it does not explicitly model noise. 
PRPCA, SRTC, and SS-RTD reduced the effect of noise through the total variation regularization, but their estimated components remain noisy or oversmoothed (Figs.~\ref{fig:result_image_cell2_g},~\ref{fig:result_image_cell1_gs}, and~\ref{fig:result_image_CAMEL_17_gsst} (f), (g), and (h)). This is because they cannot capture the specific structures (e.g., edges and shapes) of temporally discontinuous foreground objects in low frame-rate videos. 
The results of FactorDVP-T were unstable, as it completely eliminated foreground components (Fig.~\ref{fig:result_image_cell2_g} (i)), generated shapes that do not exist in the ground-truth foreground component (Fig.~\ref{fig:result_image_cell1_gs} (i)), generated brighter/darker frames than the original frames (Figs.~\ref{fig:result_image_cell2_g} and~\ref{fig:result_image_cell1_gs} (i)), and distorted objects (Fig.~\ref{fig:result_image_CAMEL_17_gsst} (i)). This is because FactorDVP-T does not take into account general features such as sparsity and piecewise smoothness in FBS. 
In contrast, due to the CSR-based foreground modeling, \Ourss recovered the foreground objects (Figs.~\ref{fig:result_image_cell2_g},~\ref{fig:result_image_cell1_gs}, and~\ref{fig:result_image_CAMEL_17_gsst} (j) and (k)) by referring to the bases (Figs.~\ref{fig:result_filter}) simultaneously estimated in the FBS process. In addition, our model removed noise due to explicit noise modeling. These facts show the effectiveness of \Ours.

\subsection{Ablation Experiments}
%

To demonstrate the contribution of the CSR-based foreground modeling, we compared \Ourss performance with the performance when the CSR-based foreground modeling was removed.
Specifically, we remove the first term, the second term, and the fifth constraint from Eq.~\eqref{prob:CSC_FBS}.
The parameters were set to the same as in \Ours.

Table~\ref{tab:results_abl} shows the MPSNR, MSSIM, and AUC results in all cases.
The higher value of \Ourss without CSR or \Ourss with CSR is highlighted in bold.
\Ourss with CSR was superior to \Ourss without CSR particularly in MPSNR $\CompFore$, MSSIM $\CompFore$, and AUC.
This implies that CSR improves to capture the nature of foreground components.

\section{Conclusion}
\label{sec:conclusion}

We have proposed \Ours, which adaptively captures the specific structures of foreground objects by CSR and promotes the sparsity and piecewise smoothness of foreground components by certain functions. Thanks to these foreground modeling, \Ourss has adequately separated the foreground and background components from noisy and low frame-rate videos. In addition, we have formulated the FBS as a constrained multiconvex optimization problem and developed an ALM algorithm to solve the problem. The experimental results have demonstrated the superiority of our model over existing foreground models.

\bibliographystyle{IEEEtran}

\bibliography{./main_FBS_CSR_ver1.bbl}

\begin{IEEEbiography}[{\includegraphics[width=1in,height=1.25in,clip,keepaspectratio]{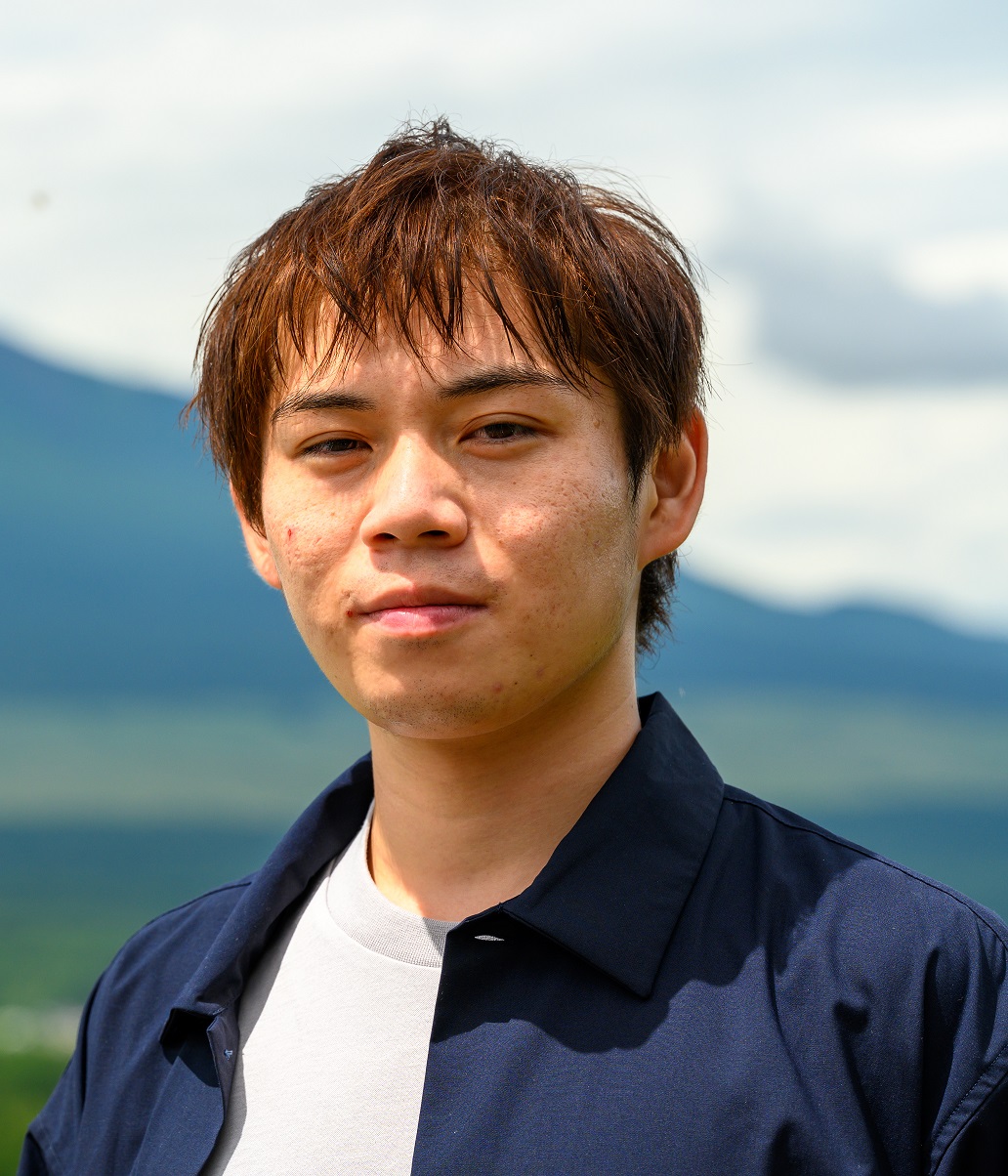}}]{Kazuki Naganuma}
	(S’21) received a B.E. degree in 2020 from the Kanagawa Institute of Technology and M.E. and Ph.D. degrees in Information and Computer Sciences 2022 and 2024 from the Tokyo Institute of Technology, respectively.
	
	He is an assistant professor at the Institute of Engineering of Tokyo University of Agriculture and Technology. 
	From April 2023 to March 2025, he was a Research Fellow (DC2) of the Japan Society for the Promotion of Science (JSPS). From October 2023 and April 2025 to present, He is a Research Fellow (PD) of JSPS and a Researcher of ACT-X of the Japan Science and Technology Corporation (JST), Tokyo, Japan.
	His current research interests are in signal and image processing and optimization theory.
	
	Dr. Naganuma received the Student Conference Paper Award from IEEE SPS Japan Chapter in 2023.
\end{IEEEbiography}

\begin{IEEEbiography}[{\includegraphics[width=1in,height=1.25in,clip,keepaspectratio]{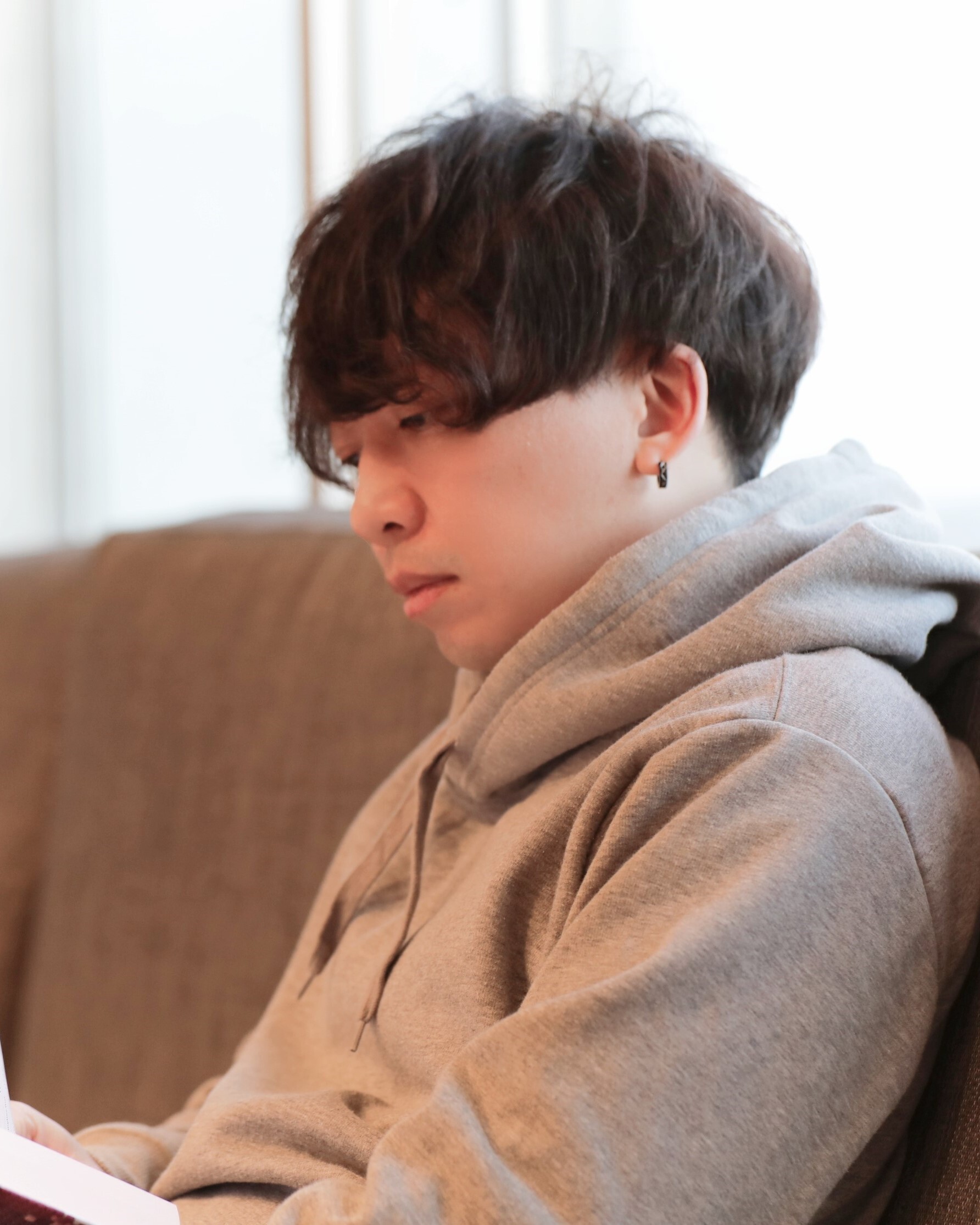}}]{Shunsuke Ono}
	(S’11–M’15–SM'23) received a B.E. degree in Computer Science in 2010 and M.E. and Ph.D. degrees in Communications and Computer Engineering in 2012 and 2014 from the Tokyo Institute of Technology, respectively.
	
	From 2012 to 2014, he was a Research Fellow (DC1) of the Japan Society for the Promotion of Science (JSPS). He was an Assistant, then an Associate Professor with Tokyo Institute of Technology (TokyoTech), Tokyo, Japan, from 2014 to 2024. From 2016 to 2020, he was a Researcher of Precursory Research for Embryonic Science and Technology (PRESTO), Japan Science and Technology Agency (JST), Tokyo, Japan. Currently, he is an Associate Professor with Institute of Science Tokyo (Science Tokyo), Tokyo, Japan. His research interests include signal processing, image analysis, optimization, remote sensing, and measurement informatics. He has served as an Associate Editor for IEEE TRANSACTIONS ON SIGNAL AND INFORMATION PROCESSING OVER NETWORKS (2019--2024).
	
	Dr. Ono was a recipient of the Young Researchers’ Award and the Excellent Paper Award from the IEICE in 2013 and 2014, respectively, the Outstanding Student Journal Paper Award and the Young Author Best Paper Award from the IEEE SPS Japan Chapter in 2014 and 2020, respectively, and the Best Paper Award in APSIPA ASC 2024. He also received the Funai Research Award in 2017, the Ando Incentive Prize in 2021, the MEXT Young Scientists’ Award in 2022, and the IEEE SPS Outstanding Editorial Board Member Award in 2023. 
\end{IEEEbiography}

\vfill

\end{document}

%% file: fig_tex/overall.tex
\begin{figure*}[!t]
	
	\begin{center}
		
		\begin{minipage}{0.99\hsize}
			\centerline{\includegraphics[width=\hsize]{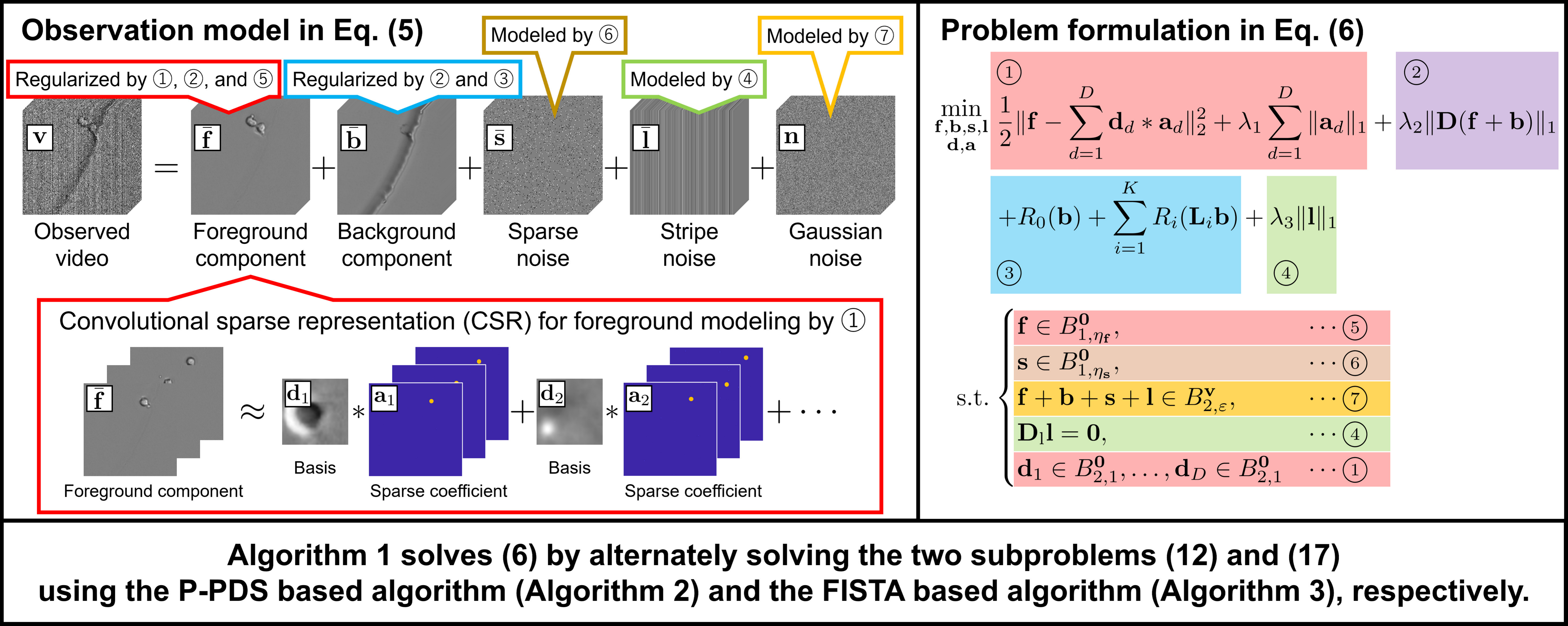}}
		\end{minipage}

	\end{center}

	\vspace{-1mm}
	
	\caption{Illustration of the proposed method, i.e., \Ours.}
	\label{fig:overall}
\end{figure*}

%% file: algo/algo_CSR_FBS.tex
%
%
%
%
%

\begin{algorithm}[t]
	\caption{Solver for the main FBS problem~\eqref{prob:CSC_FBS}}
	\label{algo:CSR_FBS}
	\begin{algorithmic}[1]
		\Require{$\CompFore^{(0)}, \CompBack^{(0)}, \CompSpar^{(0)}, \CompStri^{(0)}, \CoefSpar^{(0)}, \Dict^{(0)} $}
		\Ensure{$\CompFore^{(\IterOuter)}, \CompBack^{(\IterOuter)}, \Dict^{(\IterOuter)}$}
		\State Initialize $\IterOuter = 0, \CompFore^{(0)}, \CompBack^{(0)}, \CompSpar^{(0)}, \CompStri^{(0)}, \CoefSpar^{(0)}, \Dict^{(0)}$;
		\While {A stopping criterion are not satisfied}
		
		\vspace{1mm}
		
		\State $\CompFore^{(\IterOuter + 1)}, \CompBack^{(\IterOuter + 1)}, \CompSpar^{(\IterOuter + 1)}, \CompStri^{(\IterOuter + 1)}, \CoefSpar^{(\IterOuter + 1)}$
		
		$\leftarrow \argmin_{\CompFore, \CompBack, \CompSpar, \CompStri, \CoefSpar} \ALFuncMultiConvFixed{1}{(\IterOuter)}{\CompFore, \CompBack, \CompSpar, \CompStri, \CoefSpar} + \ALFuncConv{1}{\CompFore, \CompBack, \CompSpar, \CompStri, \CoefSpar}$ 
		
		i.e., solving Prob.~\eqref{prob:sub_FBSA_PDS} by Algorithm~\ref{algo:sub_FBSA_PDS}; 
		
		\vspace{1mm}
		
		\State $\Dict^{(\IterOuter + 1)} \leftarrow \argmin_{\Dict} \ALFuncMultiConvFixed{2}{(\IterOuter)}{\Dict} + \ALFuncConv{2}{\Dict}$ 
		
		i.e., solving Prob.~\eqref{prob:sub_D_FISTA} by Algorithm~\ref{algo:sub_D_FISTA}; 
		
		\vspace{1mm}
		
		\State $\IterOuter \leftarrow \IterOuter + 1$;
		\EndWhile
	\end{algorithmic}
\end{algorithm}

%% file: algo/algo_sub_FBSA_PDS_no_conj.tex
\begin{algorithm}[t]
	\caption{P-PDS solver for the subproblem~\eqref{prob:sub_FBSA_PDS}}
	\label{algo:sub_FBSA_PDS}
	\begin{algorithmic}[1]
		\Require{$\CompFore^{(\IterOuter)}, \CompBack^{(\IterOuter)}, \CompSpar^{(\IterOuter)}, \CompStri^{(\IterOuter)}, \CoefSpar^{(\IterOuter)}, \widetilde{\Dict} = \Dict^{(\IterOuter)}$}
		\Ensure{$\CompFore^{(\IterInner)}, \CompBack^{(\IterInner)}, \CompSpar^{(\IterInner)}, \CompStri^{(\IterInner)}, \CoefSpar^{(\IterInner)}$}
		\State Initialize $\IterInner = 0, \VarDual{1,1}, \ldots, \VarDual{1,\NumFuncBack}, \VarDual{2}, \VarDual{3}$;
		\State Set  $\CompFore^{(\IterInner)}, \CompBack^{(\IterInner)}, \CompSpar^{(\IterInner)}, \CompStri^{(\IterInner)}, \CoefSpar^{(\IterInner)} \leftarrow \CompFore^{(\IterOuter)}, \CompBack^{(\IterOuter)}, \CompSpar^{(\IterOuter)}, \CompStri^{(\IterOuter)}, \CoefSpar^{(\IterOuter)}$;
		\State Set stepsizes $\gamma_{\CompFore},\gamma_{\CompBack},\gamma_{\CompSpar},\gamma_{\CoefSpar}, \gamma_{\VarSymDual}$ as~\eqref{eq:stepsize_setting_P_PDS};
		\While {A stopping criterion is not satisfied}
		
		\State $\CompFore^{\prime} \leftarrow \CompFore^{(\IterInner)}  - \gamma_{\CompFore}( \VarDual{2}^{(\IterInner)} + \DiffSpa^{\top}\VarDual{3}^{(\IterInner)} + \VarDual{4}^{(\IterInner)})$;
		\State $\CompFore^{(\IterInner + 1)} \leftarrow \mathrm{prox}_{\gamma_{\CompFore}\iota_{\BallFore}}(\CompFore^{\prime})$ by the algorithm in~\cite{fast_l1_ball_projection};
		
		\State $\CompBack^{\prime} \leftarrow \CompBack^{(\IterInner)} - \gamma_{\CompBack} ( \sum_{\ValFuncBack = 1}^{\NumFuncBack} \LinOpBack{\NumFuncBack}^{\top}\VarDual{1, \NumFuncBack}^{(\IterInner)} + \DiffSpa^{\top}\VarDual{3}^{(\IterInner)} + \VarDual{4}^{(\IterInner)})$;
		\State $\CompBack^{(\IterInner + 1)} \leftarrow \prox_{\gamma_{\CompBack}\FuncBack{0}}(\CompBack^{\prime})$;
		
		\State $\CompSpar^{\prime} \leftarrow \CompSpar^{(\IterInner)} - \gamma_{\CompSpar}\VarDual{4}^{(\IterInner)}$;
		\State $\CompSpar^{(\IterInner + 1)} \leftarrow \prox_{\gamma_{\CompSpar} \iota_{\BallSpar}}(\CompSpar^{\prime})$ by the algorithm in~\cite{fast_l1_ball_projection};
		
		\State $\CompStri^{(\IterInner + 1)} \leftarrow \mathrm{prox}_{\ParamStri \| \cdot \|_{1}}(\CompStri^{(\IterInner)} - \gamma_{\CompStri}(\VarDual{4}^{(\IterInner)} + \DiffVerTemp^{\top}\VarDual{5}^{(\IterInner)}))$;
		
		\For{$\ValDict = 1, \cdots, \NumDict$}
		\State $\CoefSpar_{\ValDict}^{\prime} \leftarrow \CoefSpar_{\ValDict}^{(\IterInner)} - \gamma_{\CoefSpar_{\ValDict}} ( \ConvD{\widetilde{\Dict}_{\ValDict}}^{*}(-\VarDual{2}^{(\IterInner)}))$;
		\State $\CoefSpar_{\ValDict}^{(\IterInner + 1)} \leftarrow \mathrm{prox}_{\gamma_{\CoefSpar_{\ValDict}}\ParamOptSparCoef\|\cdot\|_{1}}(\CoefSpar_{\ValDict}^{\prime})$;
		\EndFor
		
		\State $\widetilde{\CompBack} \leftarrow 2\CompBack^{(\IterInner + 1)} - \CompBack^{(\IterInner)}$;
		\State $\widetilde{\CompFore} \leftarrow 2\CompFore^{(\IterInner + 1)} - \CompFore^{(\IterInner)}$;
		\State $\widetilde{\CompSpar} \leftarrow 2\CompSpar^{(\IterInner + 1)} - \CompSpar^{(\IterInner)}$;
		\State $\widetilde{\CompStri} \leftarrow 2\CompStri^{(\IterInner + 1)} - \CompStri^{(\IterInner)}$;
		\State $\widetilde{\CoefSpar} \leftarrow \sum_{\ValDict = 1}^{\NumDict}\ConvD{\widetilde{\Dict}_{\ValDict}}( 2\CoefSpar_{\ValDict}^{(\IterInner + 1)} - \CoefSpar_{\ValDict}^{(\IterInner)})$;
		
		\For{$\ValFuncBack = 1, \ldots, \NumFuncBack$}
		\State $\VarDual{1, \ValFuncBack}^{\prime} \leftarrow \VarDual{1, \ValFuncBack}^{(\IterInner)} + \gamma_{\VarSymDual}\LinOpBack{\ValFuncBack}\widetilde{\CompBack}$;
		\State $\VarDual{1, \ValFuncBack}^{(\IterInner + 1)} 
		\leftarrow 
		\VarDual{1, \ValFuncBack}^{\prime} - 
		\gamma_{\VarSymDual}\prox_{\frac{1}{\gamma_{\VarSymDual}} \FuncBack{\ValFuncBack}} 
		(\tfrac{1}{\gamma_{\VarSymDual}}\VarDual{1, \ValFuncBack}^{\prime})$;
		\EndFor
		
		
		\State $\VarDual{2}^{\prime} 
		\leftarrow 
		\VarDual{2}^{(\IterInner)} + \gamma_{\VarSymDual}(\widetilde{\CompFore} - \widetilde{\CoefSpar})$;
		\State $\VarDual{2}^{(\IterInner + 1)} 
		\leftarrow
		\VarDual{2}^{\prime} - \gamma_{\VarSymDual}\prox_{\frac{1}{2\gamma_{\VarSymDual}}\|\cdot\|_{2}^{2}}(\tfrac{1}{\gamma_{\VarSymDual}}\VarDual{2}^{\prime})$;
		
		\State $\VarDual{3}^{\prime} \leftarrow \VarDual{3}^{(\IterInner)} + \gamma_{\VarSymDual}\DiffSpa(\widetilde{\CompBack} + \widetilde{\CompFore})$;
		\State $\VarDual{3}^{(\IterInner + 1)} \leftarrow \VarDual{3}^{\prime} - \gamma_{\VarSymDual}\prox_{\frac{\ParamOptSparCoef}{\gamma_{\VarSymDual}}\|\cdot\|_{1}}(\tfrac{1}{\gamma_{\VarSymDual}}\VarDual{3}^{\prime})$;
		
		\State $\VarDual{4}^{\prime} \leftarrow \VarDual{4}^{(\IterInner)} + \gamma_{\VarSymDual}(\widetilde{\CompBack} + \widetilde{\CompFore} + \widetilde{\CompSpar} + \widetilde{\CompStri})$;
		\State $\VarDual{4}^{(\IterInner + 1)} \leftarrow \VarDual{4}^{\prime} - \gamma_{\VarSymDual}\prox_{\frac{1}{\gamma_{\VarSymDual}}\iota_{\BallFidel}}(\tfrac{1}{\gamma_{\VarSymDual}}\VarDual{4}^{\prime})$;
		
		\State $\VarDual{5}^{(\IterInner + 1)} \leftarrow \VarDual{5}^{(\IterInner)} + \gamma_{\VarSymDual} \DiffVerTemp \widetilde{\CompStri}$;
		
		\State $\IterInner \leftarrow \IterInner + 1$;
		\EndWhile
	\end{algorithmic}
\end{algorithm}

%% file: algo/algo_sub_D_FISTA.tex
\begin{algorithm}[t]
	\caption{FISTA solver for the subproblem~\eqref{prob:sub_D_FISTA}}
	\label{algo:sub_D_FISTA}
	\begin{algorithmic}[1]
		\Require{$\Dict^{(\IterOuter)}, \widetilde{\CompFore} = \CompFore^{(\IterOuter + 1)}, \widetilde{\CoefSpar} = \CoefSpar^{(\IterOuter + 1)}, \gamma$}
		\Ensure{$\Dict^{(\IterInner)}$}
		\State Initialize $\IterInner = 0, \StepsizeSubFISTA^{(0)} = 1, \VarDual{1}, \ldots, \VarDual{\NumDict}$;
		\While {A stopping criterion is not satisfied}
		
		\State $\StepsizeSubFISTA^{(\IterInner + 1)} \leftarrow \frac{1 + \sqrt{1 + 4 \StepsizeSubFISTA^{(\IterInner)} \StepsizeSubFISTA^{(\IterInner)} }}{2}$;
		
		\For{$\ValDict = 1, \ldots, \NumDict$}
		\State $\Dict_{\ValDict}^{\prime} \leftarrow \VarDual{\ValDict}^{(\IterInner)} - \gamma \ConvA{\widetilde{\CoefSpar}_{\ValDict}}^{*}(\sum_{\ValDict = 1}^{\NumDict}\ConvA{\widetilde{\CoefSpar}_{\ValDict}}(\Dict_{\ValDict}^{(\IterInner)}) - \widetilde{\CompFore})$;
		\State $\Dict_{\ValDict}^{(\IterInner)} \leftarrow \mathrm{prox}_{\gamma \iota_{\BallDict}}(\Dict_{\ValDict}^{\prime})$;
		\State $\VarDual{\ValDict}^{(\IterInner + 1)} \leftarrow \Dict_{\ValDict}^{(\IterInner + 1)} + \frac{\StepsizeSubFISTA^{(\IterInner)} - 1}{\StepsizeSubFISTA^{(\IterInner + 1)}} (\Dict_{\ValDict}^{(\IterInner + 1)} - \Dict_{\ValDict}^{(\IterInner)})$;
		\EndFor

		\State $\IterInner \leftarrow \IterInner + 1$;
		\EndWhile
	\end{algorithmic}
\end{algorithm}

%% file: fig_tex/GT_images.tex
%
	%
	%
	%
%
%
%

\begin{figure*}[!t]
	
	\begin{center}
		
		\begin{minipage}{0.10\hsize}
			\centerline{\includegraphics[width=\hsize]{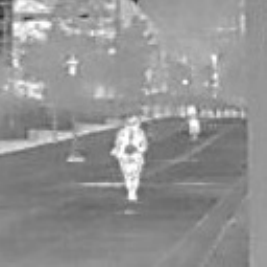}}
		\end{minipage}
		\begin{minipage}{0.10\hsize}
			\centerline{\includegraphics[width=\hsize]{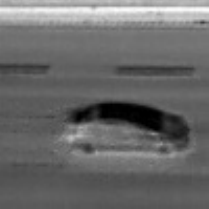}}
		\end{minipage}
		\begin{minipage}{0.10\hsize}
			\centerline{\includegraphics[width=\hsize]{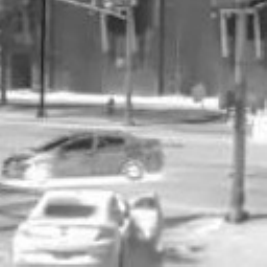}}
		\end{minipage}
		\begin{minipage}{0.10\hsize}
			\centerline{\includegraphics[width=\hsize]{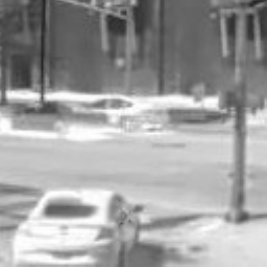}}
		\end{minipage}
		\begin{minipage}{0.10\hsize}
			\centerline{\includegraphics[width=\hsize]{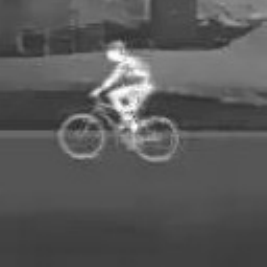}}
		\end{minipage}
		\begin{minipage}{0.10\hsize}
			\centerline{\includegraphics[width=\hsize]{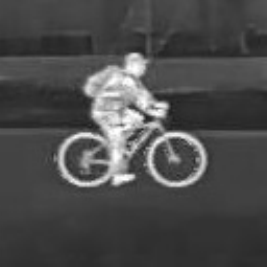}}
		\end{minipage}
		\begin{minipage}{0.10\hsize}
			\centerline{\includegraphics[width=\hsize]{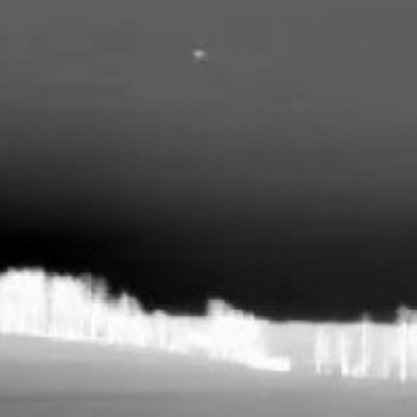}}
		\end{minipage}
		\begin{minipage}{0.10\hsize}
			\centerline{\includegraphics[width=\hsize]{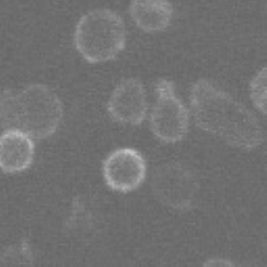}}
		\end{minipage}
		\begin{minipage}{0.10\hsize}
			\centerline{\includegraphics[width=\hsize]{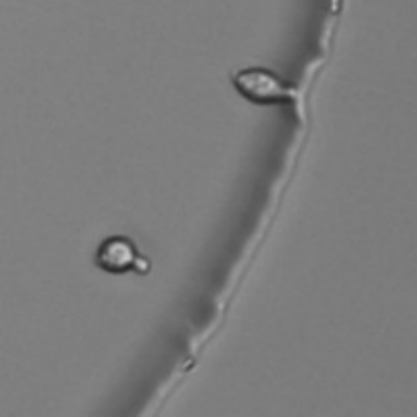}}
		\end{minipage}
		
		\vspace{1mm}
		
		\begin{minipage}{0.10\hsize}
			\centerline{\includegraphics[width=\hsize]{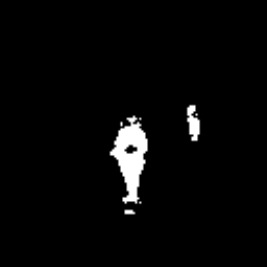}}
		\end{minipage}
		\begin{minipage}{0.10\hsize}
			\centerline{\includegraphics[width=\hsize]{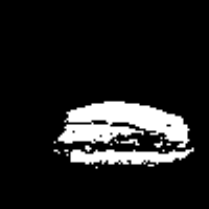}}
		\end{minipage}
		\begin{minipage}{0.10\hsize}
			\centerline{\includegraphics[width=\hsize]{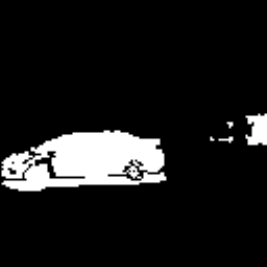}}
		\end{minipage}
		\begin{minipage}{0.10\hsize}
			\centerline{\includegraphics[width=\hsize]{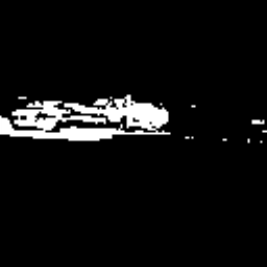}}
		\end{minipage}
		\begin{minipage}{0.10\hsize}
			\centerline{\includegraphics[width=\hsize]{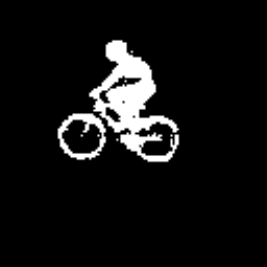}}
		\end{minipage}
		\begin{minipage}{0.10\hsize}
			\centerline{\includegraphics[width=\hsize]{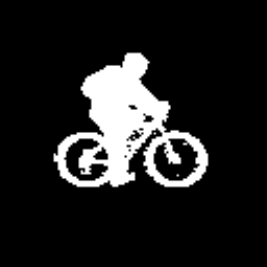}}
		\end{minipage}
		\begin{minipage}{0.10\hsize}
			\centerline{\includegraphics[width=\hsize]{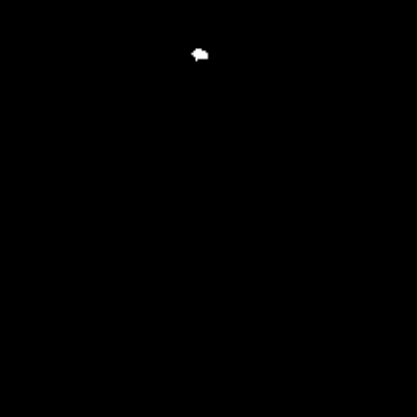}}
		\end{minipage}
		\begin{minipage}{0.10\hsize}
			\centerline{\includegraphics[width=\hsize]{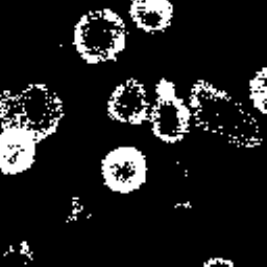}}
		\end{minipage}
		\begin{minipage}{0.10\hsize}
			\centerline{\includegraphics[width=\hsize]{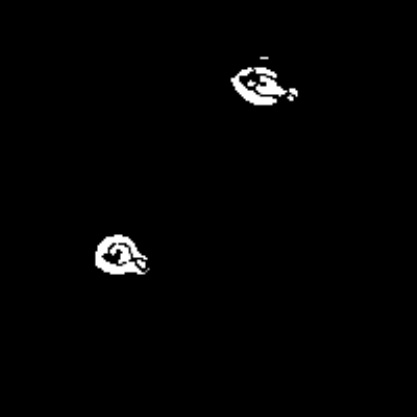}}
		\end{minipage}
		
		\vspace{1mm}
		
		\begin{minipage}{0.10\hsize}
			\centerline{\footnotesize{\textit{CAMEL seq8}}}
		\end{minipage}
		\begin{minipage}{0.10\hsize}
			\centerline{\footnotesize{\textit{CAMEL seq10}}}
		\end{minipage}
		\begin{minipage}{0.10\hsize}
			\centerline{\footnotesize{\textit{CAMEL seq17}}}
		\end{minipage}
		\begin{minipage}{0.10\hsize}
			\centerline{\footnotesize{\textit{CAMEL seq18}}}
		\end{minipage}
		\begin{minipage}{0.10\hsize}
			\centerline{\footnotesize{\textit{CAMEL seq20}}}
		\end{minipage}
		\begin{minipage}{0.10\hsize}
			\centerline{\footnotesize{\textit{CAMEL seq21}}}
		\end{minipage}
		\begin{minipage}{0.10\hsize}
			\centerline{\footnotesize{\textit{Bird}}}
		\end{minipage}
		\begin{minipage}{0.10\hsize}
			\centerline{\footnotesize{\textit{Cell1}}}
		\end{minipage}
		\begin{minipage}{0.10\hsize}
			\centerline{\footnotesize{\textit{Cell2}}}
		\end{minipage}

	\end{center}

		\vspace{-3mm}
	
	\caption{The top row shows the ground-truth video frames. The bottom row shows the pseudo ground-truth foreground maps. \textit{CAMEL seq8}, \textit{CAMEL seq10}, \textit{CAMEL seq17}, \textit{CAMEL seq18}, \textit{CAMEL seq20}, \textit{CAMEL seq21}, and \textit{Bird} are IR videos. \textit{Cell1} and \textit{Cell2} are MS videos.}
	\label{fig:GT_images}
	\vspace{-5mm}
\end{figure*}

%% file: table_tex/params.tex
\begin{table*}[t]
	\caption{Hyperparameter Settings in Each Method.}
	\vspace{-1mm}
	\label{tab:parameter_settings}
	\centering
	\begin{tabular}{ccc}
		\toprule
		Methods & Recommended parameters & Parameter search ranges \\
		\midrule
		\vspace{1mm}
		
		RPCA~\cite{RPCA} & 
		\begin{tabular}{c}  
			\vspace{1mm}
			$\lambda = \frac{1}{\sqrt{\max(n_{1}, n_{2})}}$
		\end{tabular} &
		\begin{tabular}{c} 
			$\lambda = \frac{\kappa}{\sqrt{\max(n_{1}, n_{2})}}$, 
			$\kappa \in \{ 10^{-2}, 10^{-1}, 10^{0}, 10^{1}, 10^{2} \}$ 
		\end{tabular} \\
		\vspace{2mm}
		
		GNNLSM~\cite{GNNLSM_YANG_2020} & 
		\begin{tabular}{c} 
			\vspace{1mm}
			$\tau = 300$, \\ 
			$\gamma = 0.001$ 
		\end{tabular} &
		\begin{tabular}{c} 
			\vspace{1mm}
			$\tau = 300 \kappa_{1}, \kappa_{1} \in \{ 10^{-1}, 10^{0}, 10^{1} \}$, \\ 
			$\gamma = 0.001 \kappa_{2}, \kappa_{2} \in \{ 10^{-1}, 10^{0}, 10^{1} \}$ 
		\end{tabular} \\
		\vspace{2mm}
		
		TVRPCA~\cite{TVRPCA} & 
		\begin{tabular}{c} 
			\vspace{1mm}
			$\lambda_{1} = \frac{0.4}{n_{1}n_{2}}$, \\ 
			\vspace{1mm}
			$\lambda_{2} = \frac{2}{n_{1}n_{2}}$, \\ 
			$\lambda_{3} = \frac{0.1}{n_{1}n_{2}}$ 
		\end{tabular} & 
		\begin{tabular}{c} 
			\vspace{1mm}
			$\lambda_{1} = \frac{0.4 \kappa_{1}}{n_{1}n_{2}}, \kappa_{1} \in \{ 10^{-2}, 10^{-1}, 10^{0}, 10^{1}, 10^{2} \}$, \\ 
			\vspace{1mm}
			$\lambda_{2} = \frac{2 \kappa_{2}}{n_{1}n_{2}}, \kappa_{2} \in \{ 10^{-2}, 10^{-1}, 10^{0}, 10^{1}, 10^{2} \}$, \\ 
			$\lambda_{3} = \frac{0.1 \kappa_{3}}{n_{1}n_{2}}, \kappa_{3} \in \{ 10^{-2}, 10^{-1}, 10^{0}, 10^{1}, 10^{2} \}$ 
		\end{tabular} \\
		\vspace{2mm}
		
		PRPCA~\cite{PRPCA} & 
		\begin{tabular}{c} 
			\vspace{1mm}
			$\lambda_{L} \in \{ 1, 2, 4, 6, 8, 10 \}$, \\ 
			\vspace{1mm}
			$\lambda_{S} = \frac{0.01}{\sqrt{n_{1}n_{2}}}$, \\ 
			$\lambda_{E} = \frac{0.001}{\sqrt{n_{1}n_{2}}}$ 
		\end{tabular} &
		\begin{tabular}{c} 
			\vspace{1mm}
			$\lambda_{L} \in \{ 1, 2, 4, 6, 8, 10 \}$, \\ 
			\vspace{1mm}
			$\lambda_{S} = \frac{0.01 \kappa_{1}}{\sqrt{n_{1}n_{2}}}, \kappa_{1} \in \{ 10^{-2}, 10^{-1}, 10^{0}, 10^{1}, 10^{2} \}$, \\ 
			$\lambda_{E} = \frac{0.001 \kappa_{2}}{\sqrt{n_{1}n_{2}}}, \kappa_{2} \in \{ 10^{-2}, 10^{-1}, 10^{0}, 10^{1}, 10^{2} \}$ 
		\end{tabular} \\
		\vspace{2mm}
		
		SRTC~\cite{SRTC_FBS_Shen_2022} & 
		\begin{tabular}{c} 
			$\lambda = 0.5$ 
		\end{tabular} &
		\begin{tabular}{c} 
			$\lambda = 0.5 \kappa,  \kappa \in \{ 10^{-2}, 10^{-1}, 10^{0}, 10^{1}, 10^{2} \}$
		\end{tabular} \\
		\vspace{2mm}
		
		SS-RTD~\cite{RTD_FBS_Shen_2023} & $\lambda \in \{0.2, 0.4, 0.6, 0.8, 1.0\}$ & $\lambda \in \{0.2, 0.4, 0.6, 0.8, 1.0\}$ \\
		\vspace{2mm}
		
		FactorDVP-T~\cite{FDVP_Miao_2024} & 
		\begin{tabular}{c} 
			\vspace{1mm}
			Learning Rate$ \in \{ 10^{-4}, 10^{-3}, 10^{-2} \}$ \\ 
			$r \in \{1, 3\}$ 
		\end{tabular} &
		\begin{tabular}{c} 
			\vspace{1mm}
			Learning Rate$ \in \{ 10^{-4}, 10^{-3}, 10^{-2} \}$ \\ 
			$r \in \{1, 3\}$ 
		\end{tabular} \\
		\vspace{2mm}
		
		\textbf{\Ourss (Ours)} & -- & 
		\begin{tabular}{c} 
			\vspace{1mm}
			$\ParamOursTVFB \in \{ 10^{-2}, 10^{-1}, 10^{0}, 10^{1}, 10^{2} \}$, \\ 
			$\ParamOursLRB \in \{ 10^{-2}, 10^{-1}, 10^{0}, 10^{1}, 10^{2} \}$ 
		\end{tabular} \\
		\vspace{2mm}
		
		\textbf{\Ourss (Ours)} & -- & 
		\begin{tabular}{c} 
			$\ParamOursTVFB \in \{ 10^{-2}, 10^{-1}, 10^{0}, 10^{1}, 10^{2} \}$ 
		\end{tabular} \\
		\bottomrule
	\end{tabular}
\end{table*}

%% file: table_tex/params_CSR.tex
\begin{table}[t]
	\caption{Parameter Settings Related to CSR.}
	\vspace{-1mm}
	\label{tab:parameter_settings_CSR}
	\centering
	\begin{tabular}{cccc}
		\toprule
		Datasets & $\ConsFore$ & $\NumDict$ & The size of $\Dict$ \\
		\midrule
		\vspace{1mm}
		
		\textit{CAMEL seq8} 
		& $9000$ & $2$ & $25 \times 25$
		\\
		\vspace{1mm}
		
		\textit{CAMEL seq10}
		& $6000$ & $6$ & $25 \times 25$
		\\
		\vspace{1mm}
		
		\textit{CAMEL seq17}
		& $12000$ & $12$ & $51 \times 51$
		\\
		\vspace{1mm}
		
		\textit{CAMEL seq18}
		& $15000$ & $12$ & $51 \times 51$
		\\
		\vspace{1mm}
		
		\textit{CAMEL seq20}
		& $9000$ & $3$ & $41 \times 41$
		\\
		\vspace{1mm}
		
		\textit{CAMEL seq21}
		& $9000$ & $2$ & $41 \times 41$
		\\
		\vspace{1mm}
		
		\textit{Bird}
		& $2000$ & $6$ & $11 \times 11$
		\\
		\vspace{1mm}
		
		\textit{Cell1}
		& $7000$ & $6$ & $31 \times 31$
		\\
		\vspace{1mm}
		
		\textit{Cell2}
		& $7000$ & $6$ & $31 \times 31$
		\\
		\bottomrule
	\end{tabular}
\end{table}

%% file: table_tex/result_PSNR_SSIM.tex
\begin{table*}[t]
	\begin{center}
		\caption{MPSNR, MSSIM, and AUC Values of the FBS Results in Case 1.}
		\label{tab:PSNR_and_SSIM}
					\vspace{-2mm}
		\scalebox{0.75}{
			\begin{tabular}{ccccccccccc}
				\toprule
				\multirow{2}{*}{Video} &  \multirow{2}{*}{Measure} & \multicolumn{9}{c}{Methods} \\ \addlinespace[-1pt]
				\cmidrule(lr){3-11} 
				& 
				& RPCA~\cite{RPCA} & GNNLSM~\cite{GNNLSM_YANG_2020} & TVRPCA~\cite{TVRPCA} & PRPCA~\cite{PRPCA} &  SRTC~\cite{SRTC_FBS_Shen_2022} & SS-RTD~\cite{RTD_FBS_Shen_2023} & FactorDVP-T~\cite{FDVP_Miao_2024} & \textbf{\Ourss (LR)} & \textbf{\Ourss (SC)} \\ 
				\midrule
				
				\multirow{5}{*}{\textit{CAMEL seq8}} 
				& MPSNR $\CompFore$ &   19.95 &   24.18 &   19.51 &   19.23 &   30.79 &   32.16 &   25.38 & \ValSecnd{  32.32} & \Valbest{  32.68} \\ 
				& MPSNR $\CompBack$ &   30.31 &   18.60 &   27.52 &   28.35 & \ValSecnd{  31.75} &   31.57 &   26.05 &   30.78 & \Valbest{  34.87} \\ 
				& MSSIM $\CompFore$ &   0.0789 & \Valbest{  0.7638} &   0.0647 &   0.0606 &   0.4584 &   0.2961 &   0.1568 & \ValSecnd{  0.6546} &   0.6314 \\ 
				& MSSIM $\CompBack$ &   0.7938 &   0.2482 &   0.8176 &   0.7892 &   0.8493 &   0.7960 &   0.9273 & \Valbest{  0.9443} & \ValSecnd{  0.9374} \\ 
				& AUC &   0.9040 &   0.5043 &   0.9207 &   0.5697 &   0.9703 &   0.9809 &   0.9799 & \ValSecnd{  0.9867} & \Valbest{  0.9921} \\ 
				\midrule 
				
				\multirow{5}{*}{\textit{CAMEL seq10}} 
				& MPSNR $\CompFore$ &   20.04 &   27.17 &   19.60 &   19.16 &   30.68 &   27.70 &   17.92 & \Valbest{  31.17} & \ValSecnd{  31.11} \\ 
				& MPSNR $\CompBack$ &   30.56 &   18.59 &   28.06 &   30.26 &   32.00 &   30.27 &   17.97 & \Valbest{  36.32} & \ValSecnd{  32.72} \\ 
				& MSSIM $\CompFore$ &   0.1031 & \ValSecnd{  0.5854} &   0.0876 &   0.0472 &   0.5772 &   0.3062 &   0.1111 & \Valbest{  0.6211} &   0.5763 \\ 
				& MSSIM $\CompBack$ &   0.7502 &   0.2011 &   0.7828 &   0.7480 &   0.8176 &   0.7582 &   0.8635 & \ValSecnd{  0.9408} & \Valbest{  0.9549} \\ 
				& AUC &   0.8067 &   0.5070 &   0.8062 &   0.5240 &   0.9441 &   0.9256 &   0.6300 & \Valbest{  0.9659} & \ValSecnd{  0.9626} \\ 
				\midrule 
				
				\multirow{5}{*}{\textit{CAMEL seq17}} 
				& MPSNR $\CompFore$ &   19.87 &   24.97 &   19.27 &   19.20 &   29.22 &   28.76 &   19.75 & \ValSecnd{  30.54} & \Valbest{  31.45} \\ 
				& MPSNR $\CompBack$ &   28.90 &   18.55 &   26.15 &   30.61 & \Valbest{  33.01} &   31.66 &   20.16 &   30.76 & \ValSecnd{  32.42} \\ 
				& MSSIM $\CompFore$ &   0.0980 & \ValSecnd{  0.6673} &   0.0757 &   0.0553 & \Valbest{  0.7637} &   0.4141 &   0.0935 &   0.6018 &   0.5708 \\ 
				& MSSIM $\CompBack$ &   0.8381 &   0.3718 &   0.8588 &   0.8436 &   0.8884 &   0.8548 & \ValSecnd{  0.9058} &   0.9023 & \Valbest{  0.9350} \\ 
				& AUC &   0.7610 &   0.5001 &   0.7607 &   0.5210 &   0.9415 &   0.9414 &   0.6865 & \ValSecnd{  0.9612} & \Valbest{  0.9711} \\ 
				\midrule 
				
				\multirow{5}{*}{\textit{CAMEL seq18}} 
				& MPSNR $\CompFore$ &   19.86 &   22.94 &   19.24 &   18.66 &   27.95 &   28.05 &   22.52 & \ValSecnd{  30.02} & \Valbest{  30.99} \\ 
				& MPSNR $\CompBack$ &   28.41 &   18.04 &   25.80 &   29.29 & \ValSecnd{  32.47} &   31.00 &   23.13 &   30.22 & \Valbest{  32.63} \\ 
				& MSSIM $\CompFore$ &   0.1117 & \ValSecnd{  0.6391} &   0.0890 &   0.0535 & \Valbest{  0.7253} &   0.4153 &   0.2699 &   0.5942 &   0.5660 \\ 
				& MSSIM $\CompBack$ &   0.8292 &   0.3564 &   0.8538 &   0.8376 &   0.8863 &   0.8516 &   0.8517 & \ValSecnd{  0.9000} & \Valbest{  0.9331} \\ 
				& AUC &   0.8035 &   0.5002 &   0.8004 &   0.5242 &   0.9584 &   0.9593 &   0.7956 & \ValSecnd{  0.9756} & \Valbest{  0.9861} \\ 
				\midrule 
				
				\multirow{5}{*}{\textit{CAMEL seq20}} 
				& MPSNR $\CompFore$ &   19.97 &   31.06 &   19.59 &   18.87 &   31.00 &   28.50 &   25.63 & \Valbest{  33.55} & \ValSecnd{  33.35} \\ 
				& MPSNR $\CompBack$ &   30.95 &   19.30 &   28.21 &   28.77 &   30.99 &   30.19 &   26.43 & \Valbest{  38.32} & \ValSecnd{  35.34} \\ 
				& MSSIM $\CompFore$ &   0.0774 & \Valbest{  0.8179} &   0.0638 &   0.0559 &   0.5041 &   0.2967 &   0.1465 & \ValSecnd{  0.6918} &   0.6519 \\ 
				& MSSIM $\CompBack$ &   0.7373 &   0.1589 &   0.7667 &   0.7187 &   0.7938 &   0.7330 &   0.9385 & \Valbest{  0.9674} & \ValSecnd{  0.9499} \\ 
				& AUC &   0.9303 &   0.6617 &   0.9412 &   0.5548 &   0.9744 &   0.9672 &   0.9873 & \Valbest{  0.9947} & \ValSecnd{  0.9935} \\ 
				\midrule 
				
				\multirow{5}{*}{\textit{CAMEL seq21}} 
				& MPSNR $\CompFore$ &   19.99 & \ValSecnd{  37.23} &   19.62 &   18.97 &   31.44 &   28.87 &   19.37 & \Valbest{  37.97} &   34.70 \\ 
				& MPSNR $\CompBack$ &   31.11 &   19.43 &   28.47 &   29.26 &   31.18 &   30.55 &   19.83 & \ValSecnd{  35.05} & \Valbest{  35.53} \\ 
				& MSSIM $\CompFore$ &   0.0778 & \Valbest{  0.8593} &   0.0641 &   0.0556 &   0.3771 &   0.2618 &   0.0590 & \ValSecnd{  0.8396} &   0.7084 \\ 
				& MSSIM $\CompBack$ &   0.7311 &   0.1413 &   0.7603 &   0.6987 &   0.7818 &   0.7227 &   0.8302 & \ValSecnd{  0.8701} & \Valbest{  0.9507} \\ 
				& AUC &   0.9542 &   0.7216 &   0.9614 &   0.5942 &   0.9885 &   0.9832 & \ValSecnd{  0.9921} & \Valbest{  0.9991} &   0.9919 \\ 
				\midrule 
				
				\multirow{5}{*}{\textit{Bird}} 
				& MPSNR $\CompFore$ &   19.86 & \Valbest{  49.42} &   19.49 &   20.15 & \ValSecnd{  49.16} &   48.97 &    7.53 &   41.91 &   39.85 \\ 
				& MPSNR $\CompBack$ &   29.98 &   19.99 &   27.36 &   34.72 &   34.72 &   32.92 &    7.72 & \ValSecnd{  37.52} & \Valbest{  39.83} \\ 
				& MSSIM $\CompFore$ &   0.0311 & \Valbest{  0.9881} &   0.0213 &   0.0439 & \ValSecnd{  0.9839} &   0.9762 &   0.0046 &   0.9238 &   0.8969 \\ 
				& MSSIM $\CompBack$ &   0.7220 &   0.1496 &   0.7444 &   0.7829 &   0.7827 &   0.7081 &   0.2707 & \ValSecnd{  0.9258} & \Valbest{  0.9396} \\ 
				& AUC &   0.7505 &   0.5000 &   0.7617 &   0.6808 &   0.8336 & \ValSecnd{  0.8467} &   0.5986 & \Valbest{  0.9047} &   0.7140 \\ 
				\midrule 
				
				\multirow{5}{*}{\textit{Cell1}} 
				& MPSNR $\CompFore$ &   29.76 &   29.76 &   30.56 &   20.19 &   31.35 &   28.87 &    8.92 & \Valbest{  33.70} & \ValSecnd{  31.87} \\ 
				& MPSNR $\CompBack$ &   19.42 &   19.43 &   20.12 &   30.53 &   31.83 &   30.73 &    8.93 & \ValSecnd{  34.59} & \Valbest{  36.21} \\ 
				& MSSIM $\CompFore$ & \Valbest{  0.7642} & \Valbest{  0.7642} &   0.5536 &   0.0541 & \ValSecnd{  0.6970} &   0.3215 &   0.0253 &   0.6904 &   0.5652 \\ 
				& MSSIM $\CompBack$ &   0.0871 &   0.0871 &   0.0968 &   0.6456 &   0.7264 &   0.6553 &   0.3607 & \ValSecnd{  0.8460} & \Valbest{  0.9864} \\ 
				& AUC &   0.5000 &   0.5001 &   0.8456 &   0.5074 &   0.7193 &   0.6817 & \Valbest{  0.9300} & \ValSecnd{  0.9288} &   0.8043 \\ 
				\midrule 
				
				\multirow{5}{*}{\textit{Cell2}} 
				& MPSNR $\CompFore$ &   19.89 &   35.06 &   19.42 &   20.46 & \Valbest{  37.48} &   31.34 &    7.17 & \ValSecnd{  36.29} &   34.51 \\ 
				& MPSNR $\CompBack$ &   30.25 &   34.43 &   26.86 &   32.97 &   34.54 &   32.93 &    7.17 & \ValSecnd{  40.73} & \Valbest{  43.01} \\ 
				& MSSIM $\CompFore$ &   0.0388 & \ValSecnd{  0.9318} &   0.0198 &   0.0618 & \Valbest{  0.9372} &   0.4197 &   0.0036 &   0.7819 &   0.7290 \\ 
				& MSSIM $\CompBack$ &   0.6794 &   0.7554 &   0.7143 &   0.6858 &   0.7561 &   0.6871 &   0.1931 & \ValSecnd{  0.9692} & \Valbest{  0.9804} \\ 
				& AUC &   0.7605 &   0.5000 &   0.7384 &   0.5175 &   0.9243 &   0.9317 &   0.1451 & \Valbest{  0.9787} & \ValSecnd{  0.9359} \\

				\bottomrule
			\end{tabular}
		}
	\end{center}
	\vspace{-3mm}
\end{table*}

%% file: table_tex/result_PSNR_SSIM_gs.tex
\begin{table*}[t]
	\begin{center}
		\caption{MPSNR, MSSIM, and AUC Values of the FBS Results in Case 2.}
		\label{tab:PSNR_and_SSIM_gs}
		\vspace{-2mm}
		\scalebox{0.75}{
			\begin{tabular}{ccccccccccc}
				\toprule
				\multirow{2}{*}{Video} &  \multirow{2}{*}{Measure} & \multicolumn{9}{c}{Methods} \\ \addlinespace[-1pt]
				\cmidrule(lr){3-11} 
				& 
				& RPCA~\cite{RPCA} & GNNLSM~\cite{GNNLSM_YANG_2020} & TVRPCA~\cite{TVRPCA} & PRPCA~\cite{PRPCA} &  SRTC~\cite{SRTC_FBS_Shen_2022} & SS-RTD~\cite{RTD_FBS_Shen_2023} & FactorDVP-T~\cite{FDVP_Miao_2024} & \textbf{\Ourss (LR)} & \textbf{\Ourss (SC)} \\ 
				\midrule
				
				\multirow{5}{*}{\textit{CAMEL seq8}} 
				& MPSNR $\CompFore$ &   16.44 &   24.18 &   16.24 &   16.12 &   27.45 & \ValSecnd{  31.94} &   20.24 &   31.74 & \Valbest{  32.51} \\ 
				& MPSNR $\CompBack$ &   29.85 &   15.80 &   27.03 &   28.32 &   29.82 & \ValSecnd{  31.18} &   19.99 &   31.10 & \Valbest{  32.97} \\ 
				& MSSIM $\CompFore$ &   0.0482 & \ValSecnd{  0.7639} &   0.0437 &   0.0256 &   0.4597 &   0.2961 &   0.0845 & \Valbest{  0.7842} &   0.6912 \\ 
				& MSSIM $\CompBack$ &   0.7755 &   0.1607 &   0.8055 &   0.7740 &   0.7533 &   0.7788 &   0.8533 & \ValSecnd{  0.8603} & \Valbest{  0.9030} \\ 
				& AUC &   0.8732 &   0.5045 &   0.8948 &   0.5579 &   0.9711 &   0.9787 &   0.9715 & \Valbest{  0.9972} & \ValSecnd{  0.9921} \\ 
				\midrule 
				
				\multirow{5}{*}{\textit{CAMEL seq10}} 
				& MPSNR $\CompFore$ &   16.22 &   19.12 &   15.95 &   15.86 &   22.84 &   27.30 &   21.09 & \Valbest{  31.81} & \ValSecnd{  31.52} \\ 
				& MPSNR $\CompBack$ &   30.18 &   17.54 &   27.47 &   30.03 &   30.19 &   29.97 &   21.99 & \Valbest{  32.96} & \ValSecnd{  32.24} \\ 
				& MSSIM $\CompFore$ &   0.0597 &   0.2146 &   0.0520 &   0.0213 &   0.1875 &   0.2934 &   0.1482 & \Valbest{  0.6470} & \ValSecnd{  0.6116} \\ 
				& MSSIM $\CompBack$ &   0.7273 &   0.1609 &   0.7630 &   0.7256 &   0.7470 &   0.7338 &   0.7315 & \ValSecnd{  0.8964} & \Valbest{  0.9396} \\ 
				& AUC &   0.7805 &   0.5078 &   0.7776 &   0.5139 &   0.9049 &   0.9111 &   0.6578 & \Valbest{  0.9783} & \ValSecnd{  0.9694} \\ 
				\midrule 
				
				\multirow{5}{*}{\textit{CAMEL seq17}} 
				& MPSNR $\CompFore$ &   16.06 &   24.97 &   15.84 &   15.73 &   25.11 &   28.30 &   15.77 & \ValSecnd{  30.11} & \Valbest{  31.13} \\ 
				& MPSNR $\CompBack$ &   28.48 &   15.40 &   25.80 &   30.17 & \ValSecnd{  30.83} & \Valbest{  31.17} &   15.42 &   28.12 &   30.42 \\ 
				& MSSIM $\CompFore$ &   0.0595 & \Valbest{  0.6672} &   0.0495 &   0.0232 &   0.4386 &   0.3955 &   0.0447 & \ValSecnd{  0.6171} &   0.5531 \\ 
				& MSSIM $\CompBack$ &   0.8250 &   0.2563 &   0.8471 &   0.8304 &   0.8299 &   0.8405 &   0.8022 & \ValSecnd{  0.8526} & \Valbest{  0.8974} \\ 
				& AUC &   0.7370 &   0.5000 &   0.7345 &   0.5162 &   0.9279 &   0.9290 &   0.5488 & \ValSecnd{  0.9573} & \Valbest{  0.9694} \\ 
				\midrule 
				
				\multirow{5}{*}{\textit{CAMEL seq18}} 
				& MPSNR $\CompFore$ &   16.10 &   22.94 &   15.83 &   15.50 &   24.71 &   27.63 &   17.42 & \ValSecnd{  29.39} & \Valbest{  30.48} \\ 
				& MPSNR $\CompBack$ &   27.99 &   15.19 &   25.42 &   28.89 &   30.38 & \Valbest{  30.72} &   17.10 &   28.29 & \ValSecnd{  30.66} \\ 
				& MSSIM $\CompFore$ &   0.0699 & \Valbest{  0.6391} &   0.0599 &   0.0224 &   0.4353 &   0.3919 &   0.0766 &   0.5521 & \ValSecnd{  0.6143} \\ 
				& MSSIM $\CompBack$ &   0.8183 &   0.2468 &   0.8412 &   0.8253 &   0.8260 &   0.8419 & \ValSecnd{  0.8437} &   0.8288 & \Valbest{  0.9004} \\ 
				& AUC &   0.7766 &   0.5002 &   0.7706 &   0.5219 &   0.9335 &   0.9525 &   0.6261 & \ValSecnd{  0.9729} & \Valbest{  0.9836} \\ 
				\midrule 
				
				\multirow{5}{*}{\textit{CAMEL seq20}} 
				& MPSNR $\CompFore$ &   16.19 &   18.59 &   15.95 &   15.75 &   22.34 &   27.93 &   28.63 & \Valbest{  33.51} & \ValSecnd{  33.25} \\ 
				& MPSNR $\CompBack$ &   30.46 &   18.57 &   27.56 &   28.35 &   29.03 &   29.86 &   32.17 & \Valbest{  34.69} & \ValSecnd{  33.99} \\ 
				& MSSIM $\CompFore$ &   0.0491 &   0.2640 &   0.0427 &   0.0264 &   0.1144 &   0.2840 &   0.2481 & \Valbest{  0.7701} & \ValSecnd{  0.6491} \\ 
				& MSSIM $\CompBack$ &   0.7189 &   0.1394 &   0.7514 &   0.7005 &   0.7375 &   0.7128 &   0.8703 & \ValSecnd{  0.9408} & \Valbest{  0.9479} \\ 
				& AUC &   0.9058 &   0.6465 &   0.9170 &   0.5552 &   0.9375 &   0.9612 &   0.9903 & \Valbest{  0.9981} & \ValSecnd{  0.9929} \\ 
				\midrule 
				
				\multirow{5}{*}{\textit{CAMEL seq21}} 
				& MPSNR $\CompFore$ &   16.03 &   18.06 &   15.77 &   15.59 &   21.27 &   28.35 &   33.20 & \Valbest{  35.56} & \ValSecnd{  34.46} \\ 
				& MPSNR $\CompBack$ &   30.56 &   18.75 &   27.72 &   28.71 &   28.70 &   30.17 & \Valbest{  34.41} &   30.80 & \ValSecnd{  33.00} \\ 
				& MSSIM $\CompFore$ &   0.0512 &   0.2612 &   0.0444 &   0.0268 &   0.0978 &   0.2492 &   0.5148 & \Valbest{  0.7670} & \ValSecnd{  0.6755} \\ 
				& MSSIM $\CompBack$ &   0.7085 &   0.1276 &   0.7426 &   0.6764 &   0.7280 &   0.6978 &   0.8302 & \Valbest{  0.9574} & \ValSecnd{  0.9513} \\ 
				& AUC &   0.9308 &   0.6616 &   0.9379 &   0.5874 &   0.9577 &   0.9795 & \ValSecnd{  0.9954} & \Valbest{  0.9965} &   0.9830 \\ 
				\midrule 
				
				\multirow{5}{*}{\textit{Bird}} 
				& MPSNR $\CompFore$ &   15.85 &   19.91 &   15.67 &   15.93 & \Valbest{  51.09} & \ValSecnd{  48.80} &   11.08 &   48.44 &   43.47 \\ 
				& MPSNR $\CompBack$ &   29.44 &   18.06 &   26.75 &   32.41 &   29.76 &   32.46 &   11.03 & \ValSecnd{  38.09} & \Valbest{  41.77} \\ 
				& MSSIM $\CompFore$ &   0.0155 &   0.4576 &   0.0107 &   0.0203 & \Valbest{  0.9892} & \ValSecnd{  0.9759} &   0.0123 & \ValSecnd{  0.9759} &   0.9506 \\ 
				& MSSIM $\CompBack$ &   0.6999 &   0.1200 &   0.7237 &   0.6854 &   0.5952 &   0.6867 &   0.6999 & \ValSecnd{  0.9675} & \Valbest{  0.9678} \\ 
				& AUC &   0.7439 &   0.5036 &   0.7560 &   0.7187 &   0.4393 & \ValSecnd{  0.9212} &   0.5951 & \Valbest{  0.9835} &   0.7703 \\ 
				\midrule 
				
				\multirow{5}{*}{\textit{Cell1}} 
				& MPSNR $\CompFore$ &   29.76 &   29.76 &   30.19 &   16.79 &   26.72 &   28.63 &   12.04 & \Valbest{  33.36} & \ValSecnd{  32.09} \\ 
				& MPSNR $\CompBack$ &   16.30 &   16.30 &   16.35 &   30.24 &   30.29 &   30.47 &   12.18 & \Valbest{  38.85} & \ValSecnd{  36.86} \\ 
				& MSSIM $\CompFore$ & \ValSecnd{  0.7642} & \ValSecnd{  0.7642} & \Valbest{  0.7676} &   0.0248 &   0.3085 &   0.3088 &   0.0318 &   0.6726 &   0.5809 \\ 
				& MSSIM $\CompBack$ &   0.0486 &   0.0486 &   0.0488 &   0.6202 &   0.6230 &   0.6317 &   0.7034 & \ValSecnd{  0.9679} & \Valbest{  0.9915} \\ 
				& AUC &   0.5000 &   0.5000 &   0.7719 &   0.5073 &   0.6626 &   0.6698 & \ValSecnd{  0.8923} & \Valbest{  0.9228} &   0.8274 \\ 
				\midrule 
				
				\multirow{5}{*}{\textit{Cell2}} 
				& MPSNR $\CompFore$ &   16.54 &   33.48 &   16.28 &   16.91 &   29.69 &   30.93 &    6.68 & \ValSecnd{  37.90} & \Valbest{  38.23} \\ 
				& MPSNR $\CompBack$ &   29.75 &   16.54 &   26.39 &   32.35 &   32.01 &   32.42 &    6.87 & \ValSecnd{  39.12} & \Valbest{  41.46} \\ 
				& MSSIM $\CompFore$ &   0.0203 & \Valbest{  0.9232} &   0.0118 &   0.0273 &   0.5196 &   0.3977 &   0.0036 & \ValSecnd{  0.8217} &   0.8179 \\ 
				& MSSIM $\CompBack$ &   0.6570 &   0.0618 &   0.6971 &   0.6553 &   0.6364 &   0.6615 &   0.1348 & \Valbest{  0.9782} & \ValSecnd{  0.9779} \\ 
				& AUC &   0.7332 &   0.5019 &   0.7109 &   0.5125 &   0.9069 &   0.9235 &   0.1345 & \ValSecnd{  0.9844} & \Valbest{  0.9856} \\

				\bottomrule
			\end{tabular}
		}
	\end{center}
	\vspace{-3mm}
\end{table*}

%% file: table_tex/result_PSNR_SSIM_gsst.tex
\begin{table*}[t]
	\begin{center}
		\caption{MPSNR, MSSIM, and AUC Values of the FBS Results in Case 3.}
		\label{tab:PSNR_and_SSIM_gsst}
		\vspace{-2mm}
		\scalebox{0.75}{
			\begin{tabular}{ccccccccccc}
				\toprule
				\multirow{2}{*}{Video} &  \multirow{2}{*}{Measure} & \multicolumn{9}{c}{Methods} \\ \addlinespace[-1pt]
				\cmidrule(lr){3-11} 
				& 
				& RPCA~\cite{RPCA} & GNNLSM~\cite{GNNLSM_YANG_2020} & TVRPCA~\cite{TVRPCA} & PRPCA~\cite{PRPCA} & SRTC~\cite{SRTC_FBS_Shen_2022} & SS-RTD~\cite{RTD_FBS_Shen_2023} & FactorDVP-T~\cite{FDVP_Miao_2024} & \textbf{\Ourss (LR)} & \textbf{\Ourss (SC)} \\ 
				\midrule
				
				\multirow{5}{*}{\textit{CAMEL seq8}} 
				& MPSNR $\CompFore$ &   16.44 &   24.18 &   16.24 &   16.13 &   27.33 & \ValSecnd{  31.95} &   27.93 &   31.64 & \Valbest{  32.08} \\ 
				& MPSNR $\CompBack$ &   24.26 &   15.34 &   23.60 &   23.29 &   23.83 &   23.92 &   23.74 & \Valbest{  27.98} & \ValSecnd{  27.92} \\ 
				& MSSIM $\CompFore$ &   0.0484 & \Valbest{  0.7638} &   0.0440 &   0.0256 &   0.4528 &   0.2952 &   0.3197 & \ValSecnd{  0.7557} &   0.6288 \\ 
				& MSSIM $\CompBack$ &   0.4987 &   0.1472 &   0.5128 &   0.4865 &   0.4844 &   0.4913 &   0.4991 & \ValSecnd{  0.7263} & \Valbest{  0.7351} \\ 
				& AUC &   0.8739 &   0.5044 &   0.8957 &   0.5572 &   0.9695 &   0.9808 &   0.9783 & \Valbest{  0.9969} & \ValSecnd{  0.9918} \\ 
				\midrule 
				
				\multirow{5}{*}{\textit{CAMEL seq10}} 
				& MPSNR $\CompFore$ &   16.15 &   19.12 &   15.89 &   15.78 &   22.63 &   27.17 &   25.46 & \Valbest{  30.84} & \ValSecnd{  29.49} \\ 
				& MPSNR $\CompBack$ &   22.65 &   16.36 &   22.24 &   22.17 &   22.27 &   22.11 &   27.04 & \Valbest{  29.02} & \ValSecnd{  28.79} \\ 
				& MSSIM $\CompFore$ &   0.0604 &   0.2233 &   0.0534 &   0.0207 &   0.1837 &   0.2714 &   0.4240 & \Valbest{  0.6117} & \ValSecnd{  0.5442} \\ 
				& MSSIM $\CompBack$ &   0.3809 &   0.1337 &   0.3955 &   0.3671 &   0.3739 &   0.3669 &   0.7371 & \ValSecnd{  0.7878} & \Valbest{  0.8213} \\ 
				& AUC &   0.7789 &   0.5081 &   0.7769 &   0.5134 &   0.9003 &   0.9092 &   0.4695 & \Valbest{  0.9682} & \ValSecnd{  0.9341} \\ 
				\midrule 
				
				\multirow{5}{*}{\textit{CAMEL seq17}} 
				& MPSNR $\CompFore$ &   16.03 &   24.97 &   15.80 &   15.69 &   25.01 &   28.31 &   21.44 & \ValSecnd{  29.82} & \Valbest{  30.99} \\ 
				& MPSNR $\CompBack$ &   24.35 &   15.02 &   23.35 &   24.50 &   24.77 &   24.73 &   20.07 & \Valbest{  27.52} & \ValSecnd{  27.42} \\ 
				& MSSIM $\CompFore$ &   0.0589 & \Valbest{  0.6670} &   0.0489 &   0.0228 &   0.4305 &   0.3958 &   0.1228 & \ValSecnd{  0.5816} &   0.5529 \\ 
				& MSSIM $\CompBack$ &   0.6337 &   0.2373 &   0.6452 &   0.6275 &   0.6314 &   0.6342 &   0.6286 & \Valbest{  0.8205} & \ValSecnd{  0.8149} \\ 
				& AUC &   0.7411 &   0.5001 &   0.7378 &   0.5209 &   0.9223 &   0.9313 &   0.7292 & \ValSecnd{  0.9451} & \Valbest{  0.9670} \\ 
				\midrule 
				
				\multirow{5}{*}{\textit{CAMEL seq18}} 
				& MPSNR $\CompFore$ &   16.04 &   22.94 &   15.78 &   15.43 &   24.56 &   27.60 &   19.71 & \ValSecnd{  29.05} & \Valbest{  30.46} \\ 
				& MPSNR $\CompBack$ &   23.46 &   14.75 &   22.45 &   23.58 &   24.05 &   24.02 &   18.74 & \ValSecnd{  26.98} & \Valbest{  27.10} \\ 
				& MSSIM $\CompFore$ &   0.0691 & \Valbest{  0.6391} &   0.0590 &   0.0226 &   0.4271 &   0.3866 &   0.1080 &   0.5317 & \ValSecnd{  0.5667} \\ 
				& MSSIM $\CompBack$ &   0.6077 &   0.2252 &   0.6195 &   0.5991 &   0.6066 &   0.6093 &   0.5663 & \Valbest{  0.7855} & \ValSecnd{  0.7769} \\ 
				& AUC &   0.7778 &   0.5002 &   0.7730 &   0.5261 &   0.9354 &   0.9523 &   0.7053 & \ValSecnd{  0.9689} & \Valbest{  0.9828} \\ 
				\midrule 
				
				\multirow{5}{*}{\textit{CAMEL seq20}} 
				& MPSNR $\CompFore$ &   16.16 &   18.52 &   15.91 &   15.71 &   22.18 &   27.84 &   26.73 & \Valbest{  32.93} & \ValSecnd{  31.81} \\ 
				& MPSNR $\CompBack$ &   24.38 &   17.75 &   23.66 &   23.31 &   23.58 &   23.70 &   24.19 & \ValSecnd{  30.27} & \Valbest{  30.72} \\ 
				& MSSIM $\CompFore$ &   0.0497 &   0.2581 &   0.0433 &   0.0262 &   0.1087 &   0.2727 &   0.2391 & \Valbest{  0.6917} & \ValSecnd{  0.6202} \\ 
				& MSSIM $\CompBack$ &   0.3901 &   0.1214 &   0.4054 &   0.3637 &   0.3752 &   0.3690 &   0.4032 & \ValSecnd{  0.8077} & \Valbest{  0.8802} \\ 
				& AUC &   0.9037 &   0.6500 &   0.9157 &   0.5552 &   0.9333 &   0.9596 &   0.9652 & \Valbest{  0.9892} & \ValSecnd{  0.9827} \\ 
				\midrule 
				
				\multirow{5}{*}{\textit{CAMEL seq21}} 
				& MPSNR $\CompFore$ &   16.00 &   18.14 &   15.76 &   15.57 &   21.36 &   28.06 &   30.59 & \Valbest{  37.16} & \ValSecnd{  35.54} \\ 
				& MPSNR $\CompBack$ &   24.47 &   17.53 &   24.20 &   22.67 &   22.64 &   22.91 &   23.65 & \ValSecnd{  29.75} & \Valbest{  31.89} \\ 
				& MSSIM $\CompFore$ &   0.0506 &   0.2672 &   0.0444 &   0.0259 &   0.0972 &   0.2179 &   0.3614 & \Valbest{  0.8226} & \ValSecnd{  0.7141} \\ 
				& MSSIM $\CompBack$ &   0.3431 &   0.1034 &   0.3603 &   0.3080 &   0.3220 &   0.3137 &   0.3378 & \ValSecnd{  0.6790} & \Valbest{  0.9347} \\ 
				& AUC &   0.9290 &   0.6458 &   0.9362 &   0.5861 &   0.9552 &   0.9723 &   0.9790 & \Valbest{  0.9989} & \ValSecnd{  0.9943} \\ 
				\midrule 
				
				\multirow{5}{*}{\textit{Bird}} 
				& MPSNR $\CompFore$ &   15.81 &   19.79 &   15.64 &   15.89 & \Valbest{  51.09} & \ValSecnd{  48.57} &    7.40 &   40.81 &   42.75 \\ 
				& MPSNR $\CompBack$ &   23.41 &   17.16 &   22.59 &   23.79 &   23.68 &   23.79 &    7.30 & \ValSecnd{  29.58} & \Valbest{  29.97} \\ 
				& MSSIM $\CompFore$ &   0.0155 &   0.4497 &   0.0107 &   0.0203 & \Valbest{  0.9892} & \ValSecnd{  0.9745} &   0.0034 &   0.9304 &   0.9429 \\ 
				& MSSIM $\CompBack$ &   0.3597 &   0.1024 &   0.3696 &   0.3442 &   0.3314 &   0.3447 &   0.1085 & \ValSecnd{  0.7676} & \Valbest{  0.7894} \\ 
				& AUC &   0.7276 &   0.5100 &   0.7418 &   0.7049 &   0.3660 & \ValSecnd{  0.8811} &   0.5867 & \Valbest{  0.9241} &   0.7736 \\ 
				\midrule 
				
				\multirow{5}{*}{\textit{Cell1}} 
				& MPSNR $\CompFore$ &   29.76 &   29.76 &   30.15 &   16.76 &   26.34 &   28.60 &   14.91 & \Valbest{  33.13} & \ValSecnd{  31.89} \\ 
				& MPSNR $\CompBack$ &   15.72 &   15.72 &   15.79 &   23.58 &   23.58 &   23.59 &   15.13 & \ValSecnd{  29.40} & \Valbest{  38.85} \\ 
				& MSSIM $\CompFore$ & \Valbest{  0.7642} & \Valbest{  0.7642} & \ValSecnd{  0.7585} &   0.0247 &   0.3030 &   0.3114 &   0.0287 &   0.6760 &   0.5860 \\ 
				& MSSIM $\CompBack$ &   0.0421 &   0.0421 &   0.0423 &   0.2478 &   0.2492 &   0.2496 &   0.5513 & \ValSecnd{  0.7176} & \Valbest{  0.9963} \\ 
				& AUC &   0.5000 &   0.5000 &   0.7473 &   0.5044 &   0.6613 &   0.6732 &   0.7499 & \Valbest{  0.9132} & \ValSecnd{  0.8185} \\ 
				\midrule 
				
				\multirow{5}{*}{\textit{Cell2}} 
				& MPSNR $\CompFore$ &   16.50 &   28.23 &   16.25 &   16.89 &   28.93 &   30.87 &   22.71 & \Valbest{  37.30} & \ValSecnd{  35.81} \\ 
				& MPSNR $\CompBack$ &   23.84 &   16.16 &   22.57 &   24.64 &   24.60 &   24.61 &   21.13 & \ValSecnd{  31.80} & \Valbest{  32.35} \\ 
				& MSSIM $\CompFore$ &   0.0197 & \Valbest{  0.8270} &   0.0117 &   0.0274 &   0.4970 &   0.3944 &   0.0383 & \ValSecnd{  0.7994} &   0.7857 \\ 
				& MSSIM $\CompBack$ &   0.3284 &   0.0572 &   0.3434 &   0.3151 &   0.3106 &   0.3154 &   0.4254 & \ValSecnd{  0.8318} & \Valbest{  0.8600} \\ 
				& AUC &   0.7313 &   0.5035 &   0.7076 &   0.5150 &   0.9099 &   0.9231 &   0.3409 & \Valbest{  0.9825} & \ValSecnd{  0.9347} \\

				\bottomrule
			\end{tabular}
		}
	\end{center}
	\vspace{-3mm}
\end{table*}

%% file: fig_tex/result_image_cell2_g.tex
\begin{figure*}[!t]
	
	\begin{center}
		
		\begin{minipage}{0.01\hsize}
			\centerline{\rotatebox{90}{$\CompFore + \CompBack$}}
		\end{minipage}
		\begin{minipage}{0.08\hsize}
			\centerline{\includegraphics[width=\hsize]{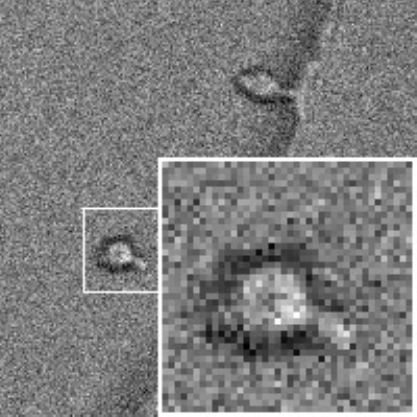}} 
		\end{minipage}
		\begin{minipage}{0.08\hsize}
			\centerline{\includegraphics[width=\hsize]{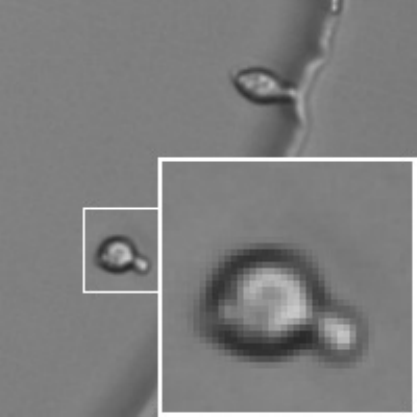}} 
		\end{minipage}
		\begin{minipage}{0.08\hsize}
			\centerline{\includegraphics[width=\hsize]{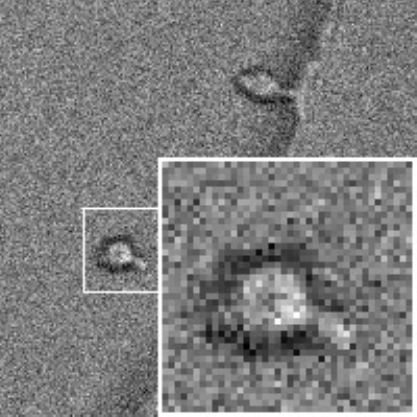}} 
		\end{minipage}
		\begin{minipage}{0.08\hsize}
			\centerline{\includegraphics[width=\hsize]{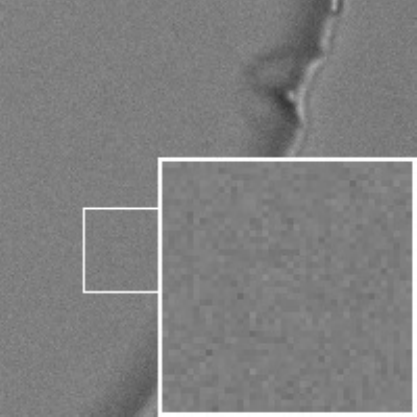}} 
		\end{minipage}
		\begin{minipage}{0.08\hsize}
			\centerline{\includegraphics[width=\hsize]{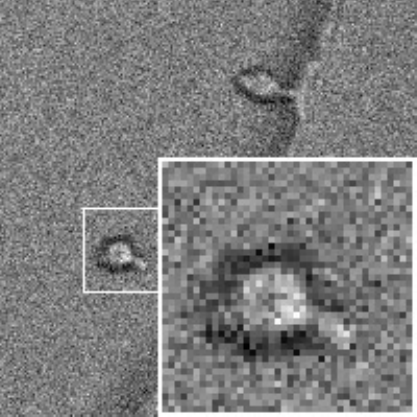}} 
		\end{minipage}
		\begin{minipage}{0.08\hsize}
			\centerline{\includegraphics[width=\hsize]{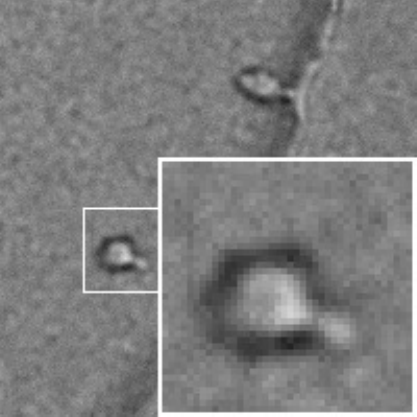}} 
		\end{minipage}
		\begin{minipage}{0.08\hsize}
			\centerline{\includegraphics[width=\hsize]{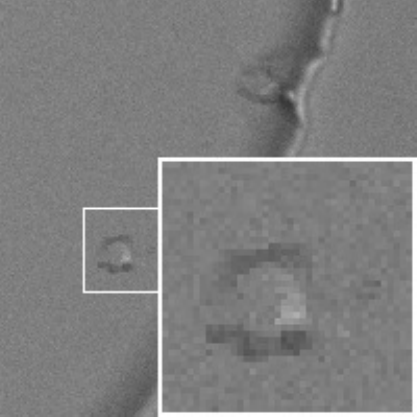}} 
		\end{minipage}
		\begin{minipage}{0.08\hsize}
			\centerline{\includegraphics[width=\hsize]{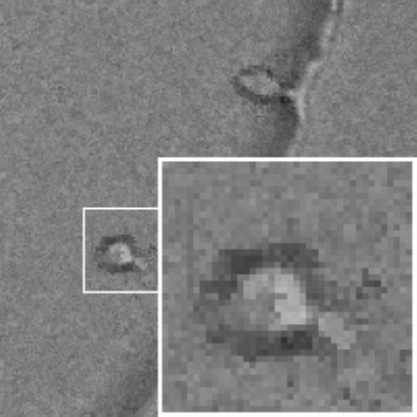}} 
		\end{minipage}
		\begin{minipage}{0.08\hsize}
			\centerline{\includegraphics[width=\hsize]{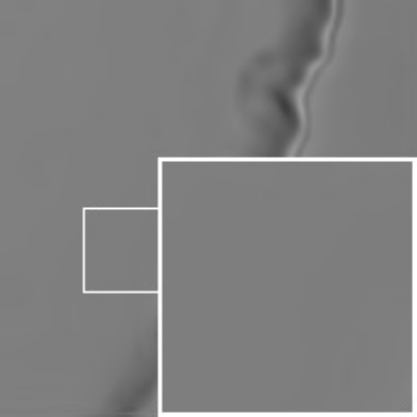}} 
		\end{minipage}
		\begin{minipage}{0.08\hsize}
			\centerline{\includegraphics[width=\hsize]{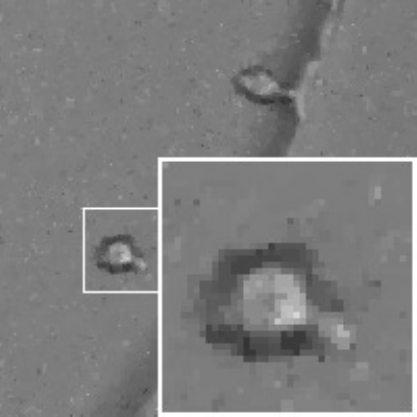}} 
		\end{minipage}
		\begin{minipage}{0.08\hsize}
			\centerline{\includegraphics[width=\hsize]{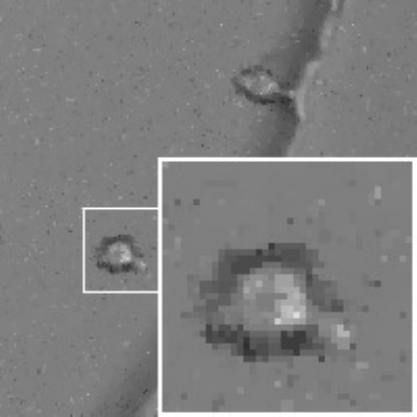}} 
		\end{minipage}
		
		\vspace{1mm}
		
		\begin{minipage}{0.01\hsize}
			\centerline{\rotatebox{90}{$\CompFore$}}
		\end{minipage}
		\begin{minipage}{0.08\hsize}
			~
		\end{minipage}
		\begin{minipage}{0.08\hsize}
			\centerline{\includegraphics[width=\hsize]{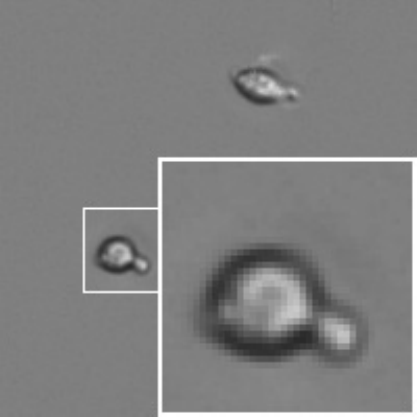}} 
		\end{minipage}
		\begin{minipage}{0.08\hsize}
			\centerline{\includegraphics[width=\hsize]{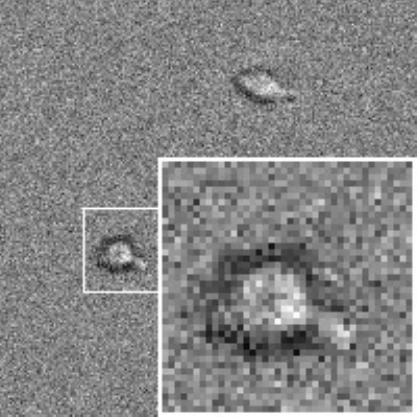}} 
		\end{minipage}
		\begin{minipage}{0.08\hsize}
			\centerline{\includegraphics[width=\hsize]{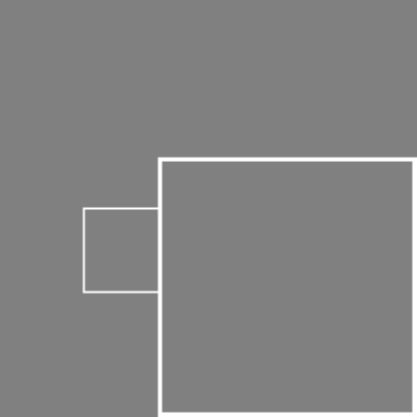}} 
		\end{minipage}
		\begin{minipage}{0.08\hsize}
			\centerline{\includegraphics[width=\hsize]{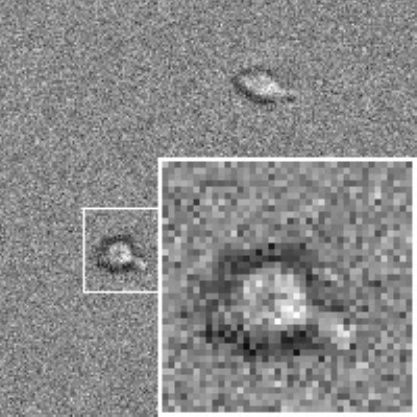}} 
		\end{minipage}
		\begin{minipage}{0.08\hsize}
			\centerline{\includegraphics[width=\hsize]{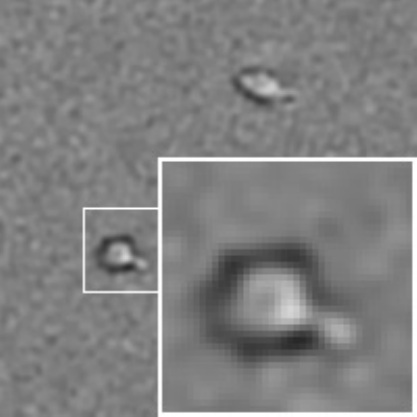}} 
		\end{minipage}
		\begin{minipage}{0.08\hsize}
			\centerline{\includegraphics[width=\hsize]{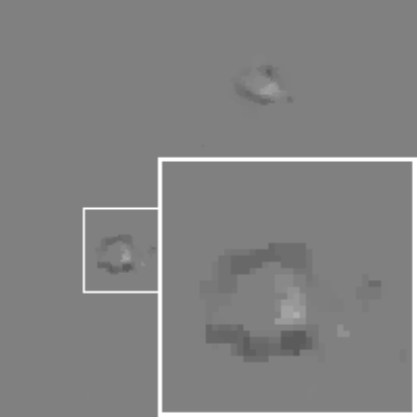}} 
		\end{minipage}
		\begin{minipage}{0.08\hsize}
			\centerline{\includegraphics[width=\hsize]{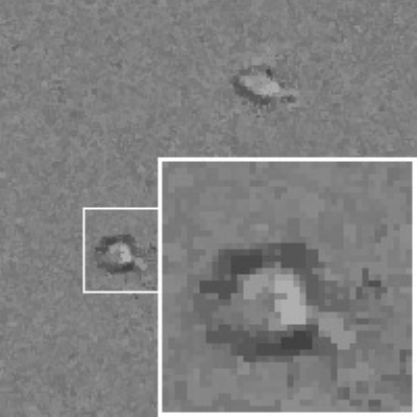}} 
		\end{minipage}
		\begin{minipage}{0.08\hsize}
			\centerline{\includegraphics[width=\hsize]{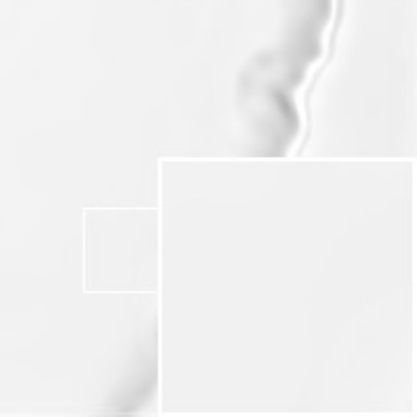}} 
		\end{minipage}
		\begin{minipage}{0.08\hsize}
			\centerline{\includegraphics[width=\hsize]{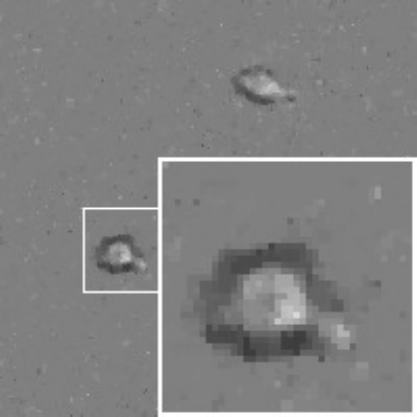}} 
		\end{minipage}
		\begin{minipage}{0.08\hsize}
			\centerline{\includegraphics[width=\hsize]{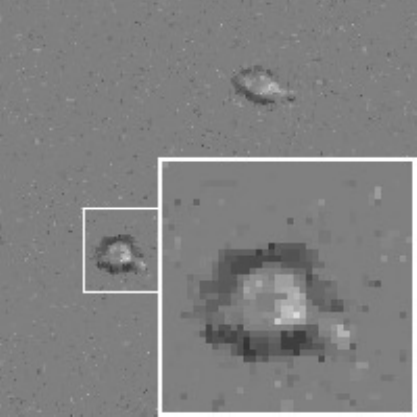}} 
		\end{minipage}
		
		\vspace{1mm}
		
		\begin{minipage}{0.01\hsize}
			\centerline{\rotatebox{90}{$\CompBack$}}
		\end{minipage}
		\begin{minipage}{0.08\hsize}
			~
		\end{minipage}
		\begin{minipage}{0.08\hsize}
			\centerline{\includegraphics[width=\hsize]{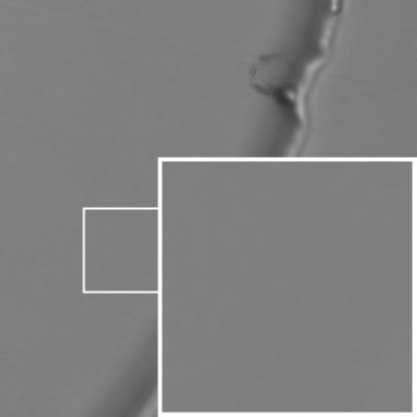}} 
		\end{minipage}
		\begin{minipage}{0.08\hsize}
			\centerline{\includegraphics[width=\hsize]{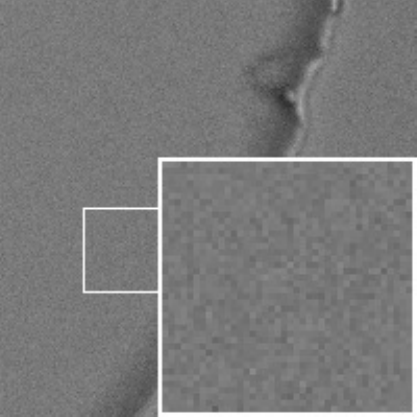}} 
		\end{minipage}
		\begin{minipage}{0.08\hsize}
			\centerline{\includegraphics[width=\hsize]{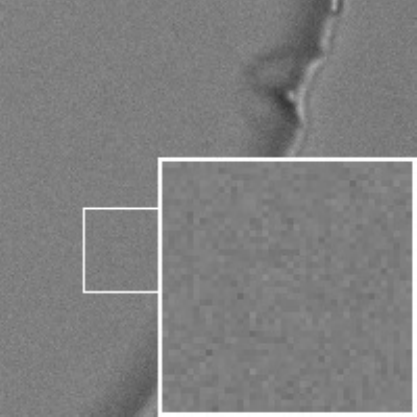}} 
		\end{minipage}
		\begin{minipage}{0.08\hsize}
			\centerline{\includegraphics[width=\hsize]{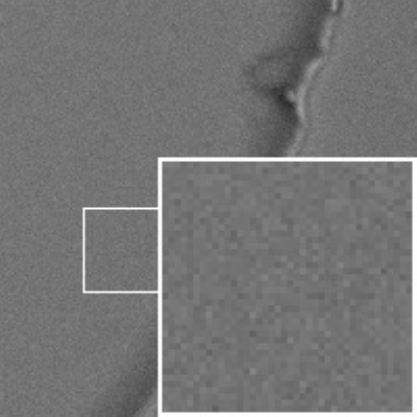}} 
		\end{minipage}
		\begin{minipage}{0.08\hsize}
			\centerline{\includegraphics[width=\hsize]{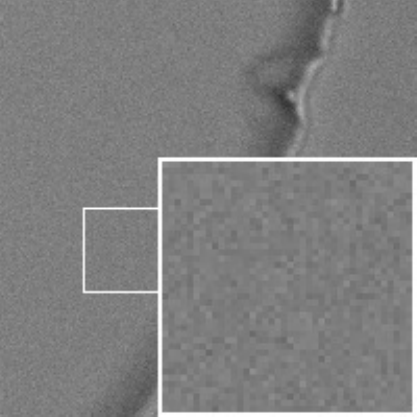}} 
		\end{minipage}
		\begin{minipage}{0.08\hsize}
			\centerline{\includegraphics[width=\hsize]{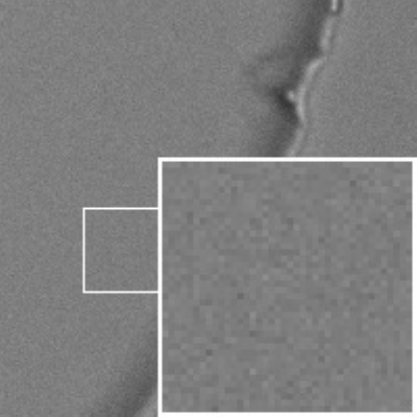}} 
		\end{minipage}
		\begin{minipage}{0.08\hsize}
			\centerline{\includegraphics[width=\hsize]{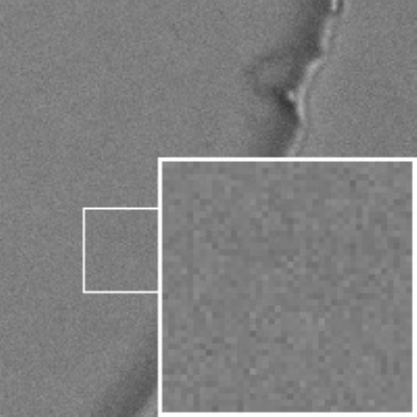}} 
		\end{minipage}
		\begin{minipage}{0.08\hsize}
			\centerline{\includegraphics[width=\hsize]{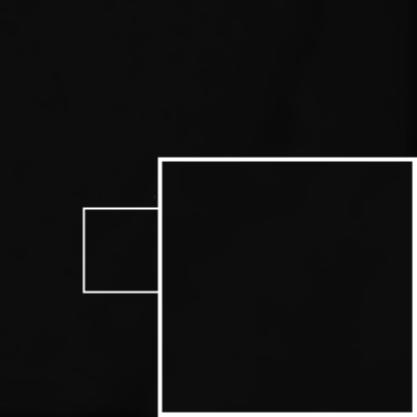}} 
		\end{minipage}
		\begin{minipage}{0.08\hsize}
			\centerline{\includegraphics[width=\hsize]{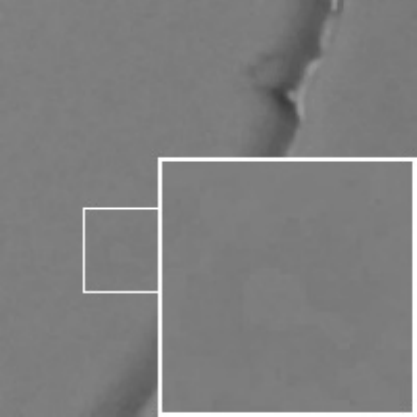}} 
		\end{minipage}
		\begin{minipage}{0.08\hsize}
			\centerline{\includegraphics[width=\hsize]{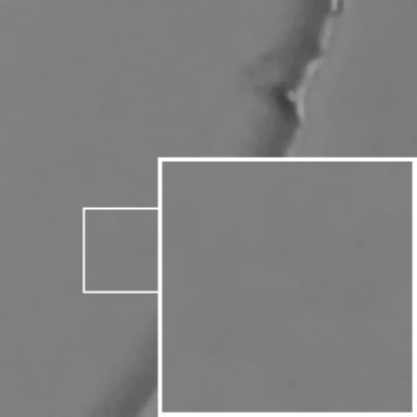}} 
		\end{minipage}
		
		\vspace{1mm}
		
		\begin{minipage}{0.01\hsize}
			~
		\end{minipage}
		\begin{minipage}{0.08\hsize}
			\centerline{\TextSizeGraph{(a)}} 
		\end{minipage}
		\begin{minipage}{0.08\hsize}
			\centerline{\TextSizeGraph{(b)}} 
		\end{minipage}
		\begin{minipage}{0.08\hsize}
			\centerline{\TextSizeGraph{(c)}}
		\end{minipage}
		\begin{minipage}{0.08\hsize}
			\centerline{\TextSizeGraph{(d)}}
		\end{minipage}
		\begin{minipage}{0.08\hsize}
			\centerline{\TextSizeGraph{(e)}} 
		\end{minipage}
		\begin{minipage}{0.08\hsize}
			\centerline{\TextSizeGraph{(f)}} 
		\end{minipage}
		\begin{minipage}{0.08\hsize}
			\centerline{\TextSizeGraph{(g)}} 
		\end{minipage}
		\begin{minipage}{0.08\hsize}
			\centerline{\TextSizeGraph{(h)}} 
		\end{minipage}
		\begin{minipage}{0.08\hsize}
			\centerline{\TextSizeGraph{(i)}}
		\end{minipage}
		\begin{minipage}{0.08\hsize}
			\centerline{\textbf{\TextSizeGraph{(j)}}}
		\end{minipage}
		\begin{minipage}{0.08\hsize}
			\centerline{\textbf{\TextSizeGraph{(k)}}} 
		\end{minipage}

	\end{center}
	
	\vspace{-3mm}
	
	\caption{FBS results for \textit{Cell2} in Case 1. 
		(a): Noisy frame. 
		(b): Ground-truth frame. 
		(c): RPCA~\cite{RPCA}. 
		(c): GNNLSM~\cite{GNNLSM_YANG_2020}. 
		(e): TVRPCA~\cite{TVRPCA}. 
		(f): PRPCA~\cite{PRPCA}. 
		(g): SRTC~\cite{SRTC_FBS_Shen_2022}. 
		(h): SS-RTD~\cite{RTD_FBS_Shen_2023}. 
		(i): FactorDVP-T~\cite{FDVP_Miao_2024}.
		\textbf{(j): \Ourss (LR).}
		\textbf{(k): \Ourss (SC).}}
	\label{fig:result_image_cell2_g}
\end{figure*}

%% file: fig_tex/result_image_cell1_gs.tex
\begin{figure*}[!t]
	
	\begin{center}
		
		\begin{minipage}{0.01\hsize}
			\centerline{\rotatebox{90}{$\CompFore + \CompBack$}}
		\end{minipage}
		\begin{minipage}{0.08\hsize}
			\centerline{\includegraphics[width=\hsize]{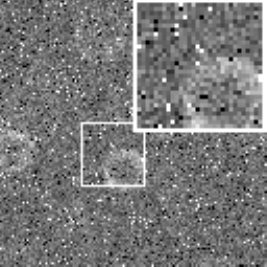}} 
		\end{minipage}
		\begin{minipage}{0.08\hsize}
			\centerline{\includegraphics[width=\hsize]{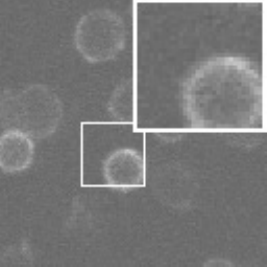}} 
		\end{minipage}
		\begin{minipage}{0.08\hsize}
			\centerline{\includegraphics[width=\hsize]{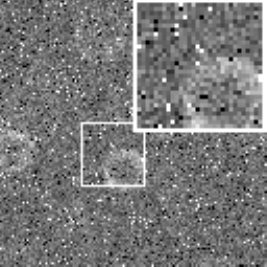}} 
		\end{minipage}
		\begin{minipage}{0.08\hsize}
			\centerline{\includegraphics[width=\hsize]{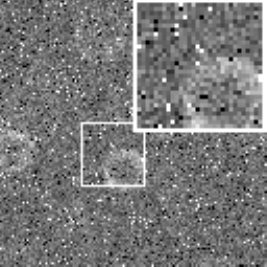}} 
		\end{minipage}
		\begin{minipage}{0.08\hsize}
			\centerline{\includegraphics[width=\hsize]{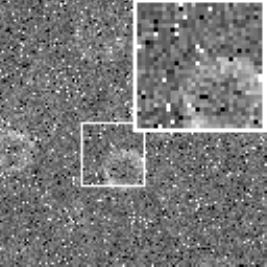}} 
		\end{minipage}
		\begin{minipage}{0.08\hsize}
			\centerline{\includegraphics[width=\hsize]{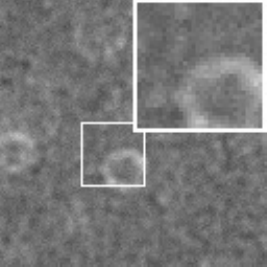}} 
		\end{minipage}
		\begin{minipage}{0.08\hsize}
			\centerline{\includegraphics[width=\hsize]{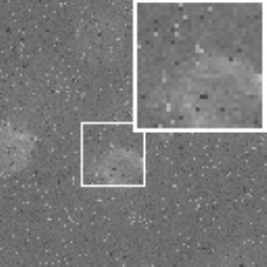}} 
		\end{minipage}
		\begin{minipage}{0.08\hsize}
			\centerline{\includegraphics[width=\hsize]{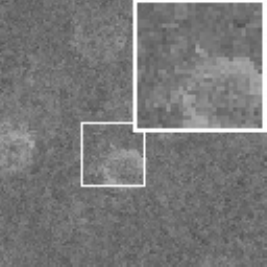}} 
		\end{minipage}
		\begin{minipage}{0.08\hsize}
			\centerline{\includegraphics[width=\hsize]{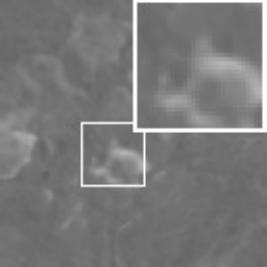}} 
		\end{minipage}
		\begin{minipage}{0.08\hsize}
			\centerline{\includegraphics[width=\hsize]{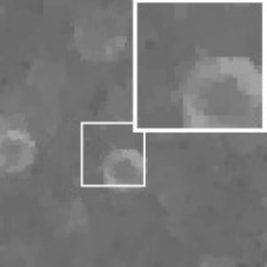}} 
		\end{minipage}
		\begin{minipage}{0.08\hsize}
			\centerline{\includegraphics[width=\hsize]{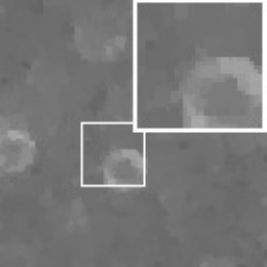}} 
		\end{minipage}
		
		\vspace{1mm}
		
		\begin{minipage}{0.01\hsize}
			\centerline{\rotatebox{90}{$\CompFore$}}
		\end{minipage}
		\begin{minipage}{0.08\hsize}
			~
		\end{minipage}
		\begin{minipage}{0.08\hsize}
			\centerline{\includegraphics[width=\hsize]{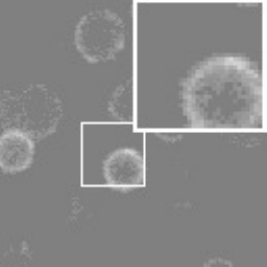}} 
		\end{minipage}
		\begin{minipage}{0.08\hsize}
			\centerline{\includegraphics[width=\hsize]{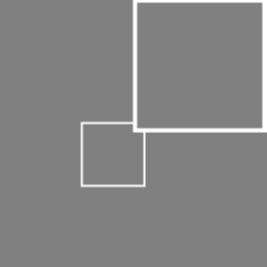}} 
		\end{minipage}
		\begin{minipage}{0.08\hsize}
			\centerline{\includegraphics[width=\hsize]{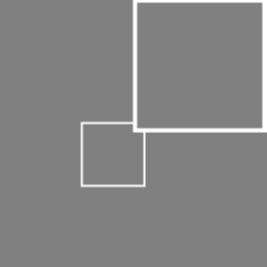}} 
		\end{minipage}
		\begin{minipage}{0.08\hsize}
			\centerline{\includegraphics[width=\hsize]{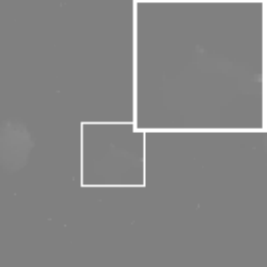}} 
		\end{minipage}
		\begin{minipage}{0.08\hsize}
			\centerline{\includegraphics[width=\hsize]{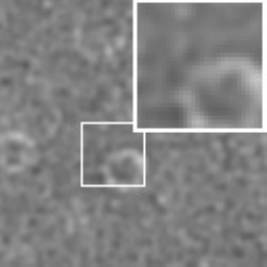}} 
		\end{minipage}
		\begin{minipage}{0.08\hsize}
			\centerline{\includegraphics[width=\hsize]{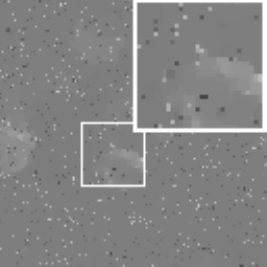}} 
		\end{minipage}
		\begin{minipage}{0.08\hsize}
			\centerline{\includegraphics[width=\hsize]{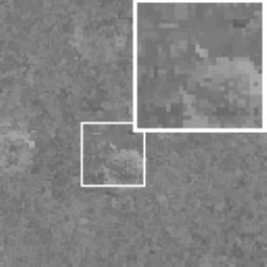}} 
		\end{minipage}
		\begin{minipage}{0.08\hsize}
			\centerline{\includegraphics[width=\hsize]{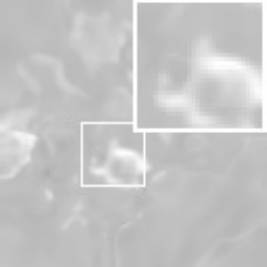}} 
		\end{minipage}
		\begin{minipage}{0.08\hsize}
			\centerline{\includegraphics[width=\hsize]{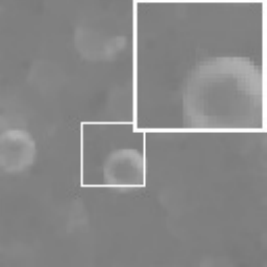}} 
		\end{minipage}
		\begin{minipage}{0.08\hsize}
			\centerline{\includegraphics[width=\hsize]{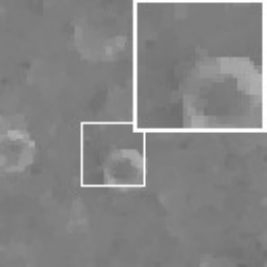}} 
		\end{minipage}
		
		\vspace{1mm}
		
		\begin{minipage}{0.01\hsize}
			\centerline{\rotatebox{90}{$\CompBack$}}
		\end{minipage}
		\begin{minipage}{0.08\hsize}
			~
		\end{minipage}
		\begin{minipage}{0.08\hsize}
			\centerline{\includegraphics[width=\hsize]{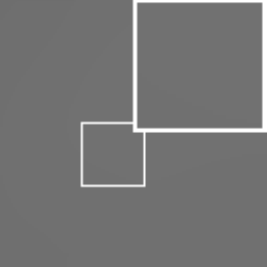}} 
		\end{minipage}
		\begin{minipage}{0.08\hsize}
			\centerline{\includegraphics[width=\hsize]{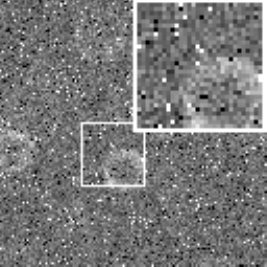}} 
		\end{minipage}
		\begin{minipage}{0.08\hsize}
			\centerline{\includegraphics[width=\hsize]{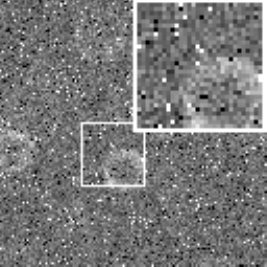}} 
		\end{minipage}
		\begin{minipage}{0.08\hsize}
			\centerline{\includegraphics[width=\hsize]{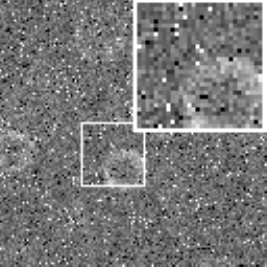}} 
		\end{minipage}
		\begin{minipage}{0.08\hsize}
			\centerline{\includegraphics[width=\hsize]{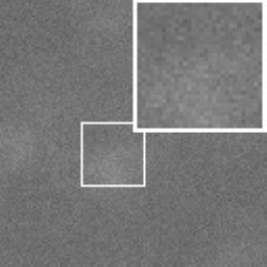}} 
		\end{minipage}
		\begin{minipage}{0.08\hsize}
			\centerline{\includegraphics[width=\hsize]{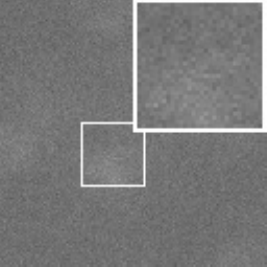}} 
		\end{minipage}
		\begin{minipage}{0.08\hsize}
			\centerline{\includegraphics[width=\hsize]{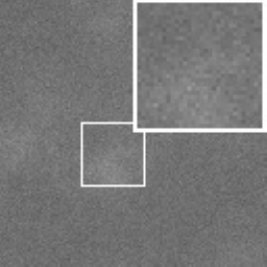}} 
		\end{minipage}
		\begin{minipage}{0.08\hsize}
			\centerline{\includegraphics[width=\hsize]{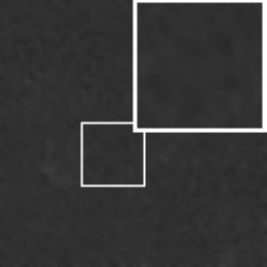}} 
		\end{minipage}
		\begin{minipage}{0.08\hsize}
			\centerline{\includegraphics[width=\hsize]{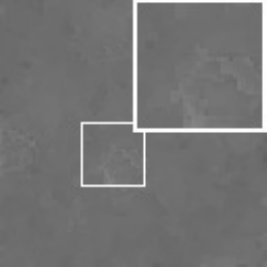}} 
		\end{minipage}
		\begin{minipage}{0.08\hsize}
			\centerline{\includegraphics[width=\hsize]{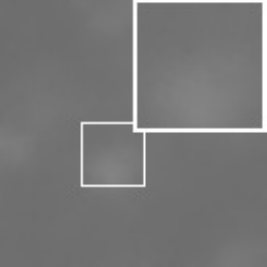}} 
		\end{minipage}
		
		\vspace{1mm}
		
		\begin{minipage}{0.01\hsize}
			~
		\end{minipage}
		\begin{minipage}{0.08\hsize}
			\centerline{\TextSizeGraph{(a)}} 
		\end{minipage}
		\begin{minipage}{0.08\hsize}
			\centerline{\TextSizeGraph{(b)}} 
		\end{minipage}
		\begin{minipage}{0.08\hsize}
			\centerline{\TextSizeGraph{(c)}}
		\end{minipage}
		\begin{minipage}{0.08\hsize}
			\centerline{\TextSizeGraph{(d)}}
		\end{minipage}
		\begin{minipage}{0.08\hsize}
			\centerline{\TextSizeGraph{(e)}} 
		\end{minipage}
		\begin{minipage}{0.08\hsize}
			\centerline{\TextSizeGraph{(f)}} 
		\end{minipage}
		\begin{minipage}{0.08\hsize}
			\centerline{\TextSizeGraph{(g)}} 
		\end{minipage}
		\begin{minipage}{0.08\hsize}
			\centerline{\TextSizeGraph{(h)}} 
		\end{minipage}
		\begin{minipage}{0.08\hsize}
			\centerline{\TextSizeGraph{(i)}}
		\end{minipage}
		\begin{minipage}{0.08\hsize}
			\centerline{\textbf{\TextSizeGraph{(j)}}}
		\end{minipage}
		\begin{minipage}{0.08\hsize}
			\centerline{\textbf{\TextSizeGraph{(k)}}} 
		\end{minipage}
		
	\end{center}
	
	\vspace{-3mm}

	\caption{FBS results for \textit{Cell1} in Case 2. 
		(a): Noisy frame. 
		(b): Ground-truth frame. 
		(c): RPCA~\cite{RPCA}. 
		(d): GNNLSM~\cite{GNNLSM_YANG_2020}. 
		(e): TVRPCA~\cite{TVRPCA}. 
		(f): PRPCA~\cite{PRPCA}. 
		(g): SRTC~\cite{SRTC_FBS_Shen_2022}. 
		(h): SS-RTD~\cite{RTD_FBS_Shen_2023}. 
		(i): FactorDVP-T~\cite{FDVP_Miao_2024}.
		\textbf{(j): \Ourss (LR).}
		\textbf{(k): \Ourss (SC).}}
	\label{fig:result_image_cell1_gs}
\end{figure*}

%% file: fig_tex/result_image_CAMEL_17_gsst.tex
\begin{figure*}[!t]
	
	\begin{center}
		
		\begin{minipage}{0.01\hsize}
			\centerline{\rotatebox{90}{$\CompFore + \CompBack$}}
		\end{minipage}
		\begin{minipage}{0.08\hsize}
			\centerline{\includegraphics[width=\hsize]{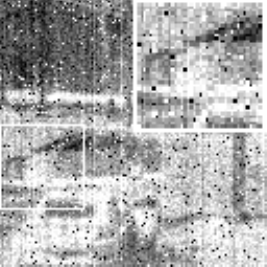}} 
		\end{minipage}
		\begin{minipage}{0.08\hsize}
			\centerline{\includegraphics[width=\hsize]{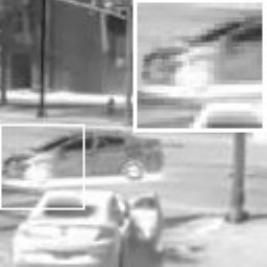}} 
		\end{minipage}
		\begin{minipage}{0.08\hsize}
			\centerline{\includegraphics[width=\hsize]{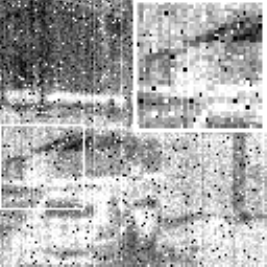}} 
		\end{minipage}
		\begin{minipage}{0.08\hsize}
			\centerline{\includegraphics[width=\hsize]{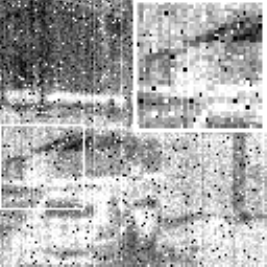}} 
		\end{minipage}
		\begin{minipage}{0.08\hsize}
			\centerline{\includegraphics[width=\hsize]{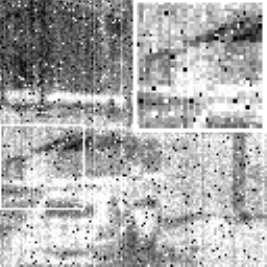}} 
		\end{minipage}
		\begin{minipage}{0.08\hsize}
			\centerline{\includegraphics[width=\hsize]{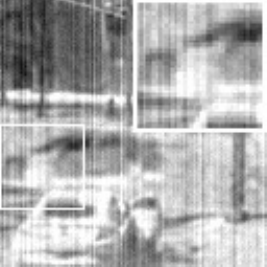}} 
		\end{minipage}
		\begin{minipage}{0.08\hsize}
			\centerline{\includegraphics[width=\hsize]{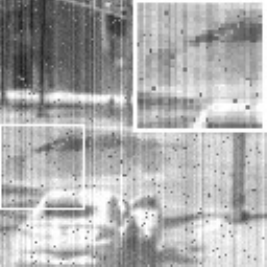}} 
		\end{minipage}
		\begin{minipage}{0.08\hsize}
			\centerline{\includegraphics[width=\hsize]{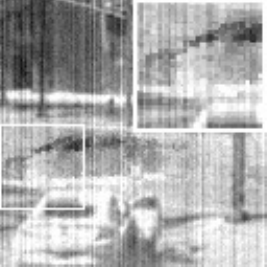}} 
		\end{minipage}
		\begin{minipage}{0.08\hsize}
			\centerline{\includegraphics[width=\hsize]{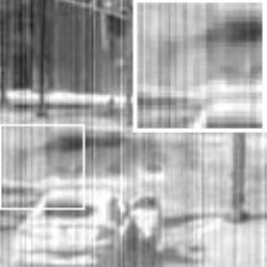}} 
		\end{minipage}
		\begin{minipage}{0.08\hsize}
			\centerline{\includegraphics[width=\hsize]{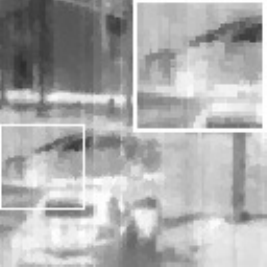}} 
		\end{minipage}
		\begin{minipage}{0.08\hsize}
			\centerline{\includegraphics[width=\hsize]{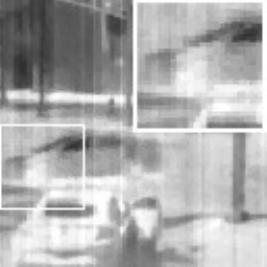}} 
		\end{minipage}
		
		\vspace{1mm}
		
		\begin{minipage}{0.01\hsize}
			\centerline{\rotatebox{90}{$\CompFore$}}
		\end{minipage}
		\begin{minipage}{0.08\hsize}
			~
		\end{minipage}
		\begin{minipage}{0.08\hsize}
			\centerline{\includegraphics[width=\hsize]{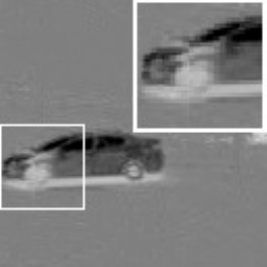}} 
		\end{minipage}
		\begin{minipage}{0.08\hsize}
			\centerline{\includegraphics[width=\hsize]{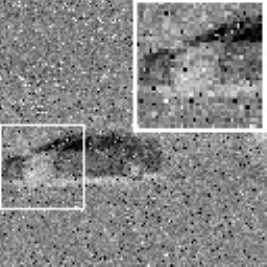}} 
		\end{minipage}
		\begin{minipage}{0.08\hsize}
			\centerline{\includegraphics[width=\hsize]{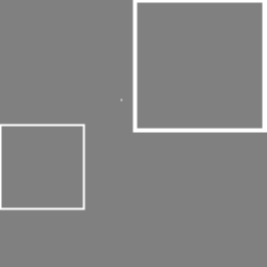}} 
		\end{minipage}
		\begin{minipage}{0.08\hsize}
			\centerline{\includegraphics[width=\hsize]{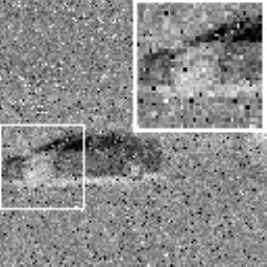}} 
		\end{minipage}
		\begin{minipage}{0.08\hsize}
			\centerline{\includegraphics[width=\hsize]{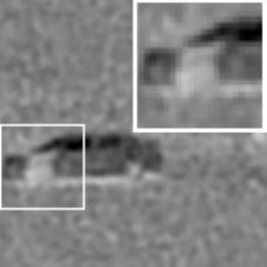}} 
		\end{minipage}
		\begin{minipage}{0.08\hsize}
			\centerline{\includegraphics[width=\hsize]{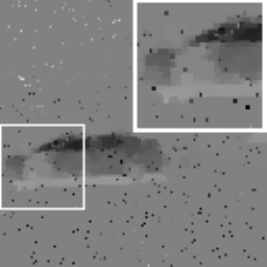}} 
		\end{minipage}
		\begin{minipage}{0.08\hsize}
			\centerline{\includegraphics[width=\hsize]{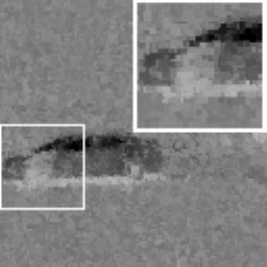}} 
		\end{minipage}
		\begin{minipage}{0.08\hsize}
			\centerline{\includegraphics[width=\hsize]{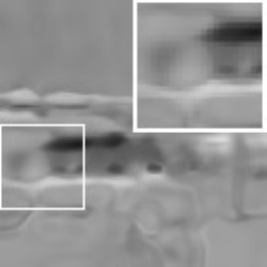}} 
		\end{minipage}
		\begin{minipage}{0.08\hsize}
			\centerline{\includegraphics[width=\hsize]{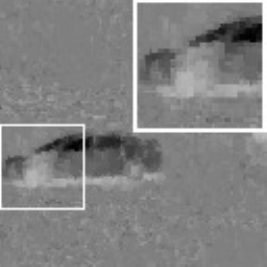}} 
		\end{minipage}
		\begin{minipage}{0.08\hsize}
			\centerline{\includegraphics[width=\hsize]{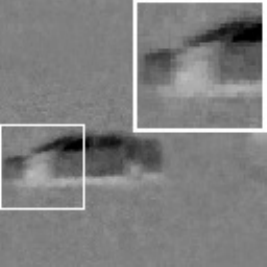}} 
		\end{minipage}
		
		\vspace{1mm}
		
		\begin{minipage}{0.01\hsize}
			\centerline{\rotatebox{90}{$\CompBack$}}
		\end{minipage}
		\begin{minipage}{0.08\hsize}
			~
		\end{minipage}
		\begin{minipage}{0.08\hsize}
			\centerline{\includegraphics[width=\hsize]{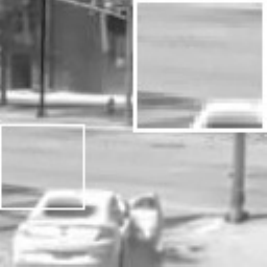}} 
		\end{minipage}
		\begin{minipage}{0.08\hsize}
			\centerline{\includegraphics[width=\hsize]{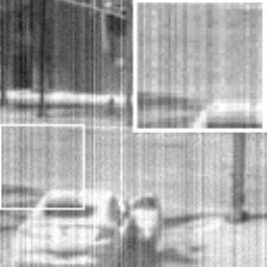}} 
		\end{minipage}
		\begin{minipage}{0.08\hsize}
			\centerline{\includegraphics[width=\hsize]{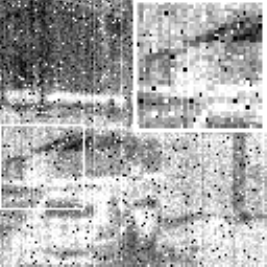}} 
		\end{minipage}
		\begin{minipage}{0.08\hsize}
			\centerline{\includegraphics[width=\hsize]{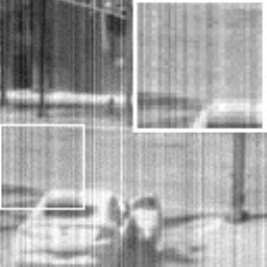}} 
		\end{minipage}
		\begin{minipage}{0.08\hsize}
			\centerline{\includegraphics[width=\hsize]{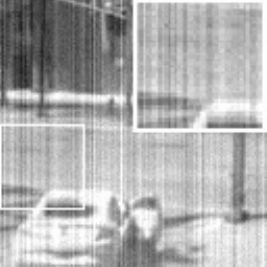}} 
		\end{minipage}
		\begin{minipage}{0.08\hsize}
			\centerline{\includegraphics[width=\hsize]{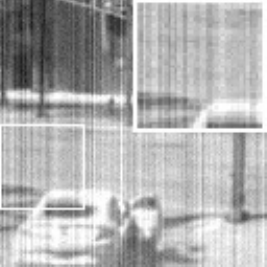}} 
		\end{minipage}
		\begin{minipage}{0.08\hsize}
			\centerline{\includegraphics[width=\hsize]{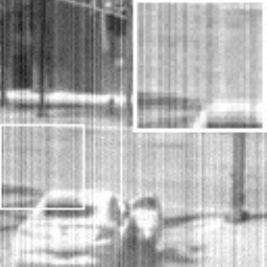}} 
		\end{minipage}
		\begin{minipage}{0.08\hsize}
			\centerline{\includegraphics[width=\hsize]{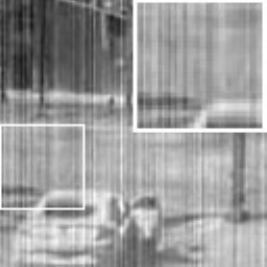}} 
		\end{minipage}
		\begin{minipage}{0.08\hsize}
			\centerline{\includegraphics[width=\hsize]{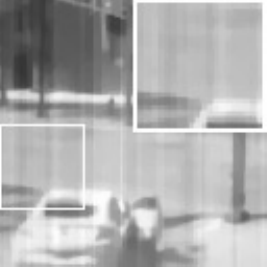}} 
		\end{minipage}
		\begin{minipage}{0.08\hsize}
			\centerline{\includegraphics[width=\hsize]{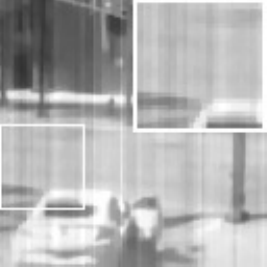}} 
		\end{minipage}
		
		\vspace{1mm}
		
		\begin{minipage}{0.01\hsize}
			~
		\end{minipage}
		\begin{minipage}{0.08\hsize}
			\centerline{\TextSizeGraph{(a)}} 
		\end{minipage}
		\begin{minipage}{0.08\hsize}
			\centerline{\TextSizeGraph{(b)}} 
		\end{minipage}
		\begin{minipage}{0.08\hsize}
			\centerline{\TextSizeGraph{(c)}}
		\end{minipage}
		\begin{minipage}{0.08\hsize}
			\centerline{\TextSizeGraph{(d)}}
		\end{minipage}
		\begin{minipage}{0.08\hsize}
			\centerline{\TextSizeGraph{(e)}} 
		\end{minipage}
		\begin{minipage}{0.08\hsize}
			\centerline{\TextSizeGraph{(f)}} 
		\end{minipage}
		\begin{minipage}{0.08\hsize}
			\centerline{\TextSizeGraph{(g)}} 
		\end{minipage}
		\begin{minipage}{0.08\hsize}
			\centerline{\TextSizeGraph{(h)}} 
		\end{minipage}
		\begin{minipage}{0.08\hsize}
			\centerline{\TextSizeGraph{(i)}}
		\end{minipage}
		\begin{minipage}{0.08\hsize}
			\centerline{\textbf{\TextSizeGraph{(j)}}}
		\end{minipage}
		\begin{minipage}{0.08\hsize}
			\centerline{\textbf{\TextSizeGraph{(k)}}} 
		\end{minipage}

	\end{center}
	
	\vspace{-3mm}

	\caption{FBS results for \textit{CAMEL seq17} in Case 3. 
		(a): Noisy frame. 
		(b): Ground-truth frame. 
		(c): RPCA~\cite{RPCA}. 
		(d): GNNLSM~\cite{GNNLSM_YANG_2020}. 
		(e): TVRPCA~\cite{TVRPCA}. 
		(f): PRPCA~\cite{PRPCA}. 
		(g): SRTC~\cite{SRTC_FBS_Shen_2022}. 
		(h): SS-RTD~\cite{RTD_FBS_Shen_2023}. 
		(i): FactorDVP-T~\cite{FDVP_Miao_2024}.
		\textbf{(j): \Ourss (LR).}
		\textbf{(k): \Ourss (SC).}}
	\label{fig:result_image_CAMEL_17_gsst}
\end{figure*}

%% file: fig_tex/result_filter.tex
%
%
%
%
%
%
%

\begin{figure*}[!t]
	
	\begin{center}
		
		\begin{minipage}{0.01\hsize}
			\centerline{\scriptsize{\rotatebox{90}{\Ourss (LR)}}}
		\end{minipage}
		\begin{minipage}{0.095\hsize}
			\centerline{\includegraphics[width=\hsize]{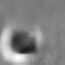}}
		\end{minipage}
		\begin{minipage}{0.095\hsize}
			\centerline{\includegraphics[width=\hsize]{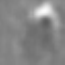}}
		\end{minipage}
		\begin{minipage}{0.095\hsize}
			\centerline{\includegraphics[width=\hsize]{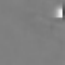}}
		\end{minipage}
		\begin{minipage}{0.095\hsize}
			\centerline{\includegraphics[width=\hsize]{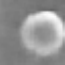}}
		\end{minipage}
		\begin{minipage}{0.095\hsize}
			\centerline{\includegraphics[width=\hsize]{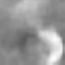}}
		\end{minipage}
		\begin{minipage}{0.095\hsize}
			\centerline{\includegraphics[width=\hsize]{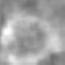}}
		\end{minipage}
		\begin{minipage}{0.095\hsize}
			\centerline{\includegraphics[width=\hsize]{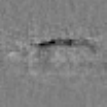}}
		\end{minipage}
		\begin{minipage}{0.095\hsize}
			\centerline{\includegraphics[width=\hsize]{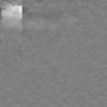}}
		\end{minipage}
		\begin{minipage}{0.095\hsize}
			\centerline{\includegraphics[width=\hsize]{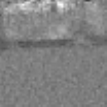}}
		\end{minipage}
		
		\vspace{1mm}
		
		\begin{minipage}{0.01\hsize}
			\centerline{\scriptsize{\rotatebox{90}{\Ourss (SC)}}}
		\end{minipage}
		\begin{minipage}{0.095\hsize}
			\centerline{\includegraphics[width=\hsize]{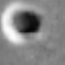}}
		\end{minipage}
		\begin{minipage}{0.095\hsize}
			\centerline{\includegraphics[width=\hsize]{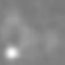}}
		\end{minipage}
		\begin{minipage}{0.095\hsize}
			\centerline{\includegraphics[width=\hsize]{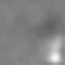}}
		\end{minipage}
		\begin{minipage}{0.095\hsize}
			\centerline{\includegraphics[width=\hsize]{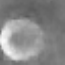}}
		\end{minipage}
		\begin{minipage}{0.095\hsize}
			\centerline{\includegraphics[width=\hsize]{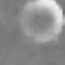}}
		\end{minipage}
		\begin{minipage}{0.095\hsize}
			\centerline{\includegraphics[width=\hsize]{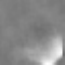}}
		\end{minipage}
		\begin{minipage}{0.095\hsize}
			\centerline{\includegraphics[width=\hsize]{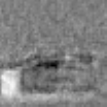}}
		\end{minipage}
		\begin{minipage}{0.095\hsize}
			\centerline{\includegraphics[width=\hsize]{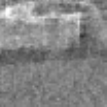}}
		\end{minipage}
		\begin{minipage}{0.095\hsize}
			\centerline{\includegraphics[width=\hsize]{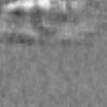}}
		\end{minipage}
		
		\vspace{1mm}
		
		\begin{minipage}{0.01\hsize}
			~
		\end{minipage}
		\begin{minipage}{0.095\hsize}
			\centerline{\TextSizeGraph{(a)}} 
		\end{minipage}
		\begin{minipage}{0.095\hsize}
			\centerline{\TextSizeGraph{(b)}} 
		\end{minipage}
		\begin{minipage}{0.095\hsize}
			\centerline{\TextSizeGraph{(c)}}
		\end{minipage}
		\begin{minipage}{0.095\hsize}
			\centerline{\TextSizeGraph{(d)}}
		\end{minipage}
		\begin{minipage}{0.095\hsize}
			\centerline{\TextSizeGraph{(e)}} 
		\end{minipage}
		\begin{minipage}{0.095\hsize}
			\centerline{\TextSizeGraph{(f)}} 
		\end{minipage}
		\begin{minipage}{0.095\hsize}
			\centerline{\TextSizeGraph{(g)}} 
		\end{minipage}
		\begin{minipage}{0.095\hsize}
			\centerline{\TextSizeGraph{(h)}} 
		\end{minipage}
		\begin{minipage}{0.095\hsize}
			\centerline{\TextSizeGraph{(i)}}
		\end{minipage}
		
	\end{center}

	\vspace{-3mm}
	
	\caption{Some bases $\Dict$ obtained in the FBS process of \Ours. 
		(a)-(c): \textit{Cell2} in Case 1. 
		(d)-(f): \textit{Cell1} in Case 2.
		(g)-(i): \textit{CAMEL seq17} in Case 3.}
	\label{fig:result_filter}
	\vspace{-2mm}
\end{figure*}

%% file: table_tex/result_abr.tex
\begin{table*}[!h]
	\begin{center}
		\caption{MPSNR, MSSIM, and AUC Values of the Ablation Experiments.}
		\label{tab:results_abl}
					\vspace{-2mm}
		\scalebox{0.80}{
			\begin{tabular}{cccccccccccccc}
				\toprule
				\multirow{3}{*}{Video} &  \multirow{3}{*}{Measure} & 
				\multicolumn{6}{c}{LR (Left: without CSR, Right: with CSR)} & 
				\multicolumn{6}{c}{SC (Left: without CSR, Right: with CSR)}
				\\
				\cmidrule(lr){3-8} \cmidrule(lr){9-14}
				& & \multicolumn{2}{c}{Case 1} & \multicolumn{2}{c}{Case 2} & \multicolumn{2}{c}{Case 3} & \multicolumn{2}{c}{Case 1} & \multicolumn{2}{c}{Case 2} & \multicolumn{2}{c}{Case 3} \\ \addlinespace[-1pt]
				\cmidrule(lr){3-4} \cmidrule(lr){5-6} \cmidrule(lr){7-8} \cmidrule(lr){9-10} \cmidrule(lr){11-12} \cmidrule(lr){13-14} 
				& 
				& -- & \checkmark  & -- & \checkmark & -- & \checkmark & -- & \checkmark & -- & \checkmark & -- & \checkmark \\ 
				\midrule
				
				\multirow{5}{*}{\textit{CAMEL seq8}} 
				& MPSNR $\CompFore$ &   31.66 & \Valbest{  32.32} &   31.27 & \Valbest{  31.74} &   30.74 & \Valbest{  31.64} &   30.85 & \Valbest{  32.68} &   30.88 & \Valbest{  32.51} &   30.15 & \Valbest{  32.08} \\ 
				& MPSNR $\CompBack$ & \Valbest{  31.80} &   30.78 &   30.27 & \Valbest{  31.10} & \Valbest{  28.69} &   27.98 &   32.17 & \Valbest{  34.87} &   30.36 & \Valbest{  32.97} & \Valbest{  29.01} &   27.92 \\ 
				& MSSIM $\CompFore$ &   0.6107 & \Valbest{  0.6546} &   0.6194 & \Valbest{  0.7842} &   0.5967 & \Valbest{  0.7557} &   0.5618 & \Valbest{  0.6314} &   0.5850 & \Valbest{  0.6912} &   0.5616 & \Valbest{  0.6288} \\ 
				& MSSIM $\CompBack$ &   0.8693 & \Valbest{  0.9443} &   0.8381 & \Valbest{  0.8603} & \Valbest{  0.7946} &   0.7263 &   0.8894 & \Valbest{  0.9374} &   0.8455 & \Valbest{  0.9030} & \Valbest{  0.8211} &   0.7351 \\ 
				& AUC &   0.9852 & \Valbest{  0.9867} &   0.9840 & \Valbest{  0.9972} &   0.9780 & \Valbest{  0.9969} &   0.9775 & \Valbest{  0.9921} &   0.9799 & \Valbest{  0.9921} &   0.9688 & \Valbest{  0.9918} \\ 
				\cmidrule(lr){1-2} \cmidrule(lr){3-8} \cmidrule(lr){9-14} 
				
				\multirow{5}{*}{\textit{CAMEL seq10}} 
				& MPSNR $\CompFore$ & \Valbest{  31.39} &   31.17 &   31.50 & \Valbest{  31.81} &   30.19 & \Valbest{  30.84} &   29.88 & \Valbest{  31.11} &   30.39 & \Valbest{  31.52} &   29.19 & \Valbest{  29.49} \\ 
				& MPSNR $\CompBack$ &   33.36 & \Valbest{  36.32} &   32.92 & \Valbest{  32.96} &   28.98 & \Valbest{  29.02} &   32.44 & \Valbest{  32.72} &   31.44 & \Valbest{  32.24} &   28.39 & \Valbest{  28.79} \\ 
				& MSSIM $\CompFore$ &   0.6162 & \Valbest{  0.6211} &   0.6344 & \Valbest{  0.6470} &   0.5864 & \Valbest{  0.6117} &   0.5292 & \Valbest{  0.5763} &   0.5967 & \Valbest{  0.6116} &   0.5343 & \Valbest{  0.5442} \\ 
				& MSSIM $\CompBack$ &   0.8992 & \Valbest{  0.9408} & \Valbest{  0.8965} &   0.8964 & \Valbest{  0.7908} &   0.7878 &   0.9394 & \Valbest{  0.9549} &   0.9150 & \Valbest{  0.9396} &   0.8153 & \Valbest{  0.8213} \\ 
				& AUC & \Valbest{  0.9737} &   0.9659 &   0.9753 & \Valbest{  0.9783} &   0.9585 & \Valbest{  0.9682} &   0.9514 & \Valbest{  0.9626} &   0.9556 & \Valbest{  0.9694} &   0.9260 & \Valbest{  0.9341} \\ 
				\cmidrule(lr){1-2} \cmidrule(lr){3-8} \cmidrule(lr){9-14} 
				
				\multirow{5}{*}{\textit{CAMEL seq17}} 
				& MPSNR $\CompFore$ &   29.94 & \Valbest{  30.54} &   29.59 & \Valbest{  30.11} &   29.02 & \Valbest{  29.82} &   29.71 & \Valbest{  31.45} &   29.55 & \Valbest{  31.13} &   28.84 & \Valbest{  30.99} \\ 
				& MPSNR $\CompBack$ &   29.51 & \Valbest{  30.76} & \Valbest{  28.18} &   28.12 &   26.94 & \Valbest{  27.52} &   30.31 & \Valbest{  32.42} &   28.52 & \Valbest{  30.42} &   27.33 & \Valbest{  27.42} \\ 
				& MSSIM $\CompFore$ &   0.5578 & \Valbest{  0.6018} &   0.5504 & \Valbest{  0.6171} &   0.5342 & \Valbest{  0.5816} &   0.5115 & \Valbest{  0.5708} &   0.5153 & \Valbest{  0.5531} &   0.4980 & \Valbest{  0.5529} \\ 
				& MSSIM $\CompBack$ &   0.8636 & \Valbest{  0.9023} &   0.8238 & \Valbest{  0.8526} &   0.7948 & \Valbest{  0.8205} &   0.8900 & \Valbest{  0.9350} &   0.8400 & \Valbest{  0.8974} & \Valbest{  0.8215} &   0.8149 \\ 
				& AUC &   0.9539 & \Valbest{  0.9612} &   0.9506 & \Valbest{  0.9573} &   0.9393 & \Valbest{  0.9451} &   0.9420 & \Valbest{  0.9711} &   0.9419 & \Valbest{  0.9694} &   0.9288 & \Valbest{  0.9670} \\ 
				\cmidrule(lr){1-2} \cmidrule(lr){3-8} \cmidrule(lr){9-14} 
				
				\multirow{5}{*}{\textit{CAMEL seq18}} 
				& MPSNR $\CompFore$ &   29.52 & \Valbest{  30.02} &   29.10 & \Valbest{  29.39} &   28.36 & \Valbest{  29.05} &   29.52 & \Valbest{  30.99} &   29.15 & \Valbest{  30.48} &   28.39 & \Valbest{  30.46} \\ 
				& MPSNR $\CompBack$ &   29.78 & \Valbest{  30.22} & \Valbest{  28.33} &   28.29 &   26.93 & \Valbest{  26.98} &   30.48 & \Valbest{  32.63} &   28.61 & \Valbest{  30.66} & \Valbest{  27.25} &   27.10 \\ 
				& MSSIM $\CompFore$ &   0.5427 & \Valbest{  0.5942} &   0.5350 & \Valbest{  0.5521} &   0.5026 & \Valbest{  0.5317} &   0.5088 & \Valbest{  0.5660} &   0.5010 & \Valbest{  0.6143} &   0.4771 & \Valbest{  0.5667} \\ 
				& MSSIM $\CompBack$ &   0.8697 & \Valbest{  0.9000} &   0.8280 & \Valbest{  0.8288} &   0.7853 & \Valbest{  0.7855} &   0.8872 & \Valbest{  0.9331} &   0.8384 & \Valbest{  0.9004} & \Valbest{  0.7975} &   0.7769 \\ 
				& AUC &   0.9724 & \Valbest{  0.9756} &   0.9673 & \Valbest{  0.9729} &   0.9595 & \Valbest{  0.9689} &   0.9713 & \Valbest{  0.9861} &   0.9665 & \Valbest{  0.9836} &   0.9593 & \Valbest{  0.9828} \\ 
				\cmidrule(lr){1-2} \cmidrule(lr){3-8} \cmidrule(lr){9-14} 
				
				\multirow{5}{*}{\textit{CAMEL seq20}} 
				& MPSNR $\CompFore$ &   33.36 & \Valbest{  33.55} &   33.42 & \Valbest{  33.51} &   32.39 & \Valbest{  32.93} &   32.10 & \Valbest{  33.35} &   33.11 & \Valbest{  33.25} &   31.66 & \Valbest{  31.81} \\ 
				& MPSNR $\CompBack$ &   36.84 & \Valbest{  38.32} & \Valbest{  36.11} &   34.69 & \Valbest{  31.34} &   30.27 &   35.14 & \Valbest{  35.34} &   33.90 & \Valbest{  33.99} &   30.60 & \Valbest{  30.72} \\ 
				& MSSIM $\CompFore$ &   0.6779 & \Valbest{  0.6918} &   0.7118 & \Valbest{  0.7701} &   0.6654 & \Valbest{  0.6917} &   0.6002 & \Valbest{  0.6519} &   0.6469 & \Valbest{  0.6491} &   0.6140 & \Valbest{  0.6202} \\ 
				& MSSIM $\CompBack$ &   0.9424 & \Valbest{  0.9674} & \Valbest{  0.9456} &   0.9408 & \Valbest{  0.8614} &   0.8077 & \Valbest{  0.9644} &   0.9499 &   0.9476 & \Valbest{  0.9479} &   0.8780 & \Valbest{  0.8802} \\ 
				& AUC & \Valbest{  0.9948} &   0.9947 &   0.9957 & \Valbest{  0.9981} & \Valbest{  0.9909} &   0.9892 &   0.9868 & \Valbest{  0.9935} &   0.9922 & \Valbest{  0.9929} &   0.9812 & \Valbest{  0.9827} \\ 
				\cmidrule(lr){1-2} \cmidrule(lr){3-8} \cmidrule(lr){9-14} 
				
				\multirow{5}{*}{\textit{CAMEL seq21}} 
				& MPSNR $\CompFore$ &   34.73 & \Valbest{  37.97} &   35.43 & \Valbest{  35.56} &   33.60 & \Valbest{  37.16} &   32.03 & \Valbest{  34.70} &   34.02 & \Valbest{  34.46} &   31.99 & \Valbest{  35.54} \\ 
				& MPSNR $\CompBack$ & \Valbest{  36.83} &   35.05 & \Valbest{  36.92} &   30.80 & \Valbest{  31.08} &   29.75 &   34.40 & \Valbest{  35.53} &   32.84 & \Valbest{  33.00} &   29.35 & \Valbest{  31.89} \\ 
				& MSSIM $\CompFore$ &   0.7366 & \Valbest{  0.8396} &   0.7619 & \Valbest{  0.7670} &   0.7231 & \Valbest{  0.8226} &   0.6058 & \Valbest{  0.7084} &   0.6730 & \Valbest{  0.6755} &   0.6375 & \Valbest{  0.7141} \\ 
				& MSSIM $\CompBack$ & \Valbest{  0.9281} &   0.8701 &   0.9443 & \Valbest{  0.9574} & \Valbest{  0.8119} &   0.6790 & \Valbest{  0.9660} &   0.9507 &   0.9510 & \Valbest{  0.9513} &   0.8304 & \Valbest{  0.9347} \\ 
				& AUC &   0.9953 & \Valbest{  0.9991} &   0.9958 & \Valbest{  0.9965} &   0.9921 & \Valbest{  0.9989} &   0.9758 & \Valbest{  0.9919} &   0.9803 & \Valbest{  0.9830} &   0.9610 & \Valbest{  0.9943} \\ 
				\cmidrule(lr){1-2} \cmidrule(lr){3-8} \cmidrule(lr){9-14} 
				
				\multirow{5}{*}{\textit{bird1}} 
				& MPSNR $\CompFore$ &   41.28 & \Valbest{  41.91} &   41.13 & \Valbest{  48.44} &   39.94 & \Valbest{  40.81} &   39.29 & \Valbest{  39.85} &   41.60 & \Valbest{  43.47} &   42.32 & \Valbest{  42.75} \\ 
				& MPSNR $\CompBack$ &   35.31 & \Valbest{  37.52} &   37.08 & \Valbest{  38.09} & \Valbest{  29.59} &   29.58 &   35.41 & \Valbest{  39.83} &   41.45 & \Valbest{  41.77} &   28.94 & \Valbest{  29.97} \\ 
				& MSSIM $\CompFore$ &   0.9186 & \Valbest{  0.9238} &   0.9327 & \Valbest{  0.9759} &   0.9230 & \Valbest{  0.9304} &   0.8869 & \Valbest{  0.8969} &   0.9395 & \Valbest{  0.9506} &   0.9357 & \Valbest{  0.9429} \\ 
				& MSSIM $\CompBack$ &   0.8863 & \Valbest{  0.9258} &   0.9484 & \Valbest{  0.9675} &   0.7672 & \Valbest{  0.7676} &   0.8115 & \Valbest{  0.9396} &   0.9633 & \Valbest{  0.9678} &   0.7045 & \Valbest{  0.7894} \\ 
				& AUC &   0.8992 & \Valbest{  0.9047} &   0.9063 & \Valbest{  0.9835} &   0.8881 & \Valbest{  0.9241} &   0.6544 & \Valbest{  0.7140} &   0.7644 & \Valbest{  0.7703} &   0.7254 & \Valbest{  0.7736} \\ 
				\cmidrule(lr){1-2} \cmidrule(lr){3-8} \cmidrule(lr){9-14} 
				
				\multirow{5}{*}{\textit{cell1}} 
				& MPSNR $\CompFore$ &   31.72 & \Valbest{  33.70} &   32.43 & \Valbest{  33.36} &   31.04 & \Valbest{  33.13} &   30.87 & \Valbest{  31.87} &   31.57 & \Valbest{  32.09} &   30.33 & \Valbest{  31.89} \\ 
				& MPSNR $\CompBack$ & \Valbest{  35.68} &   34.59 & \Valbest{  38.97} &   38.85 & \Valbest{  30.51} &   29.40 &   36.05 & \Valbest{  36.21} &   36.13 & \Valbest{  36.86} &   30.86 & \Valbest{  38.85} \\ 
				& MSSIM $\CompFore$ &   0.5856 & \Valbest{  0.6904} &   0.6352 & \Valbest{  0.6726} &   0.5632 & \Valbest{  0.6760} &   0.5448 & \Valbest{  0.5652} &   0.5804 & \Valbest{  0.5809} &   0.5266 & \Valbest{  0.5860} \\ 
				& MSSIM $\CompBack$ & \Valbest{  0.9548} &   0.8460 & \Valbest{  0.9908} &   0.9679 & \Valbest{  0.8326} &   0.7176 & \Valbest{  0.9864} & \Valbest{  0.9864} &   0.9858 & \Valbest{  0.9915} &   0.8680 & \Valbest{  0.9963} \\ 
				& AUC &   0.8098 & \Valbest{  0.9288} &   0.8669 & \Valbest{  0.9228} &   0.7724 & \Valbest{  0.9132} &   0.7676 & \Valbest{  0.8043} &   0.8109 & \Valbest{  0.8274} &   0.7129 & \Valbest{  0.8185} \\ 
				\cmidrule(lr){1-2} \cmidrule(lr){3-8} \cmidrule(lr){9-14} 
				
				\multirow{5}{*}{\textit{cell2}} 
				& MPSNR $\CompFore$ & \Valbest{  36.68} &   36.29 &   37.60 & \Valbest{  37.90} &   36.43 & \Valbest{  37.30} &   33.74 & \Valbest{  34.51} &   37.22 & \Valbest{  38.23} &   35.28 & \Valbest{  35.81} \\ 
				& MPSNR $\CompBack$ &   38.56 & \Valbest{  40.73} & \Valbest{  40.76} &   39.12 & \Valbest{  31.88} &   31.80 & \Valbest{  43.61} &   43.01 & \Valbest{  41.51} &   41.46 &   32.17 & \Valbest{  32.35} \\ 
				& MSSIM $\CompFore$ &   0.7799 & \Valbest{  0.7819} &   0.8075 & \Valbest{  0.8217} &   0.7854 & \Valbest{  0.7994} &   0.7057 & \Valbest{  0.7290} &   0.8019 & \Valbest{  0.8179} &   0.7776 & \Valbest{  0.7857} \\ 
				& MSSIM $\CompBack$ &   0.9353 & \Valbest{  0.9692} &   0.9744 & \Valbest{  0.9782} & \Valbest{  0.8363} &   0.8318 & \Valbest{  0.9828} &   0.9804 &   0.9769 & \Valbest{  0.9779} &   0.8531 & \Valbest{  0.8600} \\ 
				& AUC & \Valbest{  0.9807} &   0.9787 & \Valbest{  0.9857} &   0.9844 &   0.9728 & \Valbest{  0.9825} &   0.9064 & \Valbest{  0.9359} &   0.9839 & \Valbest{  0.9856} &   0.9261 & \Valbest{  0.9347} \\

				\bottomrule
			\end{tabular}
		}
	\end{center}
	\vspace{-3mm}
\end{table*}